*TONGJI UNIVERSITY*

# Interactive Imitation Learning in Robotics based on Simulations

BACHELOR THESIS
For the degree of Bachlor in Engineering at Tongji University

Xinjie Liu

July 1, 2021

# Interactive Imitation Learning in Robotics based on Simulations

## ABSTRACT

The transformation towards intelligence in various industries is creating more demand for intelligent and flexible products. In the field of robotics, learning-based methods are increasingly being applied, with the purpose of training robots to learn to deal with complex and changing external environments through data. In this context, reinforcement learning and imitation learning are becoming research hotspots with their respective characteristics. However, the two have their own limitations in some cases, such as the high cost of data acquisition for reinforcement learning. Moreover, it is difficult for imitation learning to provide perfect demonstrations.

As a branch of imitation learning, interactive imitation learning aims at transferring human knowledge to the agent through interactions between the demonstrator and the robot, which alleviates the difficulty of teaching. This thesis implements IIL algorithms in four simulation scenarios and conducts extensive experiments, aiming at providing exhaustive information about IIL methods both in action space and state space as well as comparison with RL methods.

# Content

# 1. Background

In recent years, thanks to the enhancement of big data technology and computing resources, the research and application of artificial intelligence (AI) has become a hotspot in the field of robotics. Unlike traditional robot control, AI technology proposes to use neural network to approximate robot's "policy" model, which outputs different actions based on the current observation of the environment.

Among them, reinforcement learning (RL) methods[1] based on trials and errors and imitation learning (IL)[2] based on human demonstrations are the two most widely studied solutions.

However, there are several difficulties in applying most model-free reinforcement learning algorithms to robotics: (1) Reinforcement learning relies on reward and punishment measures, that is, using reward functions to motivate agents. Nevertheless, it is not easy to establish effective reward functions in some task scenarios because of the multimodality of the behaviors and the implicitness of the task objective. Especially in the high-dimensional state space, the problem of "sparse rewards" will be encountered[3][4]; (2) Model-free RL methods are in a trial-and-error manner, which may be extraordinarily sample-inefficient and take a long time to converge. At the same time, in real robot learning problems, interactions with the environment to get data can be costly. So, many real robot learning researchers find it hard to be realized with insufficient data. The famous robot learning team from OpenAI was announced to be disbanded for this reason.

Imitation learning proposes to teach the robot to learn human behavior through supervised demonstrations. However, in the real-world application, it encounters the following problems: ①The phenomenon of "distribution shift"[5], which is because of the accumulation of bias from model prediction. ② The potential causal confusion problem of behavioral cloning[6]. ③ Decision-making from a human is usually non-Markovian and multi-modal, which is unnatural to learn by agents. ④ Providing the perfect "action label" through demonstration can be hard for a human.

Regarding the difficulties of building reward functions and sample-inefficiency of RL methods, the issue of providing demonstrations in IL algorithms, some researchers proposed robot learning through interactions, such as DAgger[7], HG-DAgger[8], Deep COACH[9] (DCOACH), where HG-DAgger and DCOACH are online algorithms, while DAgger requires human to relabel the data sampled from the robot trajectory. These methods are called interactive imitation learning (IIL).

Interactive imitation learning algorithms have been proven to be efficient in many task scenarios[9], but they are all based on the demonstrator's knowledge of the agent's internal dynamics model and a good understanding of its action mechanism. Then, the demonstrator can provide feedback in the robot's action space. For example, in a task where a robot arm grabs a target object, a human demonstrator needs to provide feedback signals for the torque of each joint. According to the opinions of researchers from the University of California, Berkeley, humans or animals usually perceive the transition of state in the task more intuitively, rather than the precise actions to be taken[11]. When learning to pour water into a cup, humans usually pay more attention to the state of the cup rather than the movement of the arm.

Considering this, Snehal et al. (2020)[10] suggest that policy could also be learned directly through interactions in the robot's state space, which may be more intuitive and reduce the difficulty in providing demonstrations for non-expert humans in some cases.

Overall, this thesis implements IIL algorithms in different task scenarios, where providing feedback in state space and action space, as well as regular RL methods, are compared and analyzed. Also, combined with a convolutional neural network, IIL agent is able to yield policy from high-dimensional raw images in very few training episodes in KUKA-Reach task.

Through extensive experiments, discussion about the interpretability of IIL methods is given. Furthermore, considering the suboptimality of human demonstrations, the obtained deterministic policy through imitation may not be optimal. We can also see in the experiment results that the agents learn a policy faster than in the deterministic case by using stochastic policy.

For the rest part of this chapter, sections 1.1 and 1.2 illustrate the background of research in the RL and IL communities and related issues. Section 1.3 and 1.4 introduce the IIL methods with feedback both in action space and state space. Finally, section 1.5 gives the outline of the whole thesis.

## 1.1  Model-Free Reinforcement Learning

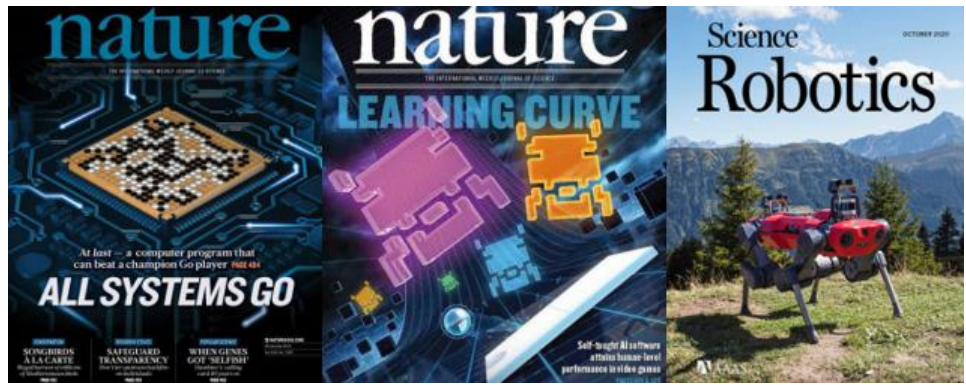

Fig. 1.1 Few applications of RL methods

In recent years, reinforcement learning has been used as an effective strategy to solve sequential decision-making problems and has become a research hotspot in artificial intelligence. The typical applications include the robot AlphaGO[12], which beat the world champion in the GO game; using RL agent to perform Atari games[13]; using RL to generate the trajectory of drone[14] autonomously, and so on.

Combined with neural networks, the deep reinforcement learning (DRL) agents are considered to be able to learn a policy in an "end-to-end" manner, skipping the feature preprocessing module, since the neural networks endow the agent with inherent feature exection ability[9].

However, it is precise because of the "end-to-end" fashion, traditional model-free DRL algorithms rely on large amounts of data to learn the policy $\pi_\theta$. Unlike in a simulation environment, interactions with the environment for getting data can be extremely expensive in real robot settings. Lacking training data and the natural sample-inefficient characteristic make model-free RL often hard to be realized in real systems. For example, in an autonomous driving scenario, one can never get data through driving randomly on the road. Besides, without human guidance during training, a trial-and-error manner makes the training procedure of RL agents quite long and challenging to converge. Even after learning an initial policy, an RL agent will still need random explorations for better policy, making the algorithms have higher variance and dangerous in some tasks.

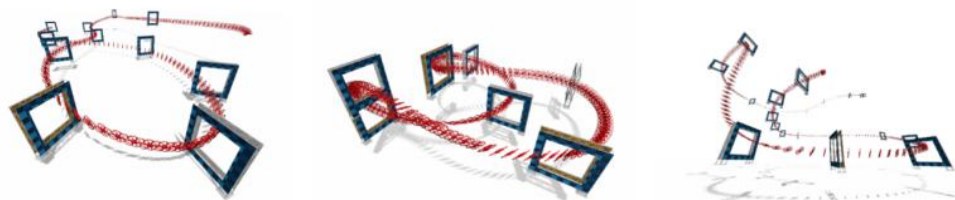

Fig. 1.2 Autonomous trajectory planning with DRL

To better construct a reward function, which can better suit and model the multi-modality and randomness of human behaviors, some researchers propose using inverse RL[19] to learn a reward function from the task environment.

The often being used model-free RL algorithms include: Policy Gradient method[15], which takes gradient ascend on the RL objective to perform optimization; Actor-Critic algorithm[16], which uses Advantage Function as evaluation and performs gradient step on the "Actor" part, aiming at reducing the variance of the training; Deep Q-Network (DQN) method[17] based on Q-Function and Value-Function. Incorporating with the thought from dynamic programming, DQN recursively optimizes the Bellman Backup to estimate future rewards better. Equation (1.1) describes the iteration of the DQN method.

$$Q^{new}\left(s_t, a_t\right) \leftarrow \underbrace{Q\left(s_t, a_t\right)}_{\text{old value}} + \underbrace{\alpha}_{\text{learning rate}} \cdot \overbrace{\left( \underbrace{r_t}_{\text{reward}} + \underbrace{\gamma}_{\text{discount factor}} \cdot \underbrace{\max_a Q\left(s_{t+1}, a\right)}_{\text{estimate of optimal future value}} - \underbrace{Q\left(s_t, a_t\right)}_{\text{old value}} \right)}^{\text{temporal difference}}$$

$$\underbrace{\phantom{r_t + \gamma \cdot \max_a Q\left(s_{t+1}, a\right) - Q\left(s_t, a_t\right)}}_{\text{new value (temporal difference target)}}$$

(1.1)

For more details of the introduction about the basic model-free RL algorithms, please refer to the Appendix.

## 1.2 Imitation Learning

### 1.2.1 Human Demonstrations as Action Labels

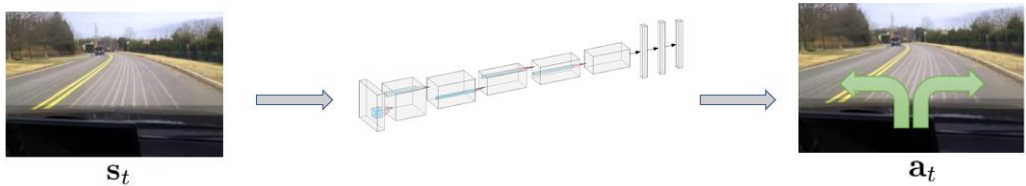

Fig. 1.3 Behavior cloning requires action labels from human

As mentioned above, reinforcement learning methods are often suffering lacking data and sample inefficiency. The IL community proposes using human demonstrations to train models. For example, in the autonomous driving scenario, cameras are used for collecting the experienced states by the agent. Then, human demonstrators are required to relabel those training data as $\{s_t, a_t\}$. Using supervised deep learning methods, a policy is trained directly to drive autonomously, called Behavior Cloning[22], and is expected to offset the data deficiency in RL methods. Also, learning through demonstration and imitation avoids the long haul of trial-and-error.

Another advantage of IL is that the state transition ($p(s_{t+1}|s_t, a_t)$, dynamics of the environment) is not necessarily known. Compared with the model-based RL methods[23], where the state transition is approximated usually by neural networks, knowledge is implicitly transferred through human demonstrations, which improves the generality between tasks. Furthermore, in IL, no perfect reward function $\mathcal{R}$ is required to form the optimization objective. Another challenge in RL is addressed.

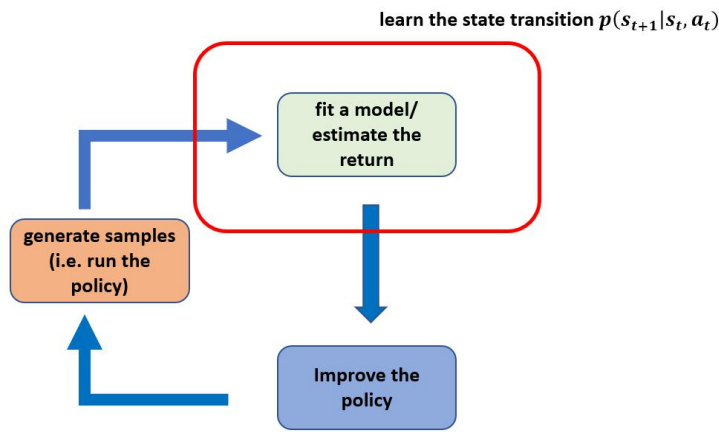

Fig. 1.4 Model-based RL learns the state transition

Combined with deep neural networks, IL can also deal with high-dimensional inputs like raw images, where autoencoders are trained to perform dimensionality reduction. For example, researchers from Nvidia[24] tried to train an autonomous car driver through raw images from cameras in an end-to-end fashion, where steering angle, controlling signals of accelerator and brake are directly being output from the neural networks.

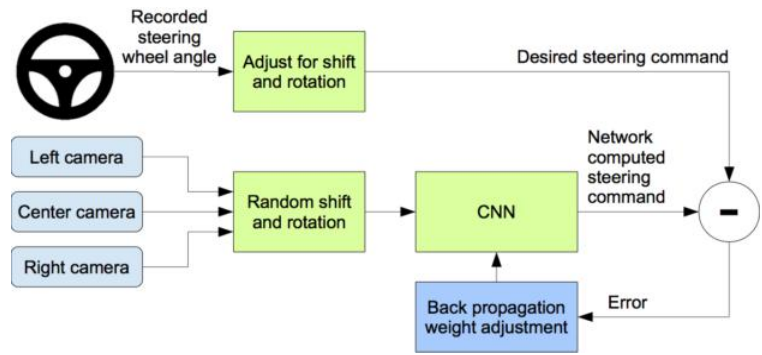

Fig. 1.5 Learning autonomous driving end-to-end from Nvidia

### 1.2.2 "Distribution Shift" and Non-Markovian Behavior

However, the naïve way of behavior cloning often fails because of its common natures[25]. One of them is the problem of "distribution shift"[7]. Models are often not perfectly precise. When the tiny errors made by the model accumulates, the shifted states, which are different from the training data, would increase the probability of making mistakes. Thus, with accumulation, the output trajectory is usually unexpected.

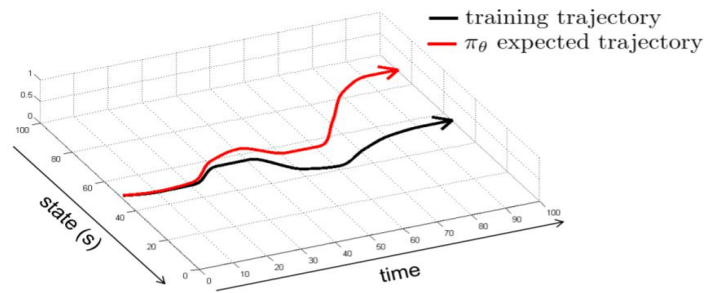

Fig. 1.6 "Distribution shift" in behavior cloning

As a solution, researchers from Nvidia using a smarter way of collecting data. As the following figure shows, three cameras are used for data collection, where the state-shift in the different cameras could compensate for each other. Although the significantly increased quantity of data requires a more prolonged training procedure, according to the videos, it works decently well.

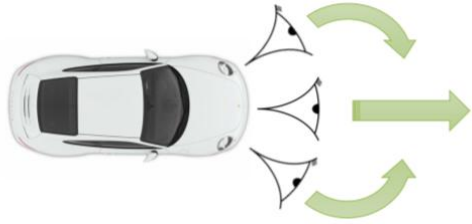

Fig. 1.7 Tackling "distribution shift" using three cameras

DAgger Algorithm

To improve IL agents' performance, researchers from Carnegie Mellon University[26]propose letting humans interactively participate in the training procedure, where they must relabel the data collected from the cameras recursively. In this way, the distribution shift is much addressed.

Table 1.1 DAgger Algorithm

| **Algorithm 1** (HG-) DAgger |
| --- |
| 1： **require**: initial demonstration database $\mathcal{D}$, policy update interval $b$ |
| 2： **for** $t = 1, 2, \dots$ **do** |
| 3：     **if** $mod(t, b)$ is 0 **then** |
| 4：         **update** $\pi_\theta$ $using\ data\ from\ D$ |
| 5：     o**bserve** state $o_t$ |
| 6：     **select** action from agent or human demonstrator |
| 7：     **execute** action |
| 8：     provide **feedback** of action label $a_t^*$ according to the observation $o_t$, when necessary |
| 9：     **aggregate** data $(o_t, a_t^*)$ to $\mathcal{D}$ |

Nevertheless, the DAgger algorithm has its natural disadvantages. First of all, for the human demonstrator, to provide the action labels for each state of the agent requires expert demonstrators. Secondly, even expert demonstrators' teaching behavior is not a Markov decision process (MDP)[27]. In other words, human behaves not only based on the current state $s_t$, but also based on their past memory. Even if they meet the same state twice, their behaviors could be different. This kind of multi-modality and randomness makes the labels provided by human demonstrators may be imperfect.

## 1.3 Interactive Imitation Learning

### 1.3.1 From Imitation to Interactive Imitation

Regarding the disadvantages of IL, a kind of interactive imitation learning (IIL) method emphasizes human participation in the training procedure for transferring human knowledge to the agents. However, humans do not need to provide the entire demonstration. Instead, the demonstrators only need to provide correction signals according to the current trajectory from the agent in a fully online manner, which addresses the issue of "distribution shift" and dramatically reduces the difficulty of giving action labels. Thus, it is more suitable for non-expert trainers. IIL methods are proven decently efficient in some task scenarios[9][10], compared with other DRL methods.

Researchers (James MacGlashan et al., 2017)[4] proved that the feedback from human demonstrators is not independent of the existing policy of the agent through the experiment of training three kinds of "virtual dogs", which adds factual basis for the rationality of IIL.

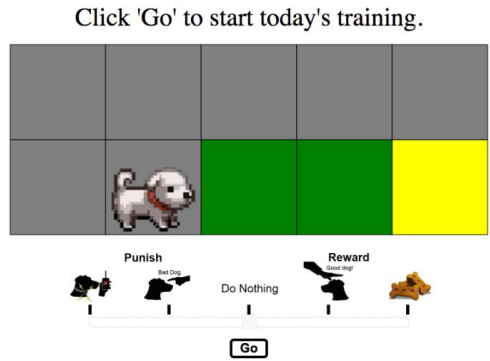

Fig. 1.8 "Virtual Dog" experiment

Then, researchers from the University of Chile (Celemin et al., 2018) proposed to train an agent based on the corrective signal provided by human demonstrators [28]. This greatly reduces the times of trial-and-error required in traditional RL algorithms and is applied to learn complex motor skills efficiently on real systems, which is validated on training a humanoid robot to play soccer.

Subsequently, researchers from the University of Chile and Delft University of Technology (Pérez-Dattar et al., 2018) proposed a COACH framework based on deep neural networks, called Deep COACH[29]. In the DCOACH algorithm, two kinds of neural network designs are employed for different task scenarios: the use of feedforward fully connected network is for the problems in low-dimensional state-space; the convolutional neural network is to address the problems of high-dimensional space, for example, the raw image data as input.

The training of a DCOACH agent is entirely online, where a so-called Memory Buffer is used to store past trajectories collected during training. The "immediate training step" and "batch training step" are both employed, aiming at making the demonstration entirely online and, at the same time, avoiding the over-fitting to the nearest trajectory.

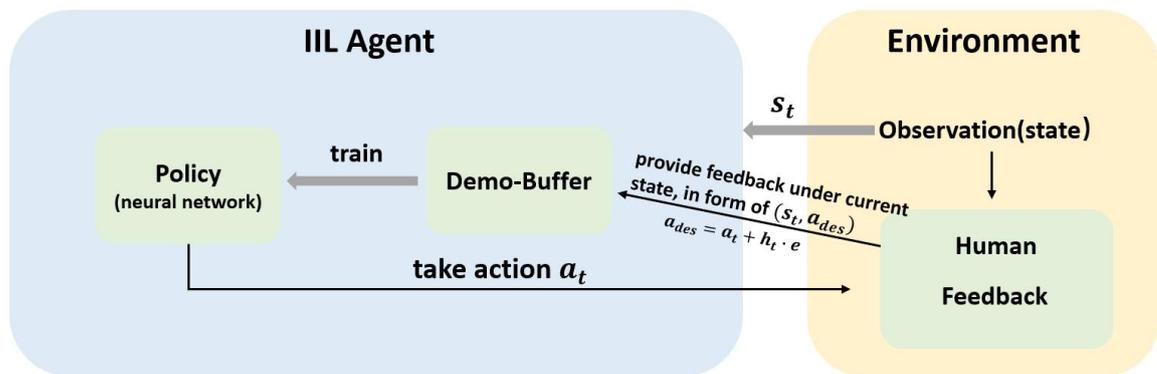

Fig. 1.9 Framework of DCOACH algorithm

The above figure illustrates the framework of the DCOACH algorithm, where a hyper-parameter $e$ (error correction constant) is defined to correct the current action executed by the agent. The corrected action, along with the current state $s_t$ form a state-action pair$(s_t, a_t)$, which is the data used to train the policy.

In the experimental part, researchers validated the proposed framework in three tasks:

(1) Cart-Pole task based on OpenAI Gym environment, which represents low-dimensional problems with discrete action space;

(2) Car-Racing game in OpenAI library, representing problems with high-dimensional inputs (raw images as input, aiming at validating the convolutional neural network with the DCOACH framework);

(3) Duckie Racing, which is a real car racing task using sensor data from the camera.

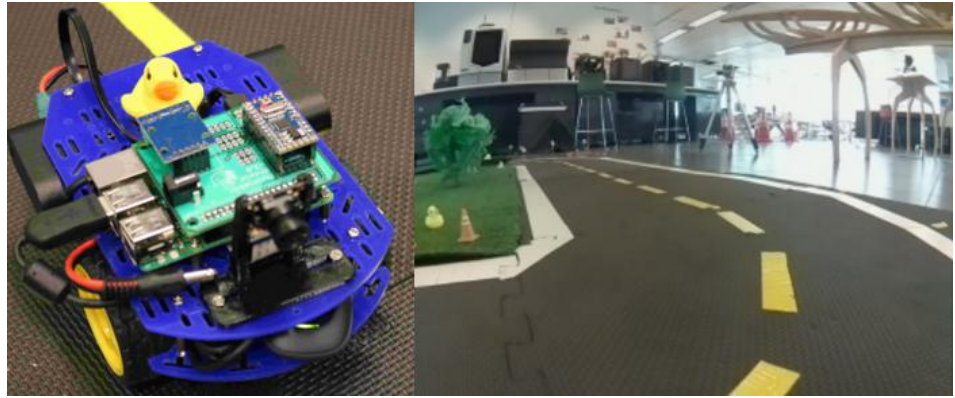

Fig. 1.10 Implementing DCOACH for solving "Duckie Racing" problem

As for the baseline, the proposed framework is compared with the Deep Deterministic Policy Gradient (DDPG).

In the experiment, the researchers first compared the performance of the proposed model with or without a memory buffer through ablation experiments, with cumulative rewards in certain rounds of training as metrics. The results verify that memory buffer improves the training operability and performance of the model by about 20% in both simulation environments. At the same time, different ratio of demonstration error from the human teacher is simulated and compared. For conclusion, in the three simulation environments, it is once again verified that DCOACH accelerates the convergence speed of the model by introducing the interactive feedback signal from the human demonstrator. Under these tasks, the final performance of the model has been nearly doubled.

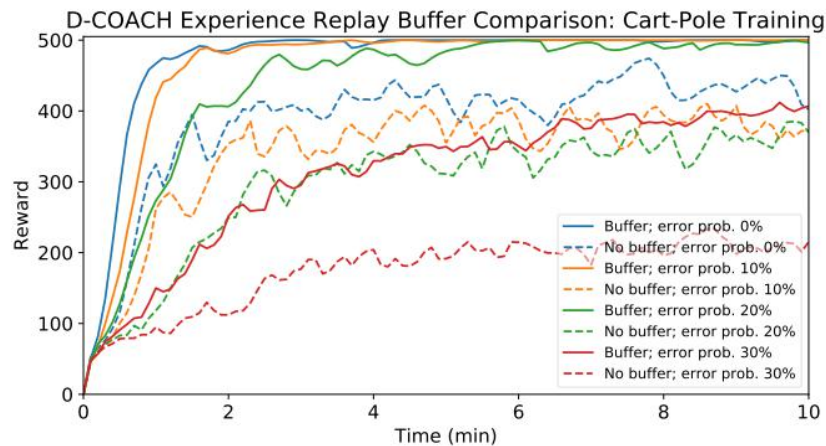

Fig. 1.11 Ablation experiment results using DCOACH

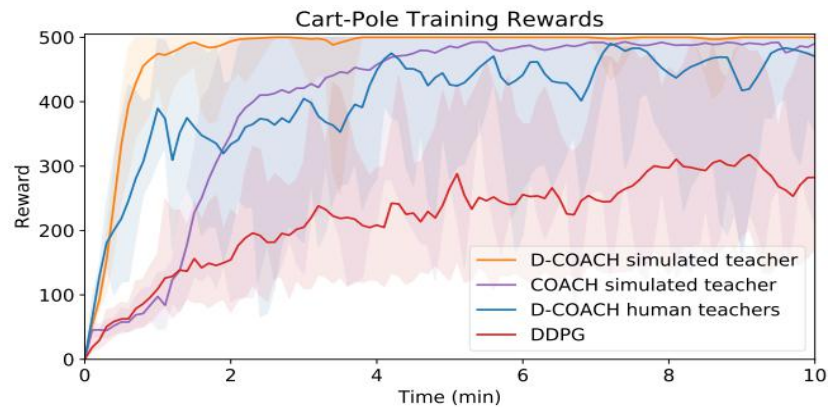

Fig. 1.12 Comparison of DCOACH and DDPG in Cart-Pole

### 1.3.2 Feedback on States

The DCOACH algorithm greatly reduces the training time by allowing human demonstrators to participate in the training process to provide feedback. At the same time, it does not require the demonstrators to demonstrate the entire action perfectly.

However, according to the opinions provided by researchers from the University of California, Berkeley, humans or animals usually percepts the transfer of states in action more intuitively when learning actions, rather than the precise actions to be made[30]. For example, when learning to pour water into a water glass, we tend to pay more attention to the state of the water glass and the kettle rather than the movements of the arms when pouring water.

The existing interactive imitation learning algorithms often transmit corrective feedback signals to the agent in the action space. This is not obvious in some tasks. For example, in the Cart-Pole problem, the state transition is equally intuitive as the action execution.

However, in specific tasks, such as controlling a multi-joint robotic arm to grab a target object, the focus should be more on the relative position of the target object and the end effector of the robotic arm. It is non-intuitive and requires specific training for human demonstrators to provide feedback in the action space of each joint. That is, **expert demonstrators are required**. By contrast, providing feedback in state space allows non-expert demonstrators to train a qualified agent.

However, it is conceivable that a new problem will be encountered: how to transfer the knowledge of ideal state transition to the agent and activate correct action execution. In the DCOACH algorithm, the corrected actions from humans can directly form the action label, and the state-action pair formed by the current state and ideal action can be used as training data to update the policy. However, in providing feedback signals based on state transitions, transforming the desired state transition into actions that the policy model should output is one of the main difficulties of this work.

In recent years, some work can be used as a reference. Inspired by model-based RL methods, researchers from the University of California, Berkeley have used convolutional neural networks to build an inverse dynamics model[30], which takes pictures of human demonstrators manipulating the rope and the image of the current shape of the rope (equivalent to $s_{t+1},s_t$) as input, inversely outputs the most possible taken action.

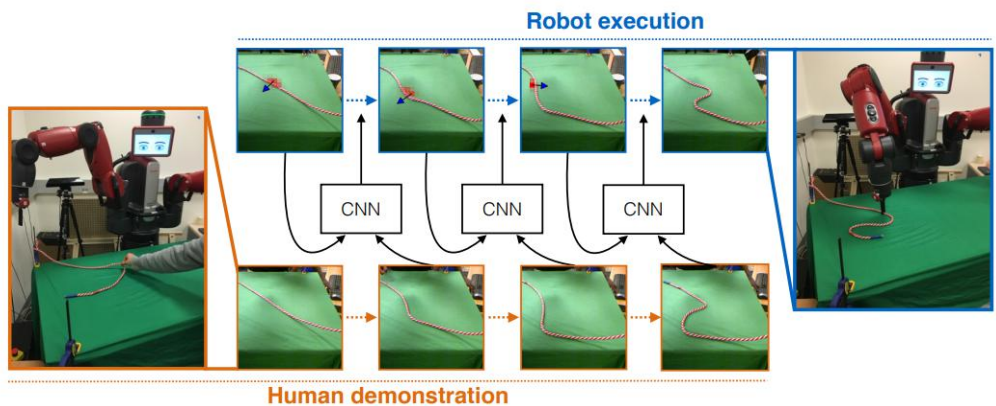

Fig. 1.13 Constructing inverse dynamics model using CNN

Furthermore, a "two-stage" training was proposed for online behavior cloning (Torabi et al., 2018)[22]. The first stage requires the agent to learn a task-independent model. The model is different from other imitation learning methods; it does not require action data from the demonstrator directly and can infer the action $a_t$ related to the state $s_t$. This method is called Behavioral Cloning from Observation (BCO). BCO avoids interacting with the environment in imitation learning, which will cause delays in the model and may be dangerous in certain scenarios (such as autonomous driving scenarios).

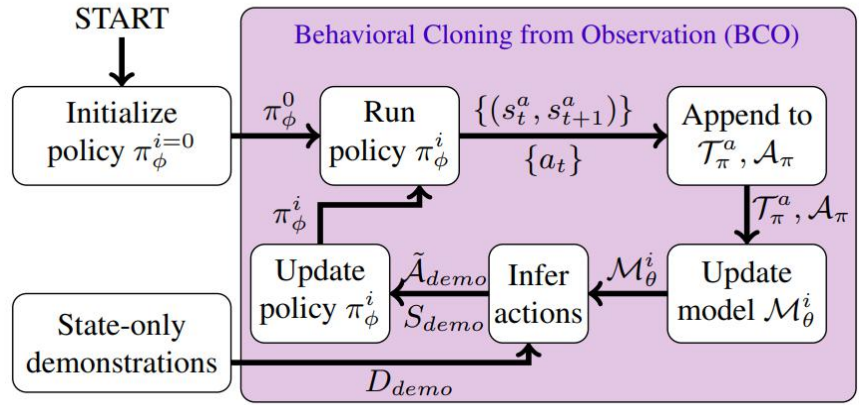

Fig. 1.14 Two-stage training in BCO

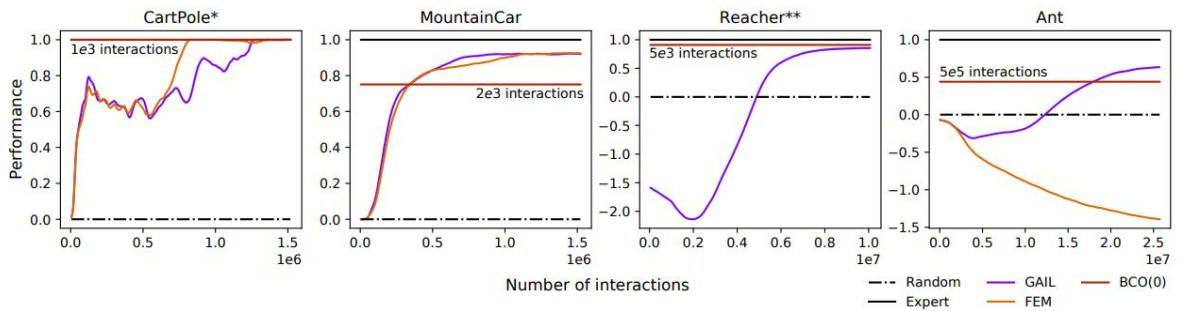

Fig. 1.15 Comparison of BCO and other algorithms in baseline tasks in OpenAI Gym

The figure above illustrates the framework of BCO. In the first stage, a random policy is executed for collecting state transition data $\{(s_t, s_{t+1}), a_t\}$, which is added to a buffer and used for training the model $M_\theta$. The model $M_\theta^i$ is able to infer the action to be taken in order to achieve the ideal state transition.

As for the experimental results, in baseline problems, it achieves equal performance as the RL methods in shorter iterations.

Snehal et al. (2020)[10] also propose to learn a forward dynamic model, which is in charge of mapping feedback in state-space back to the action space. By doing so, the teaching process could be more intuitive and does not necessarily require expert demonstrators. As the validation experiment, a KUKA robotic arm is employed to do several tasks, including reaching, putting a ball into a cup, writing.

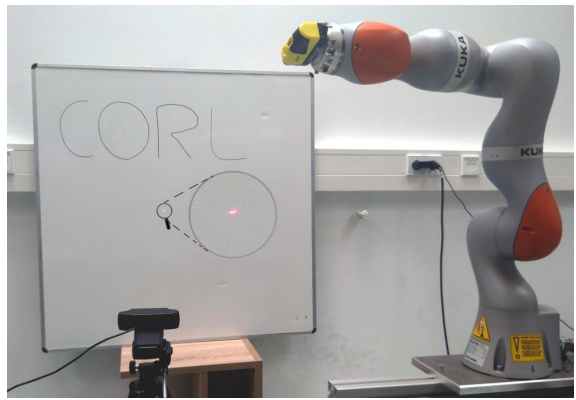

Fig. 1.16 KUKA robot is learning to write

**1.4 Description of Goal and Content**

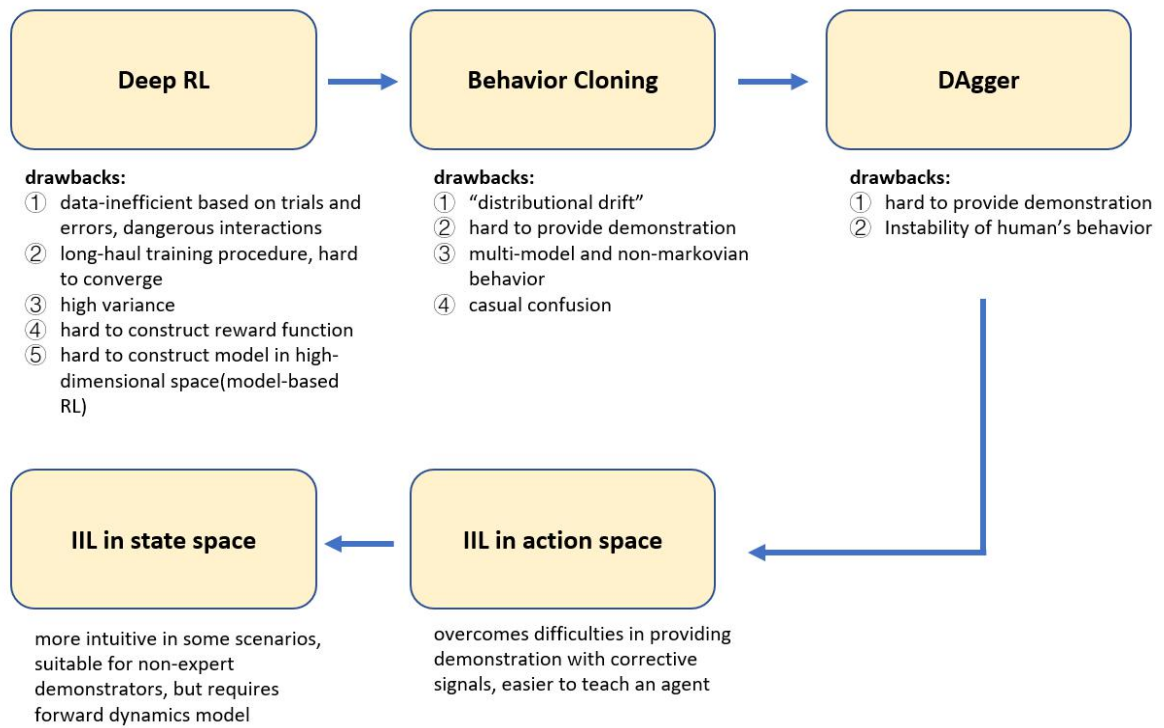

Fig. 1.17 Backgrounds summary

Summarizing the algorithms mentioned above, the conventional RL method has proved to be an effective way to solve many decision-making problems in complex state spaces. Nevertheless, due to its characteristics based on trial-and-error, it is often data-inefficient and takes an extremely long time to converge. Interacting with the real world for getting data can be pretty expensive as well. Furthermore, constructing an effective reward function tends to be tedious work in the real world.

Secondly, although IL solves the problem of getting data, it is based on the perfect demonstration provided by humans, which is unnatural for non-expert demonstrators to achieve.

Even if DAgger can tackle the "distribution shift" problem, requiring humans to constantly relabel the corresponding states also places very high requirements on demonstrators because of their non-markovian and multi-modal behavior.

By comparison, IIL methods only require human corrective signals, which significantly reduces the difficulty of demonstrating and is proved to be a practical framework. Nevertheless, in some scenarios, providing feedback in action space is regarded as non-intuitive and not corresponding the human nature. Giving feedback in state-space addresses this problem by employing the forward dynamics model, which maps the state-space-based corrective signals back to the action space.

This thesis implements IIL algorithms DCOACH and TIPS in simulation environments. Validation tasks include Cart-Pole, Reacher, Lunar-Lander, which represent control tasks with discrete action space and continuous action space in a highly dynamical scenario and its counterpart. Through extensive experiments, the two algorithms are compared with model-free RL methods. They are discussed regarding every aspect, aiming at providing exhaustive information about IIL methods both in action space and state space. Furthermore, in highly dynamic scenarios, using stochastic policy with IIL shows a better performance than its deterministic counterpart. Finally, combined with a convolutional neural network, IIL methods can also solve reaching tasks with high-dimensional raw images as input.

**1.5 Thesis Outline**

This thesis is composed of 8 chapters. The following content is arranged as:

    (1)   Chapter 2 mainly introduces environments and libraries, which have been used in this article, as well as the validation task settings.

    (2)   Chapter 3 illustrates the details of IIL algorithm DCOACH as well as the implementations in different tasks.

    (3)   Chapter 4 gives the details of the state-space-based counterpart TIPS. The implementations are also included.

    (4)   The experimental part is chapter 5, where extensive experiments are carried out. Moreover, further hypothesis-driven experiments have been conducted, along with further discussion and analysis regarding the problems observed in the experiments.

    (5)   Chapter 6 concludes the IIL methods and rethinks the whole research process.

    (6)   References are listed in chapter 7.

    (7)   Chapter 8 is the Appendix, where more experimental details and basic model-free RL methods are introduced.

# 2. Environment and Scenarios

## 2.1 Overview

As follows, this article's implementations are mainly based on the Operating System Ubuntu 18.04 LTS, where the two kinds of agents (DCOACH and TIPS) are validated in different task scenarios based on OpenAI Gyms and MUJOCO as well as PyBullet. The numerical computation is conducted using the library TensorFlow 1.14.0 and Numpy. In the data-analysis and visualization part, Matplotlib and Seaborn are employed.

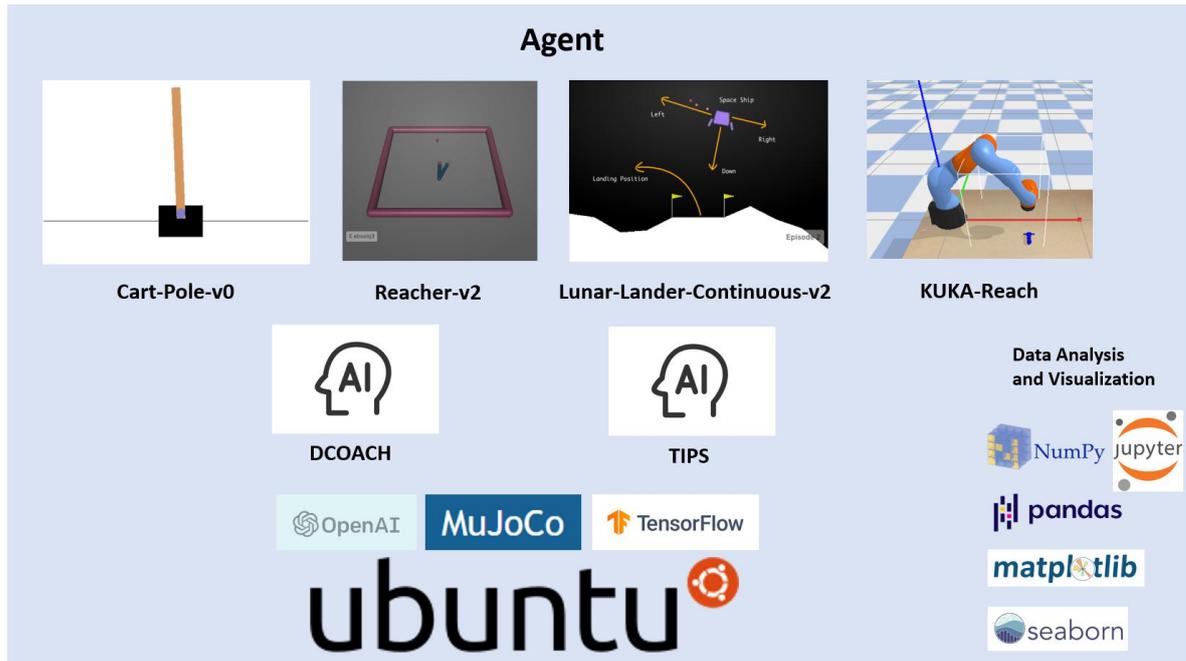

Fig. 2.1 Platforms and libraries used in this thesis

## 2.2 RL Libraries

### 2.2.1 OpenAI Gym

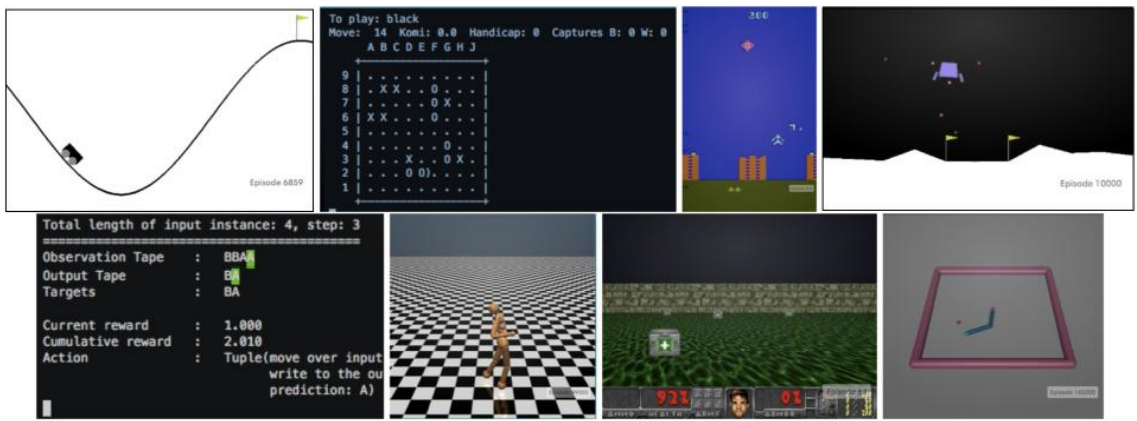

Fig. 2.2 Few task scenarios in OpenAI Gym

OpenAI Gym[31] is a reinforcement learning library that supports Python as the programming language.

It has designed various task scenarios, including classic control tasks (such as Cart-Pole, Car-Racing, etc.), Atari games, robot control scenarios based on MUJOCO component, which is similar to real-world dynamics. In all the tasks, the designed task, reward, action, and state-space are available. Also, it is possible to customize one's task scenario. Moreover, OpenAI Gym is compatible with any mainstream numerical computing library, such as TensorFlow, PyTorch, etc. People can also use the interface between OpenAI Gym and Robot Operating System (ROS) to control their robots with ROS and perform reinforcement learning algorithm tests in simulation scenarios.

In addition, the OpenAI Gym community allows people to share their algorithms and compare the performance of algorithms, which implements an ecosystem based on reinforcement learning.

### 2.2.2 MUJOCO

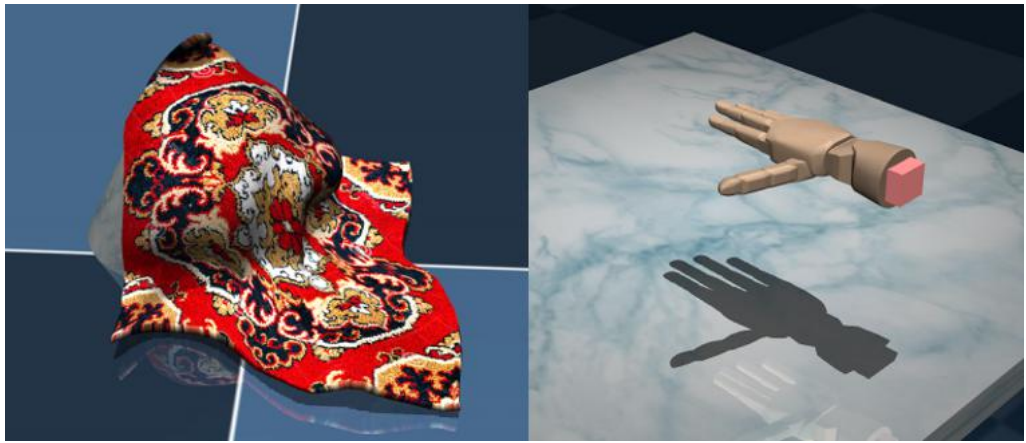

Fig. 2.3 Few scenarios in MUJOCO

MUJOCO[32] is a physical simulation engine. Based on OpenAI Gym, it strengthens the support for models in numerical optimization, including faster simulation speed, accuracy, and better modeling capabilities. It primarily supports complex dynamic systems with rich contact behaviors by promoting the simulation of computationally intensive technologies (such as optimal control, state estimation, system identification, etc.) in robotics, biomechanics, graphics, and animation.

### 2.3 Task Scenarios

A goal of this article is to apply IIL algorithms to solve three typical task scenarios in OpenAI Gym, namely: Cart-Pole v0, Reacher v2, Lunar-Lander-Continuous-v2, and to compare and analyze the differences between IIL algorithms and RL methods.

### 2.3.1 Cart-Pole-v0

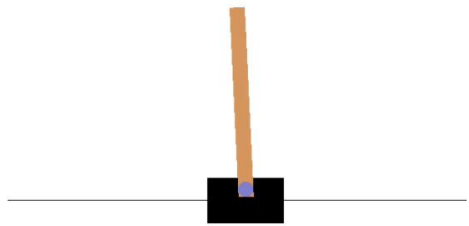

Fig. 2.4 Cart-Pole-v0

In the Cart-Pole scenario, the agent to be controlled is composed of a cart and a swing rod (pole) that are connected. The cart can move left and right freely. The system controls the cart by applying a force of +1 (representing right) or -1 (representing left) to the cart. The goal is to keep the pendulum as vertical as possible and keep the system in the middle of the screen.

After each time step, the system will get a +1 reward. The round ends when the angle between the swing bar and the vertical direction exceeds 12°, or the car touches the edge of the screen.

When the agent can get an average of 195.0 or more rewards for 100 consecutive rounds, the task is considered "solved."

Table 2.1 Action space of Cart-Pole-v0

| Dimension | Action |
|---|---|
| **0** | 0 (unit force towards the left) |
| | 1 (unit force towards the right) |

Table 2.2 State space of Cart-Pole-v0

| Dimension | State |
|---|---|
| **0** | Position of the cart |
| **1** | Velocity of the cart |
| **2** | Angle between the pole and the vertical direction |
| **3** | Tip velocity of the pole |

## 2.3.2 Reacher v2

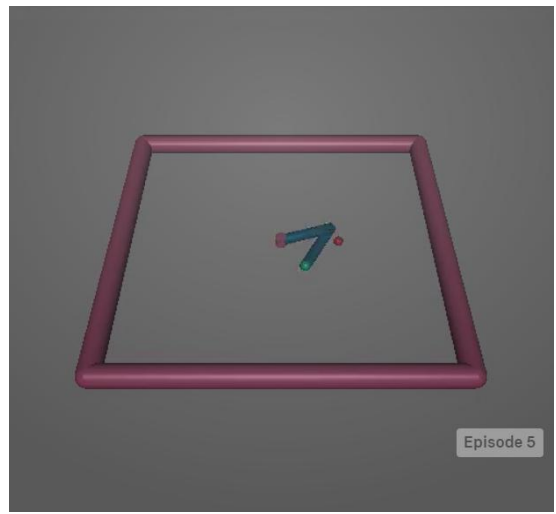

Fig. 2.5 Reacher-v2

Reacher -v2 is a simulation environment based on MUJOCO in OpenAI Gym. A square fence is located in the center of the screen. The controlled agent is a two-degree-of-freedom robotic arm whose fixed end is hinged to the center. The objective is to use the end effector of the robotic arm to touch the target object.

This section aims to compare the difficulties in providing demonstration when using the two algorithms and related model performance. Thus, we use the modified version of Reacher task:

  (1)  Removing the location information of the object, achieving the goal only depends on the corrective signals providing by a human.

(2) Keeping the initial location of the target in every episode fixed, aiming at shortening the training procedure.

The reward in this scenario is a negative value representing the distance between the end effector of the robotic arm and the target. The state space and the action space are as following:

Table 2.3 Action space of Rearcher-v2

| Dimension | Action |
|-----------|--------|
| **0** | Motion of the first joint (continuous), a positive value means anticlockwise rotation |
| 1 | Motion of the second joint (continuous), a positive value means anticlockwise rotation |

Table 2.4 State space of Reacher-v2

| Dimension | State |
|-----------|-------|
| **0** | Cosine value of the first joint |
| **1** | Cosine value of the second joint |
| **2** | Sine value of the first joint |
| **3** | Sine value of the second joint |
| **4** | The x coordinate of the target (removed) |
| **5** | The y coordinate of the target (removed) |
| **6** | The x component of velocity of the end effector |
| **7** | The y component of velocity of the end effector |
| **8** | The x component of the vector pointing from target to the end effector (removed) |
| **9** | The y component of the vector pointing from target to the end effector (removed) |
| **10** | The z component of the vector pointing from target to the end effector (removed) |

### 2.3.3 Lunar-Lander-v2

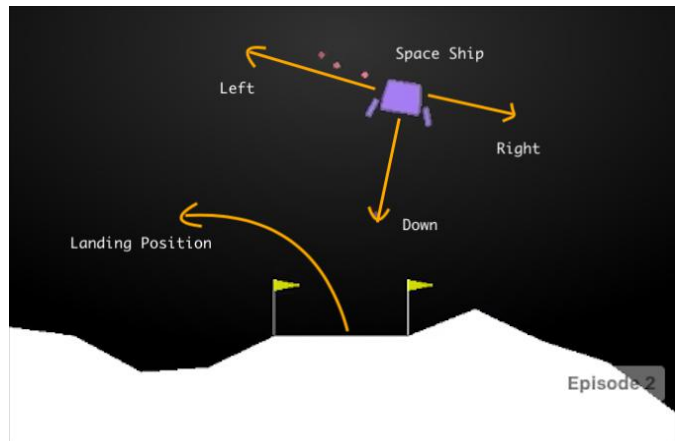

Fig. 2.6 Lunar-Lander-v2

The control object in Lunar-Lander is a spacecraft equipped with engines in three directions. The goal is to land the spacecraft smoothly on the landing platform at (0, 0) coordinates. The initial state of the spacecraft is arbitrary; that is, it may have a considerable speed and deflection angular velocity. The agent must learn to adjust for different initial states and finally achieve a smooth landing.

If the agent stops on the landing platform with a correct posture and state, it will receive a reward of 100-140 points. If the landing is stable, it will receive an additional 100 points' reward. Among them, the spacecraft has unlimited fuel, but firing the main engine (pushing the spacecraft upward) will get -0.3 points for each timestep. When the spaceship crashes (-100 points' reward) or flies out of the screen (failure), or lands smoothly (success), this round is considered to be over. When the number of cumulative rewards for this round exceeds 200, the problem is considered "solved."

There are two versions of Lunar-Lander scenario, the one with discrete action space is as follows, where only single actions are allowed to be performed at each timestep:

Table 2.5 Lunar-Lander-v2 with discrete action space

| Dimension | Action |
| --- | --- |
| **0** | Doing nothing |
| **1** | Firing the left auxiliary engine |
| **2** | Firing the main engine |
| **3** | Firing the right auxiliary engine |

By contrast, in the continuous version of Lunar-Lander, the agent is allowed to take multiple actions at the same time, such as starting the main engine and the left engine. The action space is no longer a one-dimensional binary array of length 4 but a length of 2 that is continuous on the interval [-1, 1]. The first number controls the main engine. In the interval [-1, 0], the main engine is turned off, and in the interval (0, 1], the main engine is turned on with 50%-100% power. The second number controls two sub-engines. In the interval [-1, -0.5) and (0.5, 1], the left and right engines are turned on. The sub-engines are turned off in the rest of the interval.

Table 2.6 Lunar-Lander-v2 with continuous action space

| Dimension | Action | Detail |
|:---:|:---:|:---:|
| **0** | Main engine | [-1, 0] shutting down，(0, 1] firing (50%-100) |
| **1** | Auxiliary engine | [-1, -0.5) firing the left engine，(0.5, 1]firing the right engine，the rest part means shutting down both engines |

Table 2.7 State space of Lunar-Lander-v2

| Dimension | State |
|:---:|:---:|
| **0** | Horizontal coordinates |
| **1** | Vertical coordinates |
| **2** | Horizontal velocity |
| **3** | Vertical velocity |
| **4** | Deflection angle |
| **5** | Angular velocity |
| **6** | Touching ground of foot 1 |
| **7** | Touching ground of foot 2 |

The state space is the same in both versions, where an array of length 8 constitutes the representation. The last two digits are binary, while the others are continuous on the interval [-1, 1].

### 2.3.4 KUKA-Reach

KUKA-Reach[1] is a customized simulation scenario based on the PyBullet, where a KUKA robotic arm is set to reach the target object on the desk. In this task setting, a high-dimensional raw image is used as the input.

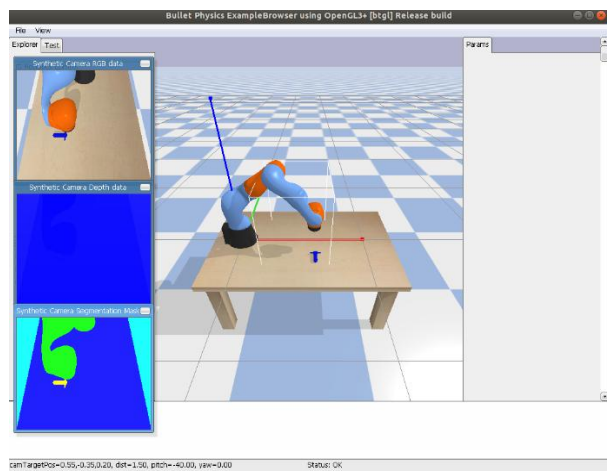

Fig. 2.7 KUKA-Reach with raw images input

Similar to the reaching task in Reacher-v2, the reward is the terminated distance between the end effector of the robotic arm and the target, where the closer they are, the lower the reward is, the better the performance is.

---

[1] https://github.com/borninfreedom/kuka-reach-drl

Differently, in this task setting, the position of the target object is random between 4 choices. The agent should learn to classify the location of the object for planning the trajectory. The executable actions are moving the end effector towards six directions along the x, y, and z axes. In the meantime, the robot should keep the whole trajectory inside the 'white box' to limit the workspace.

Note that, in this task setting, the motions of the different joints are coupled together. Thus, we only used DCOACH algorithm in this scenario.

**2.4 Summary**

This chapter mainly introduces the environment and platforms used in this thesis. Section 2.3 illustrates the task scenarios, including their action space, state space as well as objective. The following two chapters will be theoretical introductions about the two frameworks used in this thesis and the details about their implementation.

# 3. IIL Based on Action Space (DCOACH)

## 3.1 Deep COACH Algorithm

As mentioned before, part of the implementations of this article is mainly based on the IIL algorithm Deep COACH. In this section, the structure and training mechanism of DCOACH are briefly introduced.

### 3.1.1 Model Structure

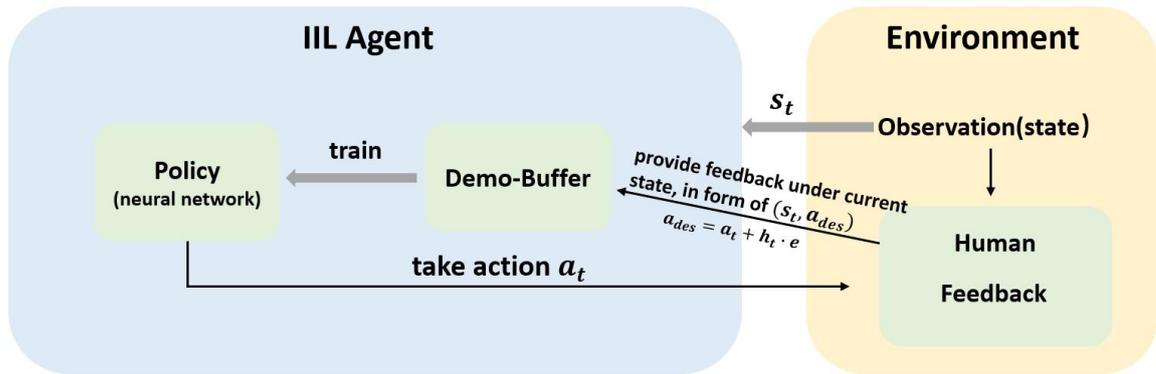

Fig. 3.1 Framework of DCOACH

The execution process of the entire DCOACH algorithm includes: ① the agent, ② the human demonstrator, who provides correction signal during the training process (often an expert demonstrator), ③the environment, where the agent performs the task.

Moreover, the agent itself contains the following:

① policy $\pi_\theta(a_t|s_t)$, where $\theta$ is the parameters of the neural network. The policy could either be deterministic or stochastic.

② Demo-Buffer, which is a container for training data $\{s_t, a_{des}\}$. Note that the action $a_{des}$ is derived from the human correction and will be used in both the immediate training step and mini-batch training step, aiming at making the training process fully online while avoiding local overfitting.

③ Feedback mechanism, where the human teacher only gives reinforce signal rather than demonstrating the whole trajectory.

### 3.1.2 Neural Network Architecture

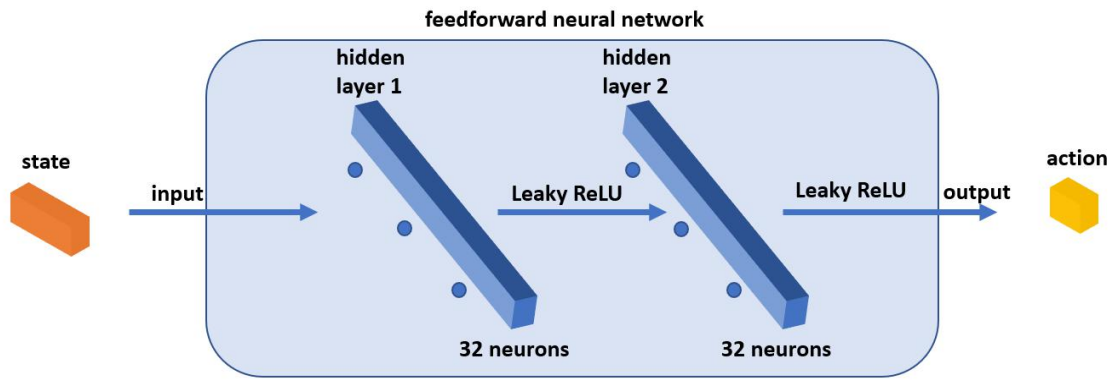

Fig. 3.2 The neural network of the policy

The DCOACH algorithm has two network designs. One is to use fully connected layers as policy approximators for tasks in relatively low-dimensional state space. Given the state data $s_t$ as input, the network outputs action in a deterministic policy or an action distribution in a stochastic setting. Another is to use several convolutional neural network (CNN) layers as encoders, taking high-dimensional data, e.g., raw images, as input and generating actions.

As the activation function, Leaky ReLU is employed between layers. Also, in relatively simple task settings with discrete action space, such as Cart-Pole, there are 16 neurons on each fully connected layer, whereas 32 neurons are on each layer in other tasks.

### 3.2 Training Procedure

Table 3.1 DCOACH Algorithm

| **Algorithm 2** Deep COACH |
| --- |
| 1： **require**:  error correction constant $e$，buffer update interval $b$ |
| 2： 初始化：$\mathcal{B} = []$ # initialize memory buffer |
| 3： **for** $t = 1, 2,...$ **do** |
| 4：     **observe** state  $o_t$ |
| 5：     **execute** action $a_t = \pi_\theta(o_t)$ |
| 6：     **feedback** providing # corrective signal $h_t$ from human demonstrator |
| 7：     **if**  $h_t$ **is not** 0 **then** |
| 8：         $error_t = h_t \cdot e$ |
| 9：         $a_t^{target} = a_t + error_t$ |
| 10：        **update** policy $\pi$ (with state-action pair $\left(o_t, a_t^{target}\right)$ and SGD) |
| 11：        **update** policy $\pi$ (with batch data sampled from buffer $\mathcal{B}$ and SGD) |
| 12：        **append** $\left(o_t, a_t^{target}\right)$ to $\mathcal{B}$ |
| 13：    **if** $mod(t, b)$ **is** 0 **then** |
| 14：    **update** policy $\pi$ (with batch data sampled from buffer $\mathcal{B}$ and SGD) |
| 15：    **if** iteration **is** over： |
| 16：        **run** 9 testing episodes, average rewards as metrics |

The above table is the standard procedure of the entire DCOACH algorithm. Also, the important hyperparameters include: ①error correction factor $e$, which defines the strength of the reinforce signal, ②interval $b$ for updating buffer, ③batch size $N$, and buffer size $K$.

The general training process takes as following:

① First, initialize the Domo-Buffer and generate a new instance of the agent. In the meantime, reset the simulation environment. Before every episode starts, the human demonstrator will be reminded.

② Let the agent run its policy. Because the agent executes a random policy in the very first beginning, every episode will fail very fast.

③ The human teacher observes the trajectory from the agent and provides correction signals $h_t$  when necessary. If $h_t$ is not 0, it means that the action should be corrected. The ideal action will be computed $a_{des} = a_t + h_t \cdot e$. That is, correction feedback will be added to the original action.

The correction signals from the human teacher are sent by a keyboard via *pyglet* library of Python. Note that it distinguishes between a long press and a short press. A counter named $h\_counter$ will record the count of timesteps with correction feedback.

④ In the next time step, the agent will execute  $a_{des}$, or just  $a_t$, in the case that human teacher did not provide any correction.

⑤ Training steps include immediate training step using single state-action pair $(s_t, a_t)$  and mini-batch training step using a batch of state-action pairs sampled from the Demo-Buffer. The goal of taking both training step is to make the training online and, at the same time, avoiding local overfitting of a certain trajectory. After training, new state-action pair  $(s_t, a_t)$  will be loaded into the Demo-Buffer to update data.

⑥ After every episode, nine extra episodes would be conducted without a human teacher, aiming to fairly evaluate the current policy performance. The average reward from the nine episodes will be calculated as the metrics. Also, the ratio of timesteps with human feedback (feedback rate) will be given via $h\_counter / t\_counter$.

⑦ Every certain interval, the policy will be updated again with mini-batch data.

**3.3 DCOACH Implementations in Different Task Scenarios**

    3.3.1 Feedback Signals

In the task settings of this thesis, feedback signals from human demonstrators are provided through a keyboard, for which the Python library *pyglet* is employed. For using the library, one should import the 'key' object from pyglet.window and set the operations through the keyboard interface given by the OpenAI Gym library, including the key pressing operation and key releasing operation.

Take the simplest feedback setting in Cart-Pole environment as an example. The code is as follows:

```python
# set global variables representing different corrective signals
H_NULL = 0 # feedback value, for distinguishing different signals
H_UP = 1
H_DOWN = 2
H_LEFT = 3
H_RIGHT = 4
H_HOLD = 5
DO_NOTHING = 6

# define key-pressing related signals
def key_press(self, k, mod):
        if k == key.SPACE:
                self.restart = True
        if k == key.LEFT:
                self.h_fb = H_LEFT
        if k == key.RIGHT:
                self.h_fb = H_RIGHT
        if k == key.UP:
                self.h_fb = H_UP
        if k == key.DOWN:
                self.h_fb = H_DOWN
        if k == key.Z:
                self.h_fb = H_HOLD
        if k == key.X:
                self.h_fb = DO_NOTHING

# define key-releasing, such that all signals are cleared
def key_release(self, k, mod):
        if (k in [key.LEFT, key.RIGHT, key.UP, key.DOWN, key.Z, key.X]):
                self.h_fb = H_NULL
```

### 3.3.2 Cart-Pole-v0

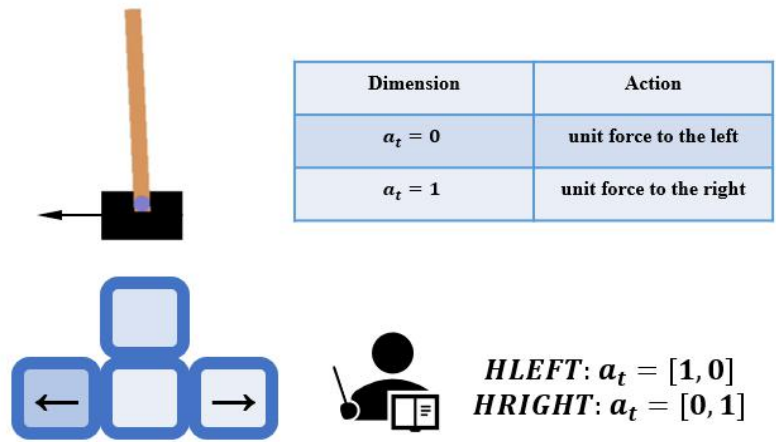

| Dimension | Action |
|---|---|
| $a_t = 0$ | unit force to the left |
| $a_t = 1$ | unit force to the right |

$HLEFT: a_t = [1, 0]$
$HRIGHT: a_t = [0, 1]$

Fig. 3.3 Feedback signals in Cart-Pole-v0

The action space in Cart-Pole-v0 is a discrete binary number, which represents applying force to the cart to the left (0) and the right (1). In this case, the feedback signal provided by the instructor is also very simple. There is no need to use the error correction constant $e$ here. HLEFT represents that the left arrow key is pressed, HRIGHT represents the human demonstrator pressing the right arrow key.

During the training process, the expert demonstrator observes the action of the agent and gives correction signals. The signal is in the form of one-hot encoding ([1, 0] to the left, [0, 1] to the right). Corrected action along with the current state form a state-action pair $(s_t, a_t)$, which is used as training data.

At each time step, if no feedback signal is provided, the agent will execute the action output from the neural network. Otherwise, the agent takes the encoded action (here, because the action is a discrete binary number, no desired action needs to be computed. It can be seen as a demonstrator can directly control the robot).

After the entire training, nine extra experiments will be conducted. The average reward and feedback rate will be computed as metrics (feedback rate is from the episode with human teacher).

### 3.3.3 Reacher-v2

| Dimension | Action |
|---|---|
| $a_t[0]$ | motion of the first joint (continuous). positive value means anticlockwise rotation |
| $a_t[1]$ | motion of the second joint (continuous). positive value means anticlockwise rotation |

$$\begin{cases} HLEFT: a_{des}[1] = 0, a_{des}[0] = a_t[0] + e \\ HRIGHT: a_{des}[1] = 0, a_{des}[0] = a_t[0] - e \\ HUP: a_{des}[0] = 0, a_{des}[1] = a_t[0] - e \\ HDOWN: a_{des}[0] = 0, a_{des}[1] = a_t[0] + e \end{cases}$$

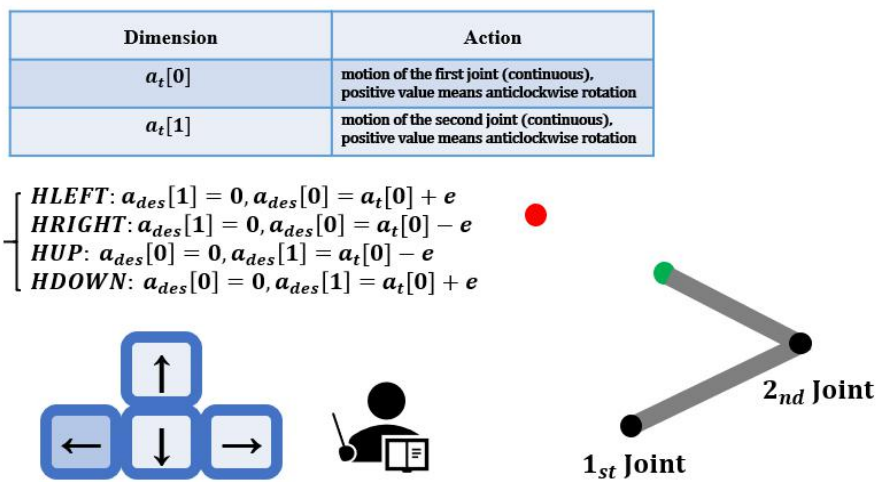

Fig. 3.4 Feedback signals in Reacher-v2

In Reacher-v2, the action space of the agent is an array of length 2 with continuous values. The first position of the array controls the rotation of the first joint of the robotic arm, where $a_t[0] > 0, a_t[1] > 0$ means that the corresponding robotic arm rotates counterclockwise.

Here the feedback signals are given to ①the first section of the robotic arm, to control the left and right movement of the end effector; ②the second section of the robotic arm, to control the up and down movement of the end effector.

When the human demonstrator presses the 'left' arrow key as feedback, the $a_{des}$ is to zero the rotation of the second section while rotating the first section of the robotic arm counterclockwise. After that, press the 'up' arrow key, the $a_{des}$ will be to zero the rotation of the first section and to rotate the second part clockwise.

Note that there could be different feedback settings along with different positions of the reaching goal. For instance, when the tiny ball locates up right to the robotic arm, the feedback signal 'up' should correspond to the anticlockwise rotation of the second section of the robotic arm, which is different from the case in our experiments. This kind of hard-coded setting also limits the generalization ability of DCOACH algorithm.

```
# feedback signals in Reacher
if(self.feedback.query(h_t) != 0):

        h_counter += 1 # feedback counter

        # action correction
        if (h_t == H_LEFT):
                a[0] += self.errorConst  # the first joint anticlockwise
                a[1] = 0                 # the second joint fixed
        elif (h_t == H_RIGHT):
                a[0] -= self.errorConst  # the first joint clockwise
                a[1] = 0                 # the second joint fixed
        elif (h_t == H_UP):
                a[0] = 0                 # the first joint fixed
                a[1] += self.errorConst  # the second joint anticlockwise
        elif (h_t == H_DOWN):
                a[0] = 0                 # the first joint fixed
                a[1] -= self.errorConst  # the second joint clockwise
```

### 3.3.4 Lunar-Lander-v2

As mentioned before, Lunar-Lander-v2 has two versions: discrete action space and continuous action space. In the discrete version of Lunar-Lander, the action vector is a one-dimensional array of length 4, where each bit controls a rocket engine (except the 'down' arrow means doing nothing). In the continuous version of Lunar-Lander, the action vector is an array of length 2, with each bit being continuous on the interval [-1, 1]. The following describes the implementation of the feedback mechanism and the details of the algorithm for the two versions.

A.  Discrete Action Space

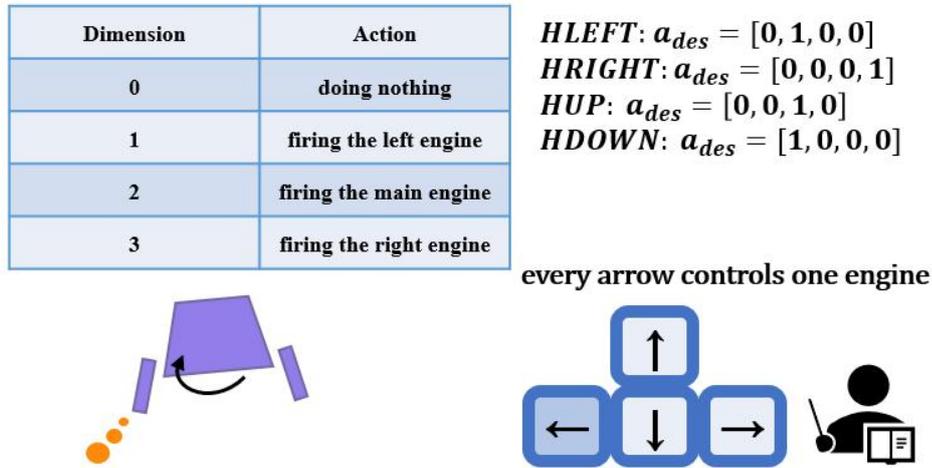

Fig. 3.5 Feedback signals in Lunar-Lander with discrete action space

In the discrete version of Lunar-Lander, the agent can only perform a single action at each time step, such as turning on the main engine to push the spacecraft upwards. In order to achieve a combined action such as moving up and left, the human teacher should firstly press the 'right' arrow key to fire the right auxiliary engine. After the agent is titled towards the left, firing the main engine will push the spaceship up and left. Here the four arrow keys correspond to three engines and doing nothing.

B.  Continuous Action Space

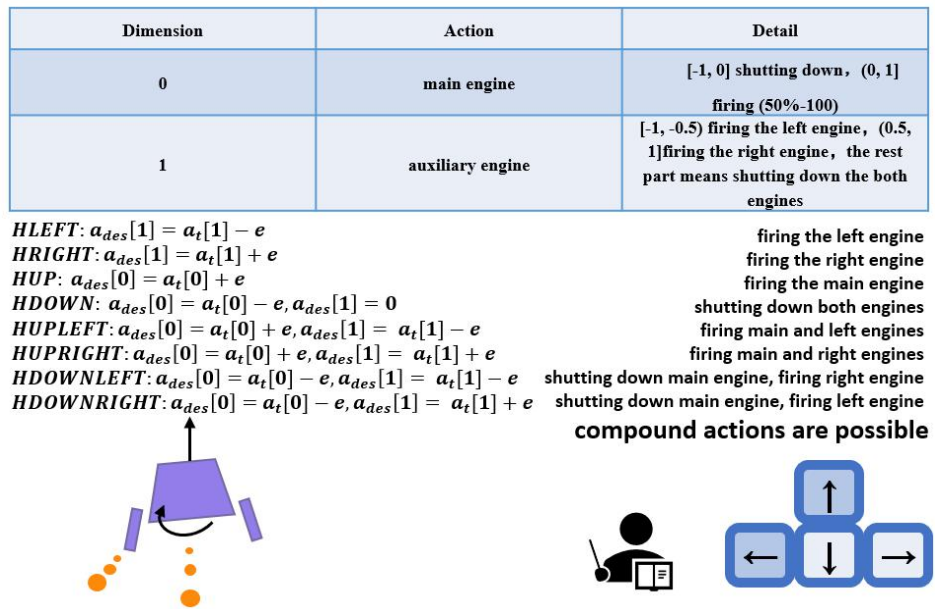

Fig. 3.6 Feedback signals in Lunar-Lander with continuous action space

In Lunar-Lander with continuous action space, the vector representing the action is an array of length 2. The feedback mechanism is designed as shown in the figure, where different actions can be performed at a single time step to achieve combined results.

```python
'''
update policy after action correction
'''
at = np.reshape(at, [-1, self.action_dim])        # immediate training steps
for itr in range(5):
        self.immediate_training_step(st, at)

self.DemoBuff.append((st[0], at[0]))              # append state-action pair

if (len(self.DemoBuff) > self.maxDemoBuffSize): # if demo-buffer is full,       #
                                                  discard the oldest data
        self.DemoBuff.pop(0)

for itr in range(5):                              # batch training steps
        self.train_policy_feedback()

at = np.reshape(at, [-1])                          # reshape the action array
A = np.copy(at)

'''
execute action and extra batch training step
'''
# execute the corrected action
state, reward, terminal, _ = self.env.step(A)

# calculate the total rewards
total_reward += reward

# reshape the state array as input of the model
state = np.reshape(state, [-1, self.state_dim])

# write down related data
self.transition_writer.write(str(st)+', '+str(A)+', '+str(total_reward)+', '
'+str(state)+', '+str(self.loop_counter)+'\n')

# extra batch training steps every b timesteps
if t_counter % self.update_interval == 0:
        for itr in range(5):
                self.update_policy_feedback()

# time counter
t_counter += 1

# display the current reward and write down related data
print('episode_reward: %5.1f' % (total_reward))
self.log_writer.write("\n" + "episode_reward: " + format(total_reward, '5.1f'))

# return the feedback rate as metrics
feedback_rate = h_counter/t_counter
return feedback_rate
```

As for the training procedure, continuous action space demands more training data and takes longer to converge. In order to make the training still online, after every time step, multiple training steps are taken, including immediate training steps and batch training steps. After each episode, nine extra experiments are conducted to evaluate the current performance of the model.

**3.4 Summary**

So far, we have discussed the DCOACH framework and its implementations in different scenarios. The next chapter will introduce its counterpart TIPS, which is based on feedback in state space.

# 4. IIL Based on State Space (TIPS)

As mentioned earlier, for Interactive Imitation Learning algorithms, in some scenarios, if the feedback can be provided based on the state space, it may significantly reduce the difficulty of teaching and improve the quality of demonstration. For example, in Reacher task, it is more intuitive to give correction directly in state-space since one can only focus on the goal's location rather than the action taken by the robotic arm itself. Moreover, in the case of DCOACH, the correction signals may vary with different goal positions. That is, the 'up' signal may correspond to different actions when the ball locates differently. This may require a hard-coded design of the feedback mechanism and can erode the generalization of the algorithm.

However, the benefits of providing feedback in state-space inevitably require a certain price. The most intuitive one is the increase in model complexity. The feedback signal in the action space can be directly added to the original action output by the policy to form the desired action. The human teacher can even control the robot directly by giving feedback in discrete action space scenarios. Nevertheless, this is not easy to accomplish based on feedback in state space. In the case of state space, the feedback given by the demonstrator is a kind of desired state, e.g., desired velocity or position of the agent, which cannot be executed directly. Thus, a mapping method is necessary to transfer the intuitive, desired state into action space.

In this regard, a forward dynamics model is employed, which learns the state transition of the environment, takes the current state-action pair $(s_t, a_t)$ as input and predicts the next state $s_{t+1}$. Under the hood, it can help choose the best action candidate to take and map the corrected state into action space. This algorithm is called TIPS (Interactive Imitation Learning from Feedback in State-space)

## 4.1 TIPS Algorithm

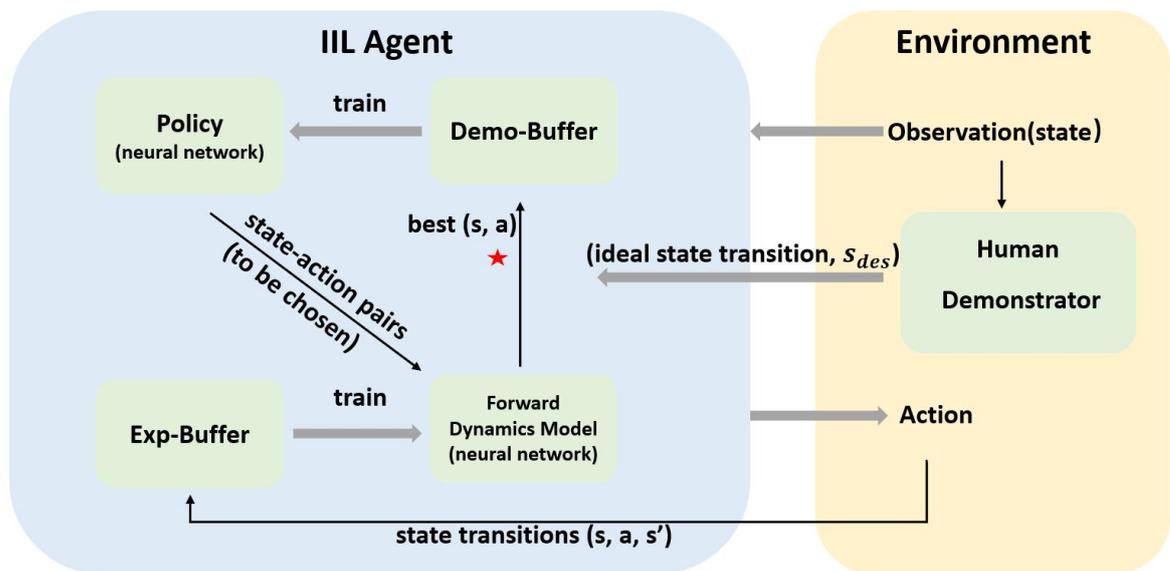

Fig. 4.1 TIPS model structure

The main difference between IIL algorithms based on state space and action space is that the feedback signal needs to be mapped from the form of the desired state to the action to be executed. Thus, the model structure of TIPS is also different: ① Forward dynamics model is learned to accomplish the mapping. ② Two different buffers are used. The Experience Buffer contains a set of state transitions $\{s_t, a_t, s_{t+1}\}$ for training the forward dynamics model. The Demo-Buffer contains state-action pairs $(s_t, a_t)$ for updating the policy. ③ Feedback mechanism also differs from the previous case; the whole interaction process is accomplished utilizing three functions under the hood, namely state correction function, action encoding function and internal cost function.

**State Correction Function**: receiving the correction signal from human teacher provided by the keyboard, encode the desired state through $s_{des} = s_t + h_t \cdot e$, where $e$ is error correction constant.

**Action Encoding Function**: mapping $s_{des}$ into action space. Under the hood, at every time step with human correction, the agent will generate a set of action candidates randomly and concatenate them with the current state $s_t$ to form a group of state-action pairs. Then, they will be fed into the forward dynamics model so that the state transitions will be predicted. The output states will be compared with $s_{des}$, the most similar one will be the chosen one, and the agent will execute the corresponding action.

**Internal Cost Function**: used as metrics for evaluating different action candidates, varying with different tasks.

Therefore, the entire TIPS model is mainly composed of: ①policy model, ②forward dynamics model, ③Exp-Buffer and Demo-Buffer, ③feedback mechanism.

## 4.2 Neural Network Structure

As mentioned earlier, in DCOACH, a fully connected neural network with two hidden layers is used to solve the control problem, where state data is accessible. Here, in order to compare with the DCOACH algorithm, both models of TIPS use the same network structure.

The policy model takes state data as input, outputs the action to be taken. Between layers, Leaky ReLU is used as activation function.

The number of neurons in each hidden layer varies according to different task scenarios. In the Cart-Pole scene, each hidden layer uses 16 neurons; in the Reacher and Lunar-Lander tasks, each layer uses 32 neurons.

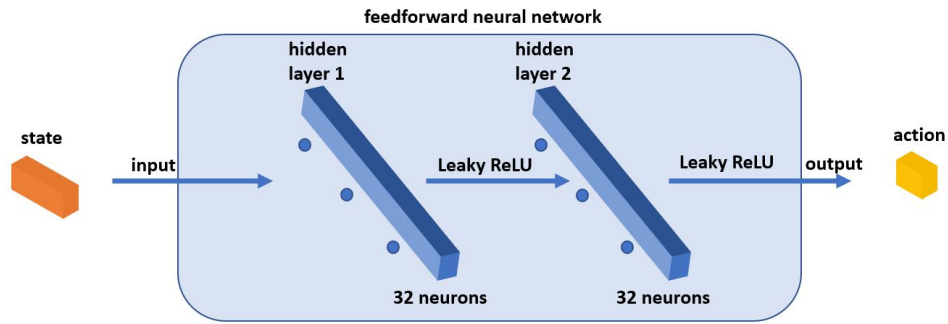

Fig. 4.2 Neural network of the policy

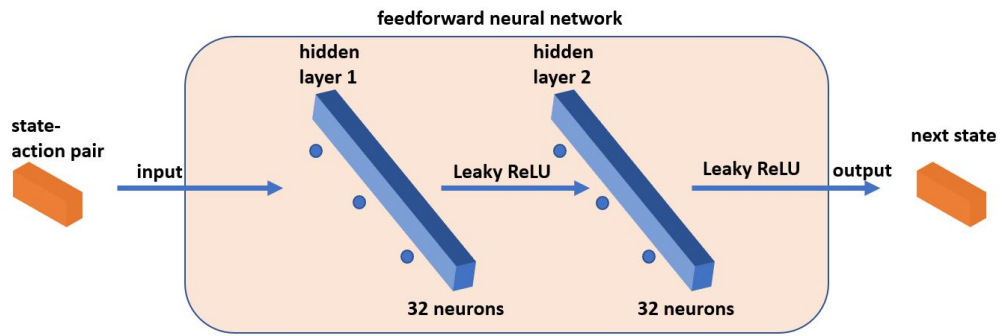

Fig. 4.3 Neural network of the forward dynamics model

The forward dynamics model is loaded with the previous state-action pair and predicts the next state. It has two hidden layers as well. Between layers, Leaky ReLU is utilized as activation function. In the task of Cart-Pole, forward dynamics model uses 16 neurons on each layer, while it has 64 neurons respectively in Reacher and Lunar-Lander scenarios.

### 4.3 Training Procedure

Table 4.1 TIPS Algorithm

| Algorithm 3 TIPS |
| --- |
| **Pre-training Stage** |
| 0: execute random policy $\pi_e$, collect data of state transition $\{s_t, a_t, s_{t+1}\}$ and add them into exp-buffer $\mathbb{E}$ |
| 1: train the forward dynamics model $i_\varphi$ |
| **Training Stage** |
| 1: **require**: error correction constant $e$, update interval of demo-buffer $b$, quantity of action candidates $ifdm\_queries$ in action encoding |
| 2: initialize: $\mathcal{D} = []$ # initialize the demo-buffer |
| 3: **for** $t = 1, 2, \dots$ **do** |
| 4:      **observe** state $s_t$ |
| 5:      **feedback** providing # corrective signal $h_t$ from human demonstrator |
| 6:      **if** $h_t$ **is not** 0 **then** |
| 7:          **state correction** $s_{des} = s_t + h_t \cdot e$ |
| 8:          **action encoding** # map $s_{des}$ to the action space $a_t^{des} = argmin_a \lVert f_\varphi(s_t, a_t) - s_{des} \rVert$ with forward dynamics model $f_\varphi$ |
| 9:          **append** state-action pair $(s_t, a_t^{des})$ to demo-buffer $\mathcal{D}$ |
| 10:         **immediate policy update** (using state-action pair $(s_t, a_t^{des})$ with SGD) |
| 11:        **batch policy update** (using mini-batch data from $\mathcal{D}$ with SGD) |
| 12:        **execute** action $a_t^{des}$ and reach the next state $s_{t+1}$ |
| 13:      **else** # $h_t = 0$ |
| 14:         **execute** $a_t = \pi_\theta(s_t)$ |
| 15:      **append** $(s_t, a_t, s_{t+1})$ to exp-buffer $\mathbb{E}$ |
| 16:      **if** $mod(t, b)$ **is** 0 **then** |
| 17:         **batch policy update** (using mini-batch data from $\mathcal{D}$ with SGD) |
| 18:      **if** episode terminates: |
| 19:         **update** forward dynamics model $f_\varphi$ (using mini-batch data from $\mathbb{E}$ with SGD) |
| 20:         **run** 9 testing episodes, average rewards as metrics |

In the case of TIPS, training procedure is divided into two parts: pre-training and training periods.

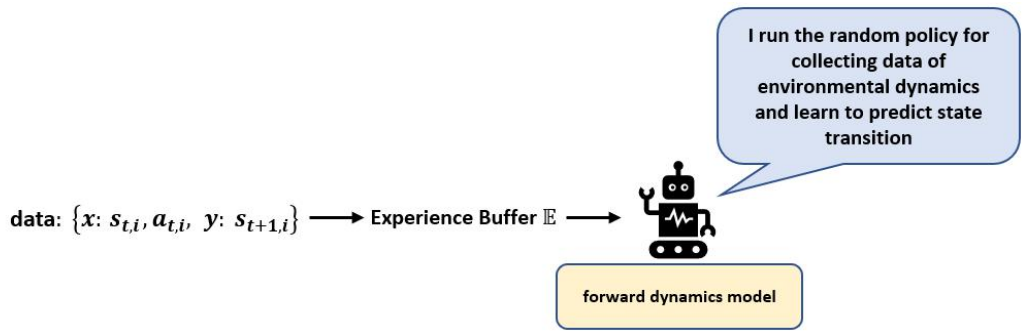

data: $\{x: s_{t,i}, a_{t,i}, \; y: s_{t+1,i}\}$ → Experience Buffer $\mathbb{E}$ →

I run the random policy for collecting data of environmental dynamics and learn to predict state transition

forward dynamics model

Fig. 4.4 Pre-training stage in TIPS

A. Pre-training Period

(1) In the pre-training period, the forward dynamics model is trained to learn the state transition of the environment. First, parameters of model $\varphi$ are initialized, a random policy will be executed on the agent, aiming at collecting multiple training data $\{s_t, a_t, s_{t+1}\}$. Then, use state-action pairs $(s_t, a_t)$ as input, $s_{t+1}$ as label to fit the model. The pre-training procedure will be walked through multiple times to reduce the cross-entropy loss.

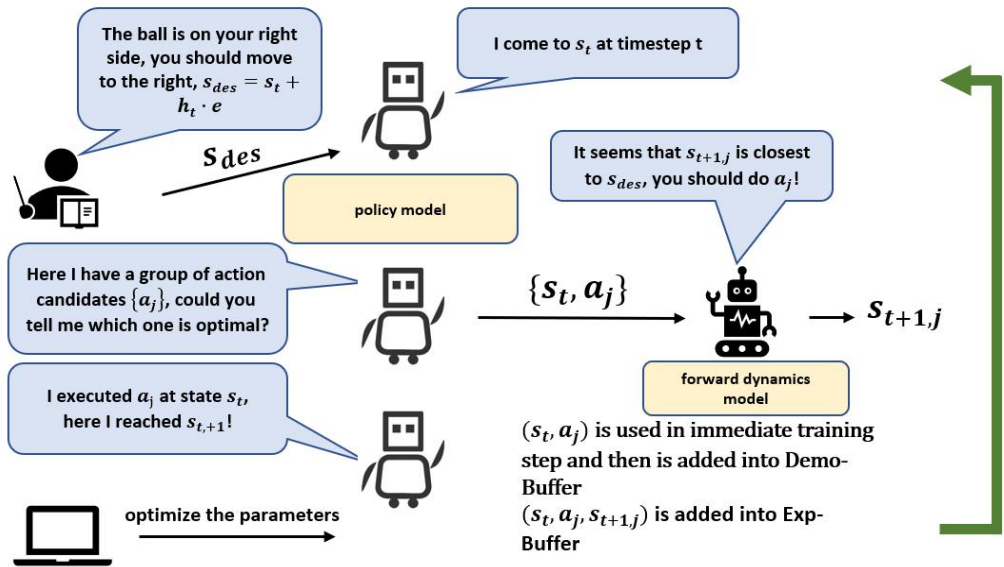

Fig. 4.5 Training stage in TIPS

### B.   Training Period

(1)   First, initialize the Domo-Buffer and generate a new instance of the agent. In the meantime, reset the simulation environment. Before every episode starts, the human demonstrator will be reminded.

(2)   Let the agent run its policy. Because the agent executes a random policy in the very first beginning, every episode will fail very fast.

(3)   At the subsequent timesteps, the human teacher observes the state of the agent and provides correction signals $h_t$ when necessary. If $h_t$ is not 0, it means that the current state should be corrected. The ideal state will be computed as  $s_{des} = s_t + h_t \cdot e$. That is, correction feedback will be added to the current state.
       The same as in the case of DCOACH, the correction signals from the human teacher are sent by a keyboard via pyglet library of Python. Note that it distinguishes between a long press and a short press. A counter named $h\_counter$ will record the count of timesteps with correction feedback.

(4)   When  $h_t$ is not 0, the desired state will be mapped back into action space as  $a_t^{des}$ via three internal functions, namely state correction function, action encoding function, and internal cost function.
       When  $h_t$  is 0, the desired action will be the output from the policy model.

(5)   Afterward, the agent executes the encoded action or the output action by the policy model, which depends on the human feedback.

(6)   After execution of an action, the policy is trained using encoded action and current state. Training steps include immediate training step using single state-action pair $\left(s_t, a_t^{des}\right)$  and mini-batch training step using a batch of state-action pairs sampled from the Demo-Buffer.
       The goal of taking both training step is still to make the training online and, at the same time, avoiding local overfitting of a certain trajectory. After training, new state-action pair$\left(s_t, a_t^{des}\right)$ will be loaded into the Demo-Buffer, new state transition $\{s_t, a_t, s_{t+1}\}$ will be set into the Exp-Buffer.

(7)   Every certain interval $b$, the policy will be updated again by mini-batch samples from Demo-Buffer.

(8)   After every episode, nine extra episodes would be conducted without a human teacher, aiming at evaluating the current policy performance fairly. The average reward from the nine episodes will be calculated as the metrics. Also, the ratio of timesteps with human feedback (feedback rate) will be given via $h\_counter$/ $t\_counter$. Besides, the forward dynamics model will be updated using samples from shuffled Exp-Buffer.

**4.4 TIPS Implementations in Different Task Scenarios**

The difference of TIPS algorithms in various scenarios mainly depends on the feedback mechanism, that is, the three internal functions. This section walks through the details of implementations of the three functions in the task scenes.

4.4.1 Cart-Pole-v0

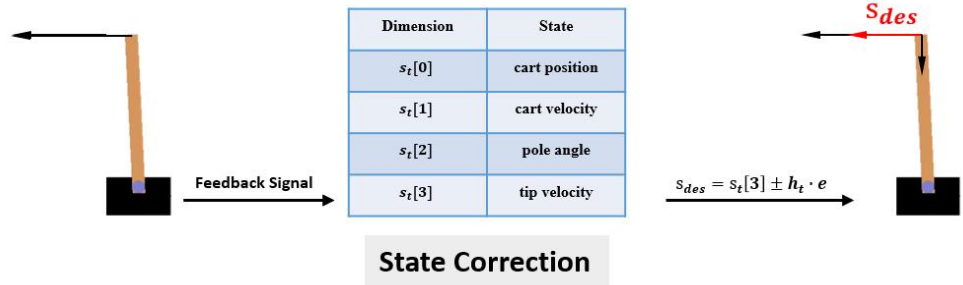

Fig. 4.6 State correction in Cart-Pole

State Correction

We define the sate correction function ($get\_state\_corrected(state, h_t)$), which takes the current state $s_t$ and the correction signal $h_t$ as input, returns the corrected state. Here we choose the linear velocity of the tip as the dimension that we want to exert feedback signal.

(1) First, initialize a zero array of length 2 using function $np.zeros()$, which is $s_{des}$.

(2) Decompose the linear velocity ($s_t[3]$) of the tip into a vertical and a horizontal part, which are stored in the zero array in step (1). The correction signal is added on the horizontal part of the linear velocity $velo_x = s_t[3] \cdot cos\theta$.

(3) If the human teacher presses the 'left' arrow key, the horizontal dimension of $s_{des}$ subtracts a correction error $s_{des}[0] = velo_x - e$, while the vertical part remains the same. Here, values of $h_t$ are different integers varying with different feedback signals.

(4) Return the encoded state $s_{des}$.

```
'''
state correction function
'''
def get_state_corrected(self, h_fb, state):

        # derive the x, y component of tip velocity and correct x component
        state_corrected = np.zeros(2)            # initialize zero array
        angle = (state[2]*np.pi)/180             # pole angle in rad
        state_corrected[0] = state[3]*np.sin(angle)  # x component
        state_corrected[1] = state[3]np.cos(angle)   # y component

        if (h_fb == H_LEFT):                     # state correction
                state_corrected[0] -= self.errorConst
        elif (h_fb == H_RIGHT):
                state_corrected[0] += self.errorConst

        return state_corrected                   # return the desired state
```

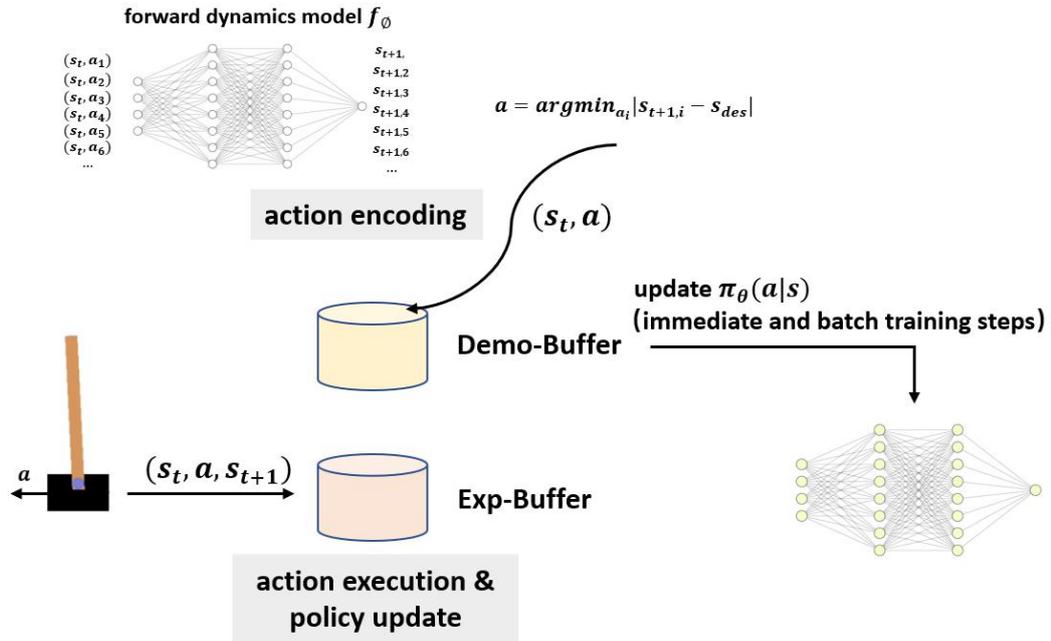

Fig. 4.7 Action encoding and policy update in Cart-Pole

Action Encoding

We define action encoding function ($get\_corrected\_action(h_{fb}, s_{des}, s_t)$, which takes expected state $s_{des}$ (one−dimensional array of length 2) from state correction function, correction signal $h_t$ from human demonstrator and current state $s_t$ (a one-dimensional array of length 4) as input, maps them to action space.

  (1) First, copy the current state array $s_t$ to a batch $ifdm\_queries$ (the size of this batch is a hyperparameter determined by the task, more complex task requires a larger value) and assign it to the variable $States$ (array in the shape of ($ifdm\_queries \times 1 \times 4$)).

  (2) Use the $np.eye()$ function to initialize an identity matrix of shape ($2 \times 2$) as the action space data, which is in the form of one-hot encoding. The first line $(1, 0)$ represents moving the agent to the left, and the second $(0, 1)$ represents moving the agent to the right.

  (3) Use the $np.random.choice()$ function to randomly generate an array $Actions$ in the shape of ($ifdm\_queries \times 2$), which represents a set of random actions, each row is either $(0., 1.)$ or $(1., 0.)$.

  (4) Define $eval\_fdm(State, Actions)$ function, which passes the two arrays in (1) (3) as input to the forward dynamics model, predicts the state transition. An array in the shape of ($ifdm\_queries \times 4$) will be returned, where each row represents an action candidate to be chosen and the projected state transition.

  (5) For each row of the $NStates$ variable, use the cosine function to take the horizontal part of the linear velocity of the tip, an array of length 10 is generated as $NStates\_xpos$.

  (6) Define the internal cost function, which is an array of length 10, each digit is the absolute value of $NStates\_xpos$ subtracted by $s_{des}$. The most desired action candidate has the least cost.

  (7) Use $np.argmin()$ function to find the index of the chosen action candidate with the least cost, which is named as $min\_a$.

  (8) Return $min\_a$.

In step (4), the random action array and the current state are passed to the forward dynamics model to predict the state transition. Because sampling cost is very small in the simulation environment, and some have known state transition functions. In this case, the new state can be obtained directly through the state transition function without using the forward dynamics model. This is a non-bias estimation. However, we still employ the forward dynamics model to simulate the agent working in complex, unpredictable, real scenarios.

After the above two functions, the state-space feedback signal provided by the human instructor is encoded into action. Combined with the current state, the state-action pair is used for later training steps, as mentioned before.

### 4.4.2 Reacher-v2

In the DCOACH algorithm, feedback signals are used to control the torque of different joints of the robotic arm, which encounters difficulties in demonstration. Non-expert human demonstrators may find it challenging and not intuitive to provide feedback, especially when coordinating the motion of the two sections for reaching the object. In contrast, in state space, one can only focus on the Cartesian coordinates of the end effector and the goal.

Reacher is a typical scenario that can distinguish between two algorithms. As mentioned earlier, the state space of the agent in this environment is represented by a one-dimensional array with a length of 11. The action space is a one-dimensional array with length of 2. Both digits are continuous on [-1, 1], representing the control input exerted on the two joints (counterclockwise rotation is positive, clockwise rotation is negative).

The code is as follows: the target object's information is removed first in the pre-training stage. The state transitions are collected through a random policy.

```python
# remove the target-related information in state space
z_index = [4,5,8,9,10]
prev_s [z_index] = 0
state[z_index] = 0

States.append(prev_s)          # collect s_t
Nstates.append(state)          # collect s_t+1
Actions.append(A)              # collect action executed at s_t
```

In completing the reaching task, the algorithm needs to control the Cartesian location of the end effector. Thus, a function $get\_arm\_pos(state)$ needs to be defined for getting the Cartesian coordinates of the end effector from current state data.

```python
'''
function for getting the coordinates of the end effector of the robotic arm
'''
def get_arm_pos(self, state):
    ang1 = np.arctan2(state[:,2], state[:,0])     # get first joint angle
    ang2 = np.arctan2(state[:,3], state[:,1])     # get second joint angle
    arm1 = 0.1                                     # length of first section
    arm2 = 0.11                                    # length of second section

    xpos = arm1*state[:,0] + arm2*np.cos(ang1+ang2)  # x component
    ypos = arm1*state[:,2] + arm2*np.sin(ang1+ang2)  # y component

    return xpos, ypos
```

State Correction

(1) Read human feedback $h_t$. When the correction signal is not 0, invoke the state correction function $get\_state\_corrected(h_t, state)$.

(2) Use the function $np.zeros(2)$ and initialize a one-dimensional array of length 2 named as $state\_corrected$ for storing the corrected Cartesian coordinates of the end effector later.

(3) Invoking the function $get\_arm\_pos(state)$ to get the current Cartesian location of the end effector, returned as variables $state\_x, state\_y$.

(4) Compute the desired state $s_{des}$. If the feedback signal from human is 'left', then, $state\_corrected[0] = state\_x - e$, $state\_corrected[1] = state\_y$, similar in other cases.
(5) Return the location of the end effector $state\_corrected$.

```
'''
Action encoding in Reacher scenario
'''
# feed states-actions array into forward dynamics model to predict transition
Nstates = self.get_transition(States, Actions)
# derive the end effector coordinates of the predicted states
Nstates_x, Nstates_y = self.get_arm_pos(Nstates)
# calculate the difference between predicted states and desired state
cost = abs(state_corrected[0] - Nstates_x) + abs(state_corrected[1] - Nstates_y)
# index the optimal action
min_cost_index = cost.argmin(axis=0)
min_action = Actions[min_cost_index]
```

Action Encoding

(1) First, remove the location information of the object in input state data.
(2) Use $np.tile(state, (ifdm\_queries, 1))$ to copy the processed state data into array $States$ in the shape of $(ifdm\_queries, 1)$;
(3) Use $np.random.uniform()$ function to randomly generate an array $Actions$, which will be used for storing a set of candidate actions.
(4) Pass arrays $States$ and $Actions$ to the forward dynamics model, get a set of projected new states $Nstates$.
(5) Use the $get\_arm\_pos()$ function ($Nstates$ as input) to get a set of x and y coordinates named as $Nstates\_x, Nstates\_y$.
(6) Define the internal cost function, which is the sum of the difference between the candidate states $Nstates\_x, Nstates\_y$ and the desired state $s_{des}$ in the form of absolute value.
(7) Use $cost.argmin()$ to select index of the ideal action, $min_{action} = Actions[min\_cost\_index]$.

### 4.4.3 Lunar-Lander-v2

In Lunar-Lander scenario, the agent needs to control itself to smoothly land to a specified location (coordinate origin) by manipulating three engines. Each timestep with firing the main engine, the agent will receive a corresponding negative reward (penalty). Thus, the agent should control both the landing position and posture in a short time horizon with the main engine fired.

Lunar-Lander has two versions, discrete action space $(4,)$ and continuous action space $(2,)$. In the discrete action space scenario, at each timestep, the agent can only take one action (that is, turn on a certain engine or extinguish all engines). Nevertheless, in the continuous case, the agent can turn on two engines simultaneously, which could accomplish combined actions.

### A.  Discrete Action Space

In Lunar-Lander scenario, because the dimension of state space higher, we increase the size of the Exp-Buffer for training the forward dynamics model. As for the choice of the dimension, where feedback signals are given, we choose the vertical velocity and the angular velocity of the spaceship. The consideration is that on the real system, the localization data always have delay. Thus, the velocity data may be a better choice in highly dynamic task scenarios.

The human instructor provides feedback through the four arrow keys on the keyboard (corresponding to different values of the variable $h_t$). The 'up' key represents the vertical velocity of the spaceship should increase. The 'left' means the angular velocity of the agent should increase counterclockwise, while 'right' representative angular velocity should increase clockwise.

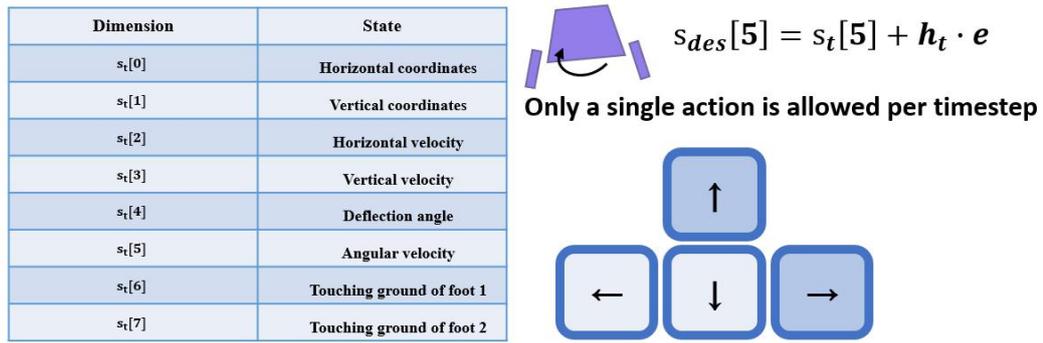

Fig. 4.8 State correction in Lunar-Lander with discrete action space

## State Correction

With the feedback signal $h_t$ and current state $s_t$ (a one-dimensional array of length 8) provided by the human demonstrator as input, the state correction function outputs the ideal $s_{des}$, which is corrected on the fourth and sixth digits.

(1) Use the function $np.copy()$ to pass the current state to the desired state $s_{des}$.

(2) Detect the feedback signal. If it is 'left', add the sixth bit of the $s_{des}$ array (angular velocity) by an error correction constant $e$, and vice versa. If the feedback is 'up', increase the fourth bit of the $s_{des}$ array (vertical velocity) by an error correction constant $e$, and vice versa.

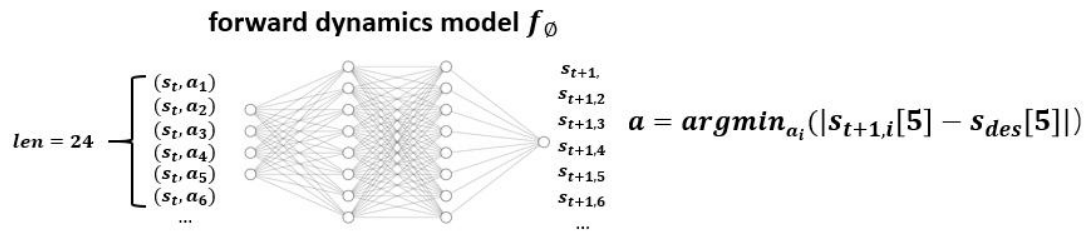

Fig. 4.9 Action encoding in Lunar-Lander with discrete action space

## Action Encoding

The action encoding function takes the correction signal $h_t$ and current state $s_t$ as input, outputs the optimal action $min\_a$, which means the action candidate with minimal cost.

(1) If the feedback signal prompts that no action should be taken, the optimal action $min\_a$ is the array $[1, 0, 0, 0]$, which means extinguishing all engines.

(2) If the feedback signal prompts to take action, copy the current state and generate a two-dimensional array $States$ in the shape of $(24, 8)$.

(3) Use the $np.eye()$ function to generate a 4-by-4 identity matrix $a$, which represents four discrete actions in the form of one-hot encoding. Then, use the $np.random.choice()$ function to randomly select actions from identity matrix $a$ and generate a two-dimensional array $Actions$ of $(24, 4)$.

(4) Pass the array $States$, $Actions$ as input to the forward dynamics model and obtain a two-dimensional array in the shape of $(24, 8)$, which represents the new states $Nstates$.

(5) Define the internal cost function. If the feedback signal is added to the vertical velocity, use the absolute value of the fourth bit of $Nstates$ subtracted by the desired state $s_{des}$ as the cost. If the feedback signal is added to the angular velocity, use the absolute value of the sixth position of $Nstates$ subtracted by the desired state $s_{des}$.

(6) Use the $argmin()$ function to select the optimal action from $Actions$ and return it as $min\_a$.

B.   Continuous Action Space

Feedback Mechanism

In the continuous version of Lunar-Lander, the agent is allowed to take multiple actions simultaneously, such as turning on the main engine and the left engine to push the spaceship aslant. The representation of the action space is no longer a one-dimensional binary array of length 4, but an array of length 2 with two digits continuous on $[-1,1]$. The first number controls the main engine. On the interval [-1, 0], the main engine is turned off. On the interval (0, 1], the main engine is fired (50%-100%, by default, power under 50% is invalid). The second digit controls the two sub-engines. On the interval [-1, -0.5) and (0.5, 1], the left and right engines are fired, while both sub-engines are turned off on the rest part of the interval.

Therefore, the feedback signal is different from the discrete case. New feedback signals representing the combined actions should be defined.

As the state transition of the agent in the environment becomes more complicated, in the exploration stage, the number of samples for training the forward dynamics model is increased to 20000, and the hyperparameter $ifdm\_queries$ is increased to 500, that is, the number of candidate actions at each timestep increase to 500.

For selecting the feedback signal, in the Lunar-Lander scenario, the agent must simultaneously control the horizontal coordinate position of landing (the landing platform is on the center of the screen) and its landing posture. Therefore, as the feedback signal, the horizontal coordinate, the vertical coordinate, the vertical velocity, the angle of the agent, and its angular velocity are most relevant to the task.

At the same time, in the continuous case, since the agent can take multiple actions simultaneously, the feedback signals can be combined. In the case that horizontal coordinate, angle and vertical coordinate are selected as feedback signal, if the spaceship is required to move towards up and right, the correction should be increasing the desired vertical coordinate and horizontal coordinate and decreasing the angle of the agent (clockwise corresponds to negative value). By contrast, if the agent's vertical velocity and angular velocity are used as the feedback signal, when the agent is expected to move to the upper right, only the angular velocity and the vertical velocity need to be increased, which simplifies the demonstration. Thus, we will choose the vertical velocity and the angular velocity of the agent as the dimensions where corrections are given.

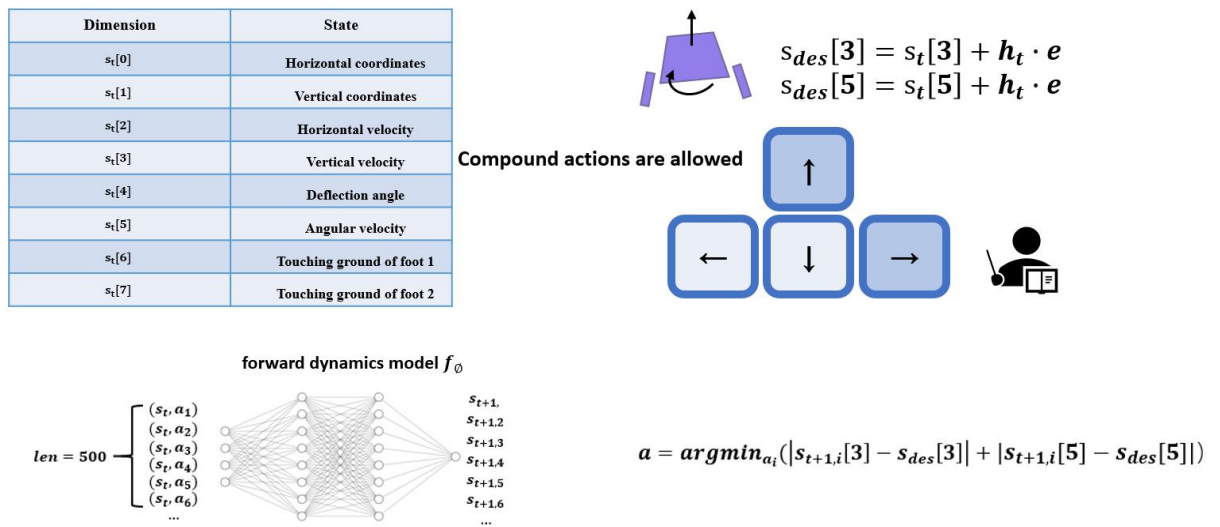

| Dimension | State |
|---|---|
| $s_t[0]$ | Horizontal coordinates |
| $s_t[1]$ | Vertical coordinates |
| $s_t[2]$ | Horizontal velocity |
| $s_t[3]$ | Vertical velocity |
| $s_t[4]$ | Deflection angle |
| $s_t[5]$ | Angular velocity |
| $s_t[6]$ | Touching ground of foot 1 |
| $s_t[7]$ | Touching ground of foot 2 |

$$s_{des}[3] = s_t[3] + h_t \cdot e$$
$$s_{des}[5] = s_t[5] + h_t \cdot e$$

Compound actions are allowed

forward dynamics model $f_\emptyset$

$$len = 500 \begin{cases} (s_t, a_1) \\ (s_t, a_2) \\ (s_t, a_3) \\ (s_t, a_4) \\ (s_t, a_5) \\ (s_t, a_6) \\ ... \end{cases} \quad \begin{array}{l} s_{t+1,1} \\ s_{t+1,2} \\ s_{t+1,3} \\ s_{t+1,4} \\ s_{t+1,5} \\ s_{t+1,6} \\ ... \end{array}$$

$$a = argmin_{a_i}(|s_{t+1,i}[3] - s_{des}[3]| + |s_{t+1,i}[5] - s_{des}[5]|)$$

Fig. 4.10 State correction and action encoding in Lunar-Lander with continuous action space

State Correction

(1)   When the feedback signal provided by the human demonstrator is simply to the left or right ($h_t = 3, 4$), the state correction function sets the fourth digit of the desired state $s_{des}$ (vertical velocity) to zero, and adds the sixth digit (angular velocity) by an error correction constant *e*.

(2)  Similarly, When the correction signal is simply to the up or down ($h_t = 1, 2$), the state correction function sets the sixth bit of the desired state $s_{des}$ (angular velocity) to zero, and adds the fourth digit (vertical velocity) by an error correction constant *e*.

(3)  If the feedback signal is compound, e.g., it suggests the desired state should be towards left and up, then both the fourth and the sixth digits of $s_{des}$ are added by an error correction constant *e*.

(4)  Return the corrected state data $s_{des}$.

As for the action encoding function, we could simply change the *Actions* array's length to 2 and increase the hyperparameter $ifdm\_queries$ to 500, which means more action candidates are employed at each timestep.

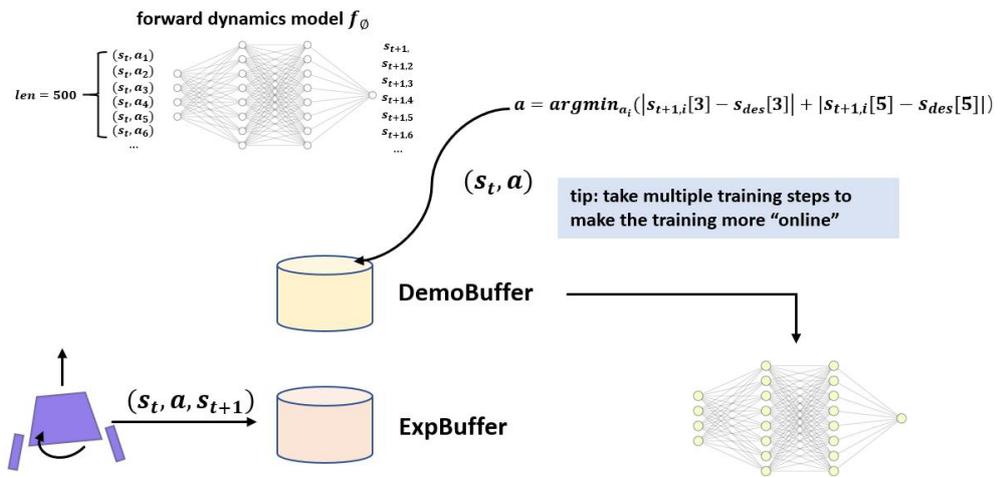

Fig. 4.11 Policy update after action encoding

## 4.5 Summary

At this point, this thesis has completed the introduction at the algorithm level. It can be seen that in terms of model structure, providing feedback based on the state space requires a more complex model. The next chapter will conduct confirmatory experiments on the TIPS algorithm and the DCOACH algorithm and compare them with other RL algorithms.

# 5. Experimental Results

In chapter 3 and chapter 4, we have introduced the details of the two IIL algorithms. This chapter focuses on validating and comparing the two algorithms and giving further analysis. In the meantime, some problems encountered in the training process are further discussed and solved.

## 5.1 Hyper-parameters

Hyperparameters in the experiments are listed below, including the learning-process-related parameters, such as learning rate, batch size, etc. Besides, for various task scenarios, different buffer sizes, sizes of candidate actions (in TIPS) and sample quantities for learning transition models are chosen as follows. Note that another essential hyperparameter is error correction const $e$, which will be further discussed in respective scenarios.

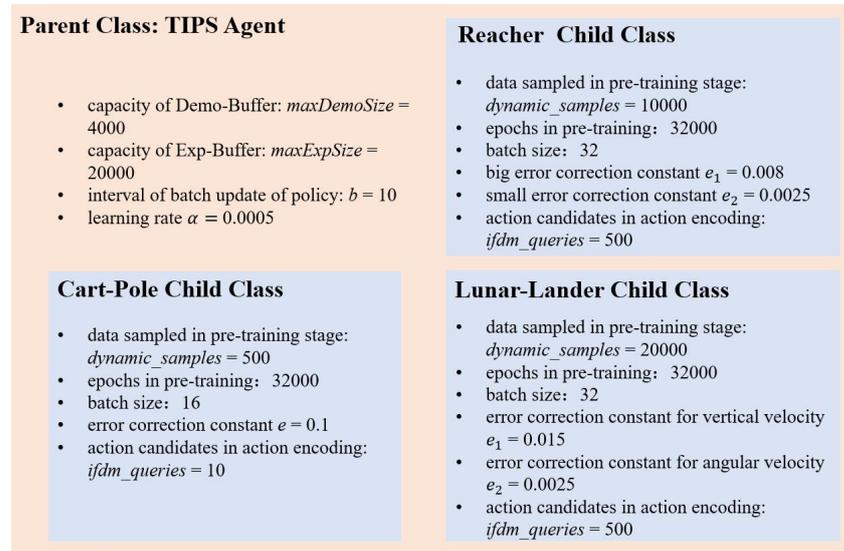

Fig. 5.1 Hyper-parameters in TIPS

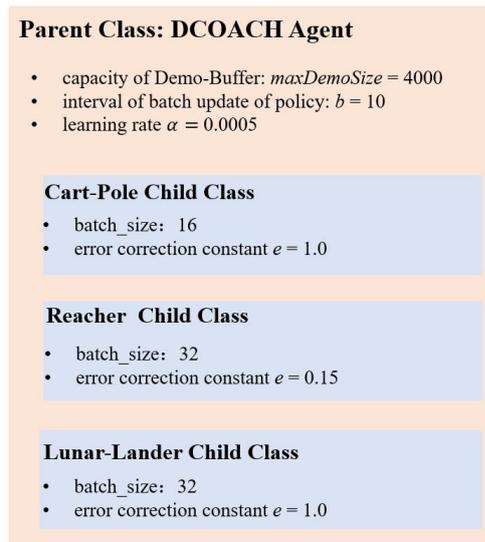

Fig. 5.2 Hyper-parameters in DCOACH

**5.2 Cart-Pole**

5.2.1 Baseline Algorithms from OpanAI Gym

As a classic reinforcement learning scenario, a few baseline algorithms are available for benchmarking in Cart-Pole. The Following figures visualize learning processes of using several model-free RL algorithms to solve the Cart-Pole problem. For each algorithm, six experiments are conducted.

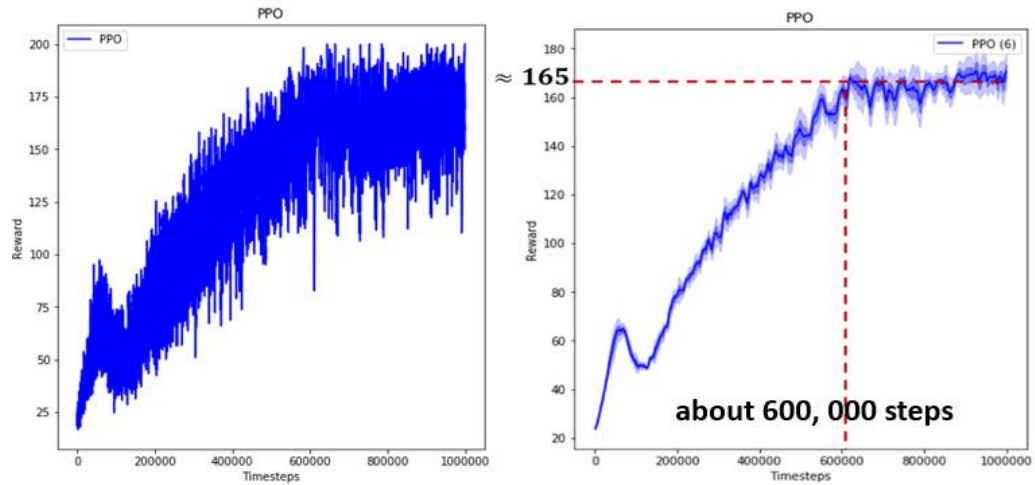

Fig. 5.3 PPO algorithm in Cart-Pole

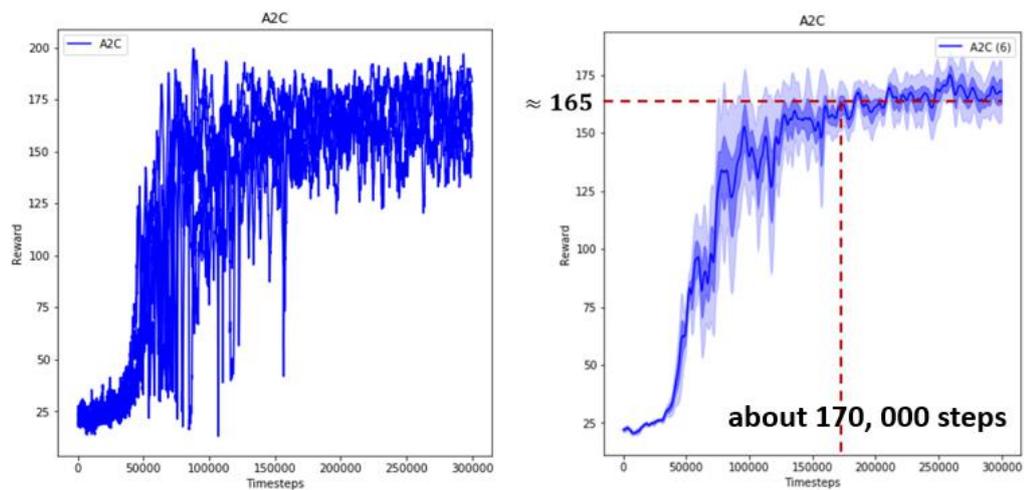

Fig. 5.4 A2C algorithm in Cart-Pole

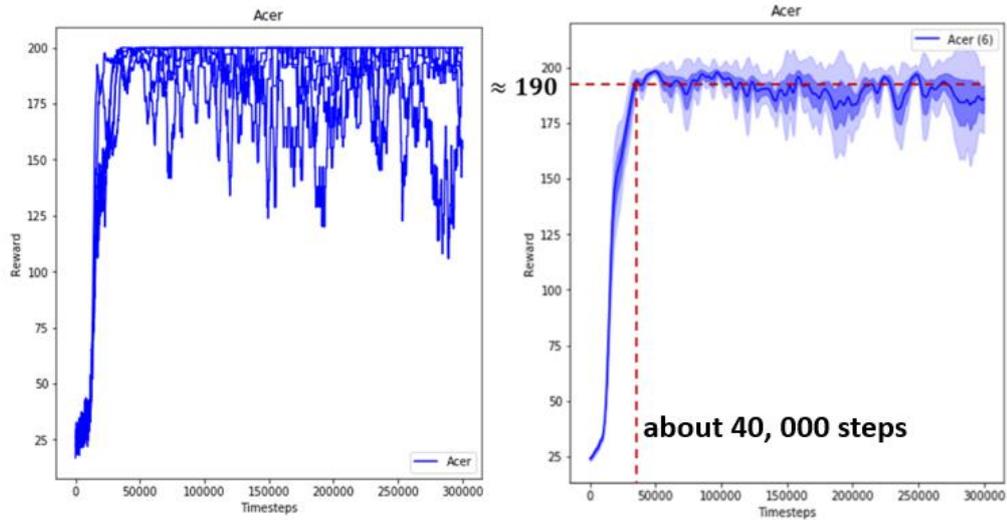

Fig. 5.5 Acer algorithm in Cart-Pole

The left side of each picture is visualized with raw data and has not been smoothed while the right is the smoothed curve. The darkest solid line in the middle is the average of 6 experiments. As can be seen from the figures,

(1) Among the three algorithms, Acer requires the least training iterations for acquiring an initial policy, which can approximately solve the problem (about 40,000 timesteps). Training procedures for the remaining two algorithms take above 170,000 and 600,000 timesteps, respectively. The equivalent episodes for Acer are above 200, even if the maximum timesteps for each iteration are assumed (the actual number of iterations tends to be much higher).

(2) From the perspective of stability, it can be seen that on the left side, learning curves are highly noisy. Even after the initial policy is learned, the quantity of obtained rewards for each iteration tends to fluctuate intensely.

(3) Judging from the rewards that can be obtained in the latter period of training, all the algorithms can train a qualified policy for solving the Cart-Pole problem.

In the following sections, learning through interaction will be validated and compared with standard RL algorithms.

5.2.2 TIPS in Cart-Pole

A. Training Process: Good and Fast!

For the training section in Cart-Pole-v0, a total 6 experiments are conducted, 50 episodes for each.

As described in Chapter 4, 9 rounds of tests are implemented after each episode with human corrections, aiming at fairly showing the current performance of the agent. The policy is not updated in the nine rounds. Thus, the agent can only perform well under the supervision of humans, can be avoided. And the experimental results are more reliable.

The following are the learning processes of each training procedure. Average rewards and feedback rate are used as metrics.

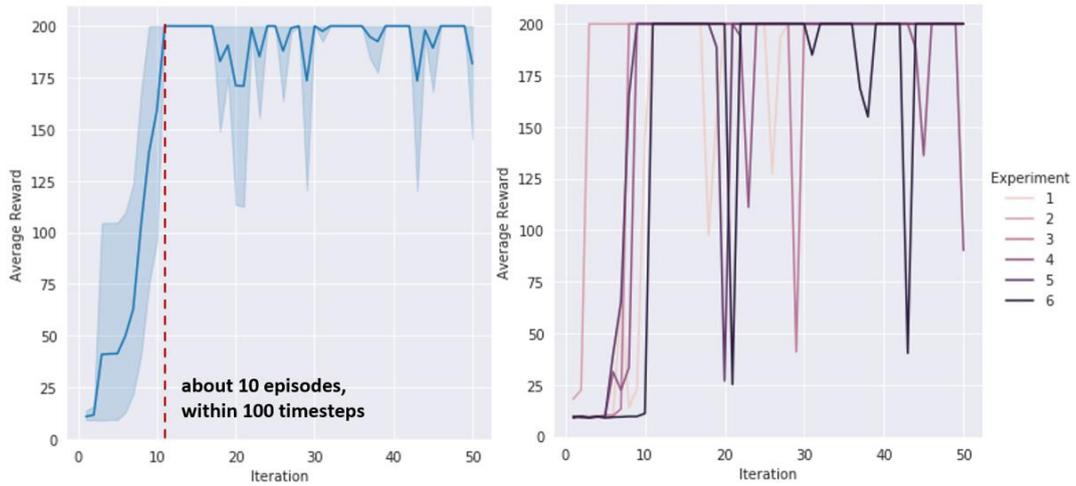

Fig. 5.6 Learning curves of TIPS in Cart-Pole

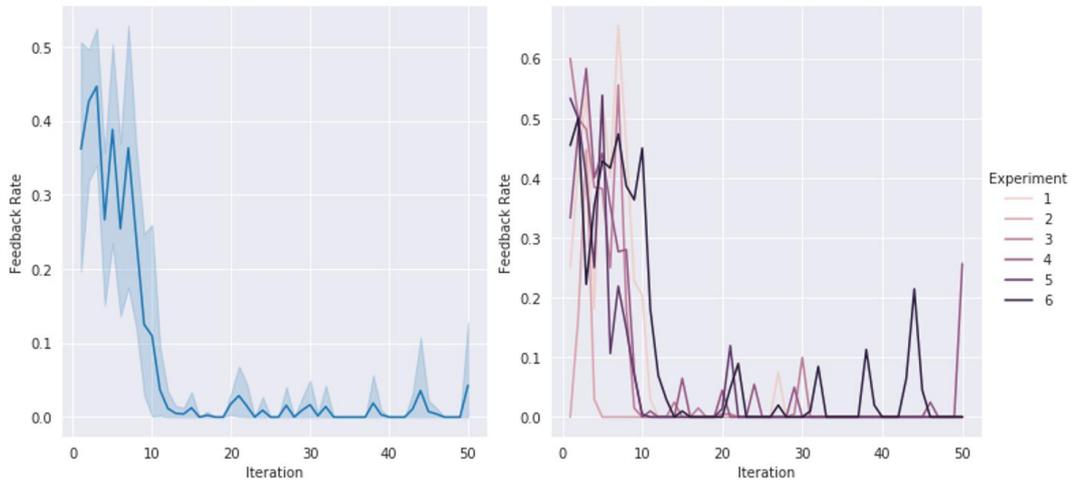

Fig. 5.7 Curves of feedback rate of TIPS in Cart-Pole

Analyzing the above figures, it is not difficult to see:

(1) In 6 experiments, the agent using the TIPS algorithm quickly learned a policy to solve the Cart-Pole problem in about 10 iterations under the guidance of a human demonstrator. It can be concluded that for the control problem in lower-dimensional space, human providing feedback signals can **significantly reduce the required training iterations** (equivalently, about 100 timesteps, compared with for Acer algorithm, at least 40,000 timesteps are required);

(2) The right-sided figures show that, after acquiring the initial policy, agents using TIPS algorithm can **stabilize at a high-level performance**, except that in some individual rounds, the obtained rewards may fall again dramatically and suddenly.

(3) It is worth noting that, in the six experiments, even after learning a good policy, **there are more or less sudden degradations of performance in each experiment** (typically 1-2 iterations per experiment, which will be further explored and discussed in the following sections). If the correction signal is not given, the performance will continue to deteriorate because the impact of the training on the agent's policy is online. That is, after each timestep, the executed action will be used as training data to update the agent's policy immediately. Also, if the executed actions are incorrect, accumulated biased data in the Demo-Buffer will further break the policy. Therefore, **the training process should be entirely under supervision from humans**, which is the price of a significantly shorter training process of learning through interaction.

The conclusion of this section: Using TIPS algorithm can **significantly reduce the required training episodes**, compared with regular model-free RL algorithms. Also, **the performance of acquired policy in learning through interaction is better than in RL scenarios.** The following few sections are arranged as follows: the automatic problem-solving process without human feedback is given in the test section, chapter $5.3.2.B$. In chapter $5.3.4$, experiments and analysis towards the sudden degradation phenomenon are further carried out.

### B．Testing Process

In the previous section, few agents are trained using TIPS, which is proved to outperform the regular model-free RL algorithm in both speed of convergence and stability aspects. This section uses the aforementioned agents in testing scenarios without further human demonstration and policy update. That is, the difference between episodes is just initial states.

Because the training process is online, if the agent's performance is unstable during the training process, the human demonstrator must intervene. That is why the training and testing are carried out separately for learning through interaction. It is also a characteristic of this kind of algorithm and should be further thought about for selecting algorithms.

Each agent trained in $5.3.2.A$. is tested with a large number of episodes. Each test has a total of 3000 rounds (about 500,000 timesteps). The following are the test results.

Result: In the test for six agents, five agents can completely solve the Cart-Pole problem. That is, the rewards obtained in each round are stable at 200, and there is no instability.

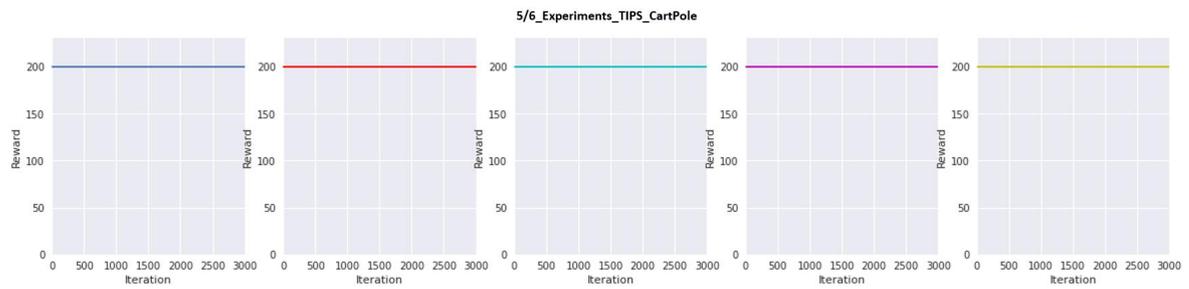

Fig. 5.8 five tests with full rewards

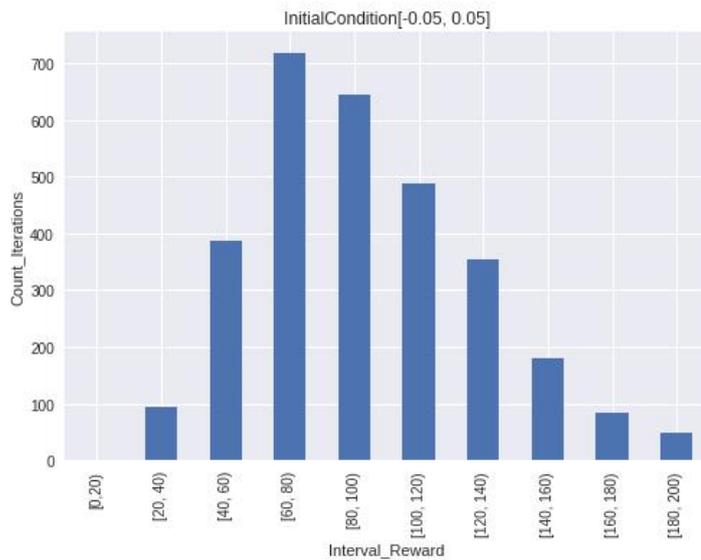

Fig. 5.9 Target agent under the original initial condition

An agent did not completely solve the Cart-Pole problem in the test. The rewards obtained in 3000 rounds were plotted in different reward intervals as above. It can be seen that the rewards it receives are mainly in the range of 60-140. We select this agent as the experimental object, analyze the policy it learned and further explore the reason for the sudden degradation phenomenon in the training process. Details are showed in section 5.3.4. The following section will train DCOACH agents to solve the Cart-Pole problem.

### 5.2.3 DCOACH in Cart-Pole

#### A．Training Process

Training and testing procedures for DOCAH agents are separate and exactly the same as in the TIPS scenarios. The primary metrics are rewards and feedback rate. Fifty episodes are conducted for each of the six experiments in this section. After each episode, independent nine autonomous runs show the average performance of the agent.

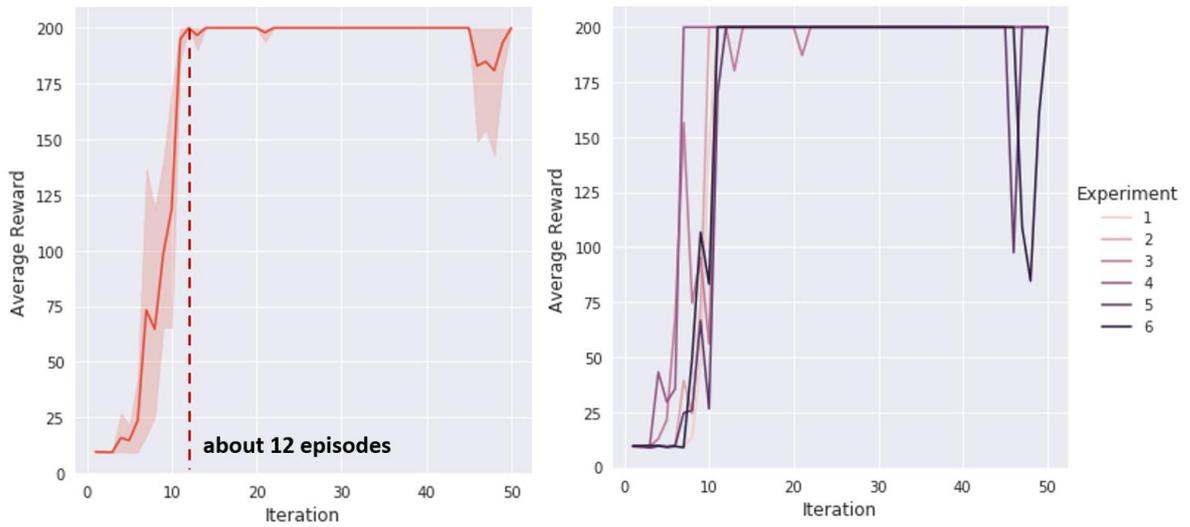

Fig. 5.10 Learning curves of DCOACH in Cart-Pole

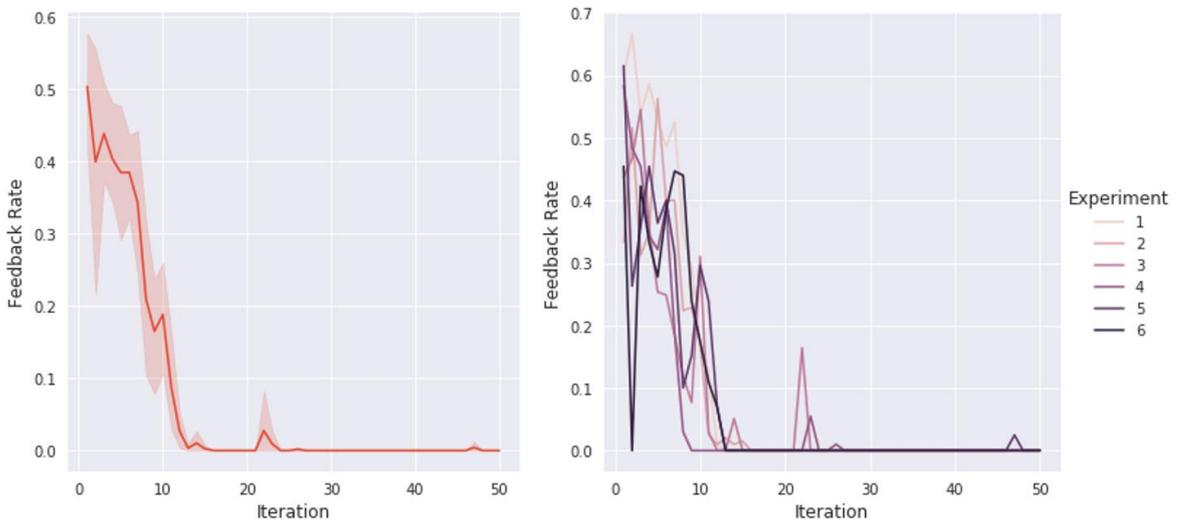

Fig. 5.11 Curves of feedback rate of DCOACH in Cart-Pole

The following conclusions can be drawn from the above results:
(1) From the perspective of required episodes for learning an initial policy, DCOACH algorithm has a similar performance as TIPS agent. Both of them could significantly outperform the regular RL algorithm.

(2)  As for the phenomenon of sudden performance degradation, **DCOACH agents show higher stability during training when compared with TIPS agents.**

B．Testing Process

This section uses the same experimental design as for the TIPS agent to test the six agents trained in the previous chapter and briefly analyzes the results. A more detailed summary and analysis are in 5.3.6 Given in.

In six 3000-rounded test experiments, five DCOACH agents completely solved the Cart-Pole problem, the number of rewards was stable at 200. Similarly, the following bar chart shows the unique agent's performance in different reward intervals.

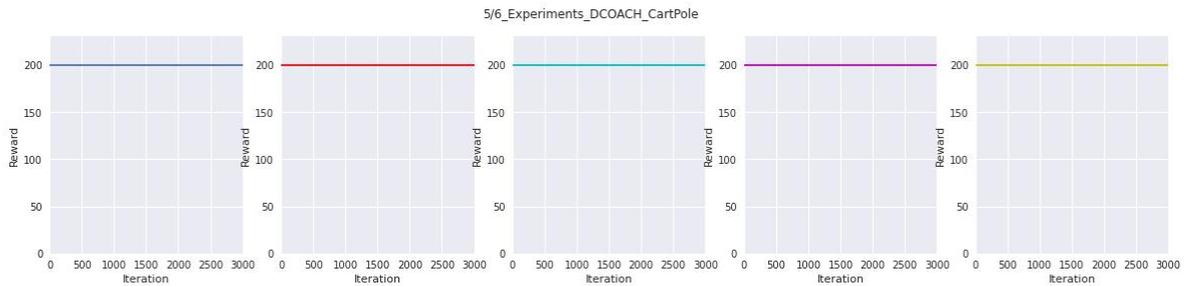

Fig. 5.12 Five tests with full rewards

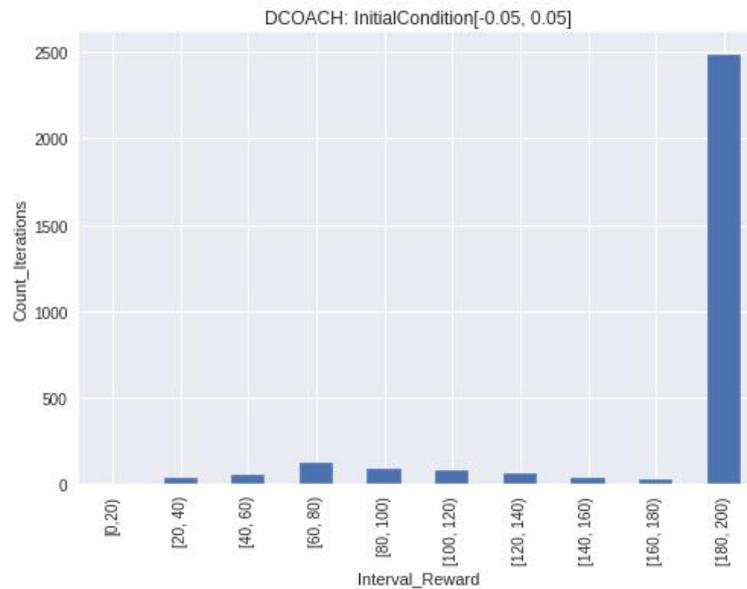

Fig. 5.13 Target DCOACH agent under the original initial condition

Based on the above results, some conclusions can be made: both the TIPS agent and the DCOACH agent can solve the Cart-Pole problem well and learn the initial policy quickly. However, in the training phase, **DCOACH algorithm converges better than TIPS algorithm, and the number of unstable phenomena is less.**

As for the test results, 5/6 of the two kinds of agents have completely solved the Cart-Pole problem after training (continuously receiving 200 rewards without any instability). Note that this is a very harsh criterion for the reinforcement learning algorithms because without receiving knowledge from humans, the RL agents still need to do some random exploration even after learning a reasonable policy, where episodes may drop dramatically.

For the two agents that have not completely solved the Cart-Pole problem, we draw their performance regarding different intervals of rewards in 3000 rounds of tests. Obviously, the DCOACH agent looks better. Nevertheless, the different training processes can be occasional. It cannot be said that DCOACH algorithm outperforms TIPS in general.

Regarding the instability of the two algorithms, more exploration and analysis are given in section 5.3.4, including hypothesis-driven experiments and testing. The overall conclusion of the two algorithms in Cart-Pole scenarios is drown in section 5.3.6.

### 5.2.4 Exploration and Analysis of Instability Phenomenon

In the previous sections, two interactive imitation learning agents were trained and tested. Both of them solved the control problem in the Cart-Pole scenario very well, but they also showed some sudden degradation phenomena in performance, which is worth discussing.

Although the existing results are already outstanding, to optimize to model, this chapter will use the worst agent in the test experiment of the TIPS algorithm, carry out analysis and solve the instability problems that arise.

### A. Exploration for Reason of Sudden Instability

First, review the training curve of the TIPS agent, which could not completely solve the problem in the testing experiment.

It can be seen that the agent was unstable in the 22nd, 45th, and 50th episodes. **The hypothesis can be made that the final policy is not qualified because of the instability in the last training episode, which 'broke' the learned policy and was not corrected on time.**

As mentioned above, if there is an unstable phenomenon in the training process and the human demonstrator does not provide correct feedback at once, biased state-action pairs will be used for policy updating in the immediate training steps and reversely 'break' the existing policy. Furthermore, accumulated biased training data also causes long-term degradation of the policy. Which means, in the training process of interactive imitation learning agents, human should always be concentrated and ready to give feedback. The following are hypotheses towards the instability phenomenon.

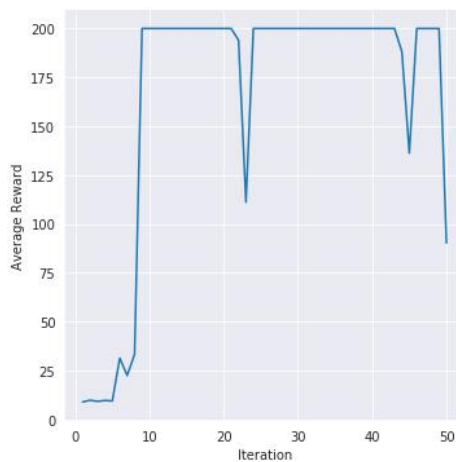

Fig 5.14 Recalling the training process of the "imperfect" agent

**Hypothesis 1：occasional extreme initial states cause the sudden instability**

Since the instability occurs suddenly and the only difference between training rounds is the initial states of the environment, we guess that maybe the initial conditions of the environment in some episodes are relatively extreme, which causes the agent to be unable and perform poorly.

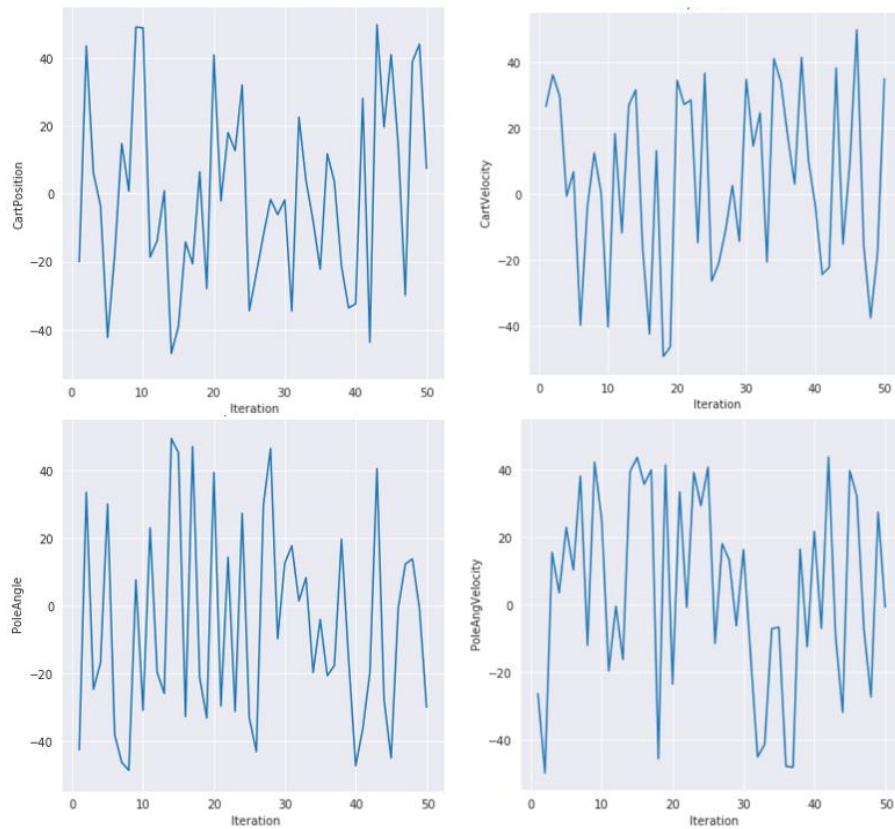

Fig. 5.15 Change of initial states in the training process

First, we draw the change of the initial conditions during the training process individually (cart position, cart speed, pole angle, and angular velocity). As a result, it cannot be seen from the figure that the initial conditions are more extreme in the rounds where the policy deteriorated.

Similarly, we can plot the change of initial states in the nine test experiments after each training episode, as shown in the figure below.

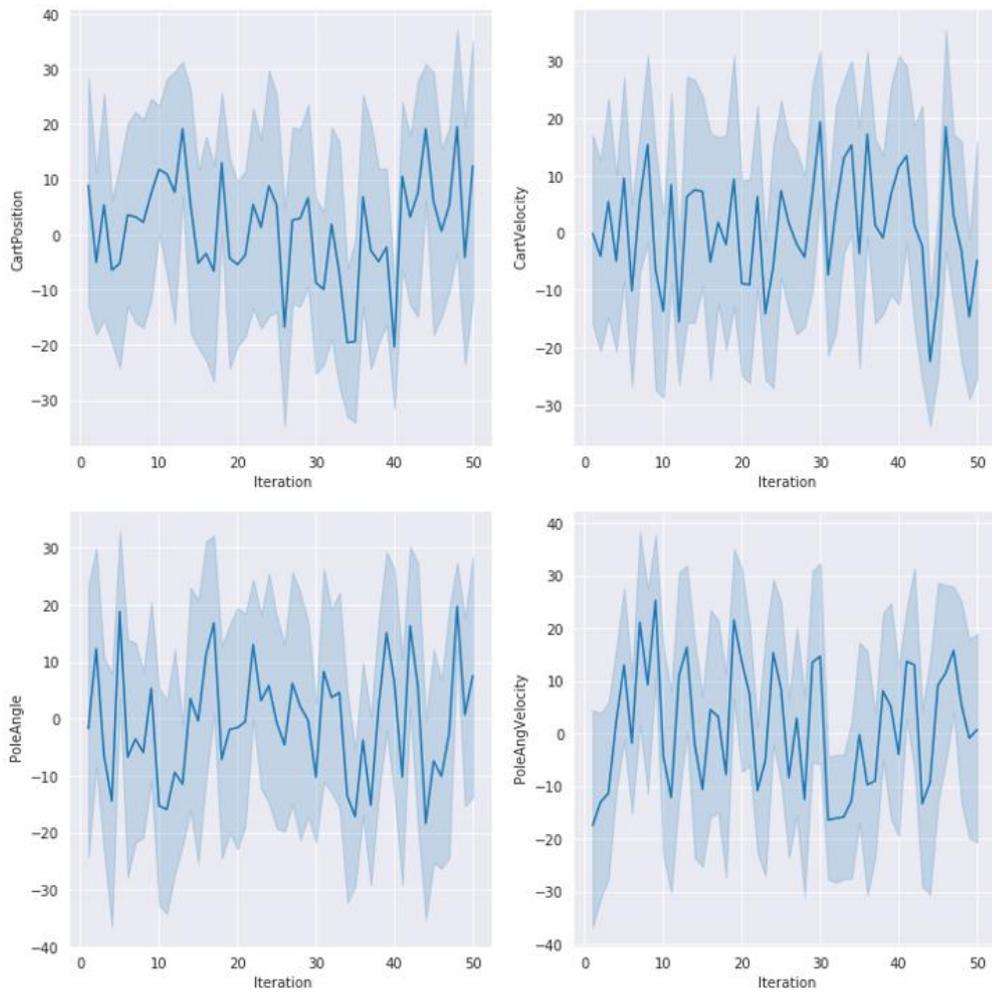

Fig. 5.16 Change of initial states in nine test episodes after each training round

The results suggest that the initial states changed randomly in the nine parallel rounds after each training episode. There is no evidence showing significant changes before and after the 22nd, 45th, and 50th episodes. For more details of the initial states, please refer to Appendix.

Conclusion:  ①The agents' policy deterioration did not start in the nine parallel tests after each training round. Because rewards that the agent obtained in the 9 test rounds were very close and equally low, which proves that the policy has already deteriorated in the training round;  ② It is not the extreme initial states causing the policy to degrade. There is no evidence showing that the initial conditions in those training rounds are more extreme. Therefore, **Hypothesis 1 does not hold**. However, we can continue to explore the relationship between the agent's policy and the initial states of the environment.

### Hypothesis 2: The Randomness of the Initial States Has a Greater Impact on Agent's Performance, Rather Than the Absolute Value of an Initial State

In the original Cart-Pole environment, the initial states are a four-dimensional uniform random variable on the interval $[-0.05, 0.05]$ (the value is just a normalized number without direct physical meaning). Draw the performance distribution of the agent on different intervals of rewards as a bar chart. The bars concentrate on the right side more shows better performance.

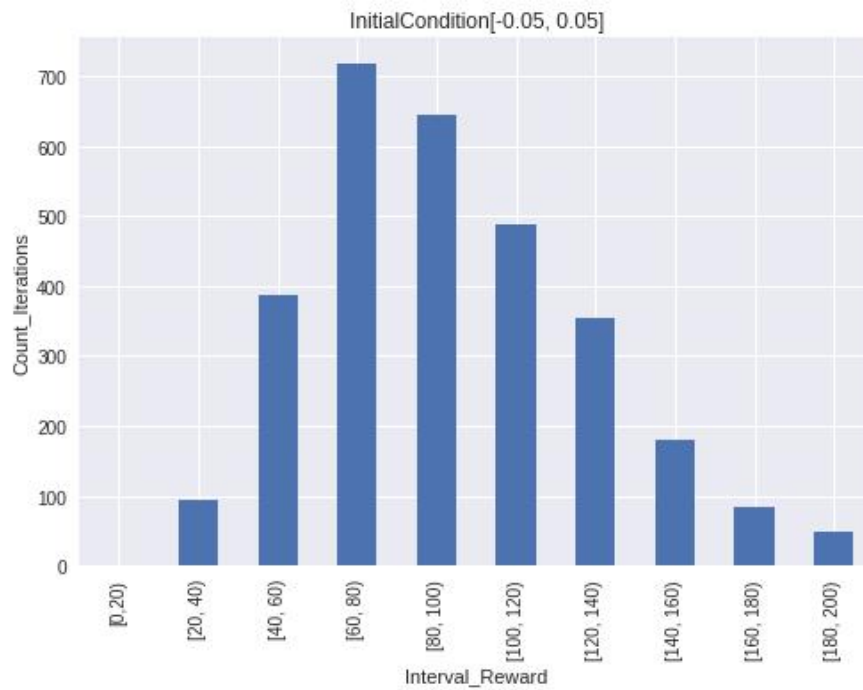

Fig. 5.17 Reward distribution by the "imperfect" agent in test experiment (original initial states)

The obtained rewards mainly concentrate on the interval $[60 - 140]$. **Next, we can modify the uniform interval where random numbers representing initial states are generated.**

As shown in the figure, we modify the intervals to $[-0.02, +0.02]$、$[-0.01, +0.01]$、$[-5e^{-3}, +5e^{-3}]$、$[-e^{-4}, +e^{-4}]$、$[-e^{-5}, +e^{-5}]$、$[-e^{-6}, +e^{-6}]$ as well as zero initial states and test the same agent in Cart-Pole scenario. **The randomness reduced from the upper left to the lower right in order.**

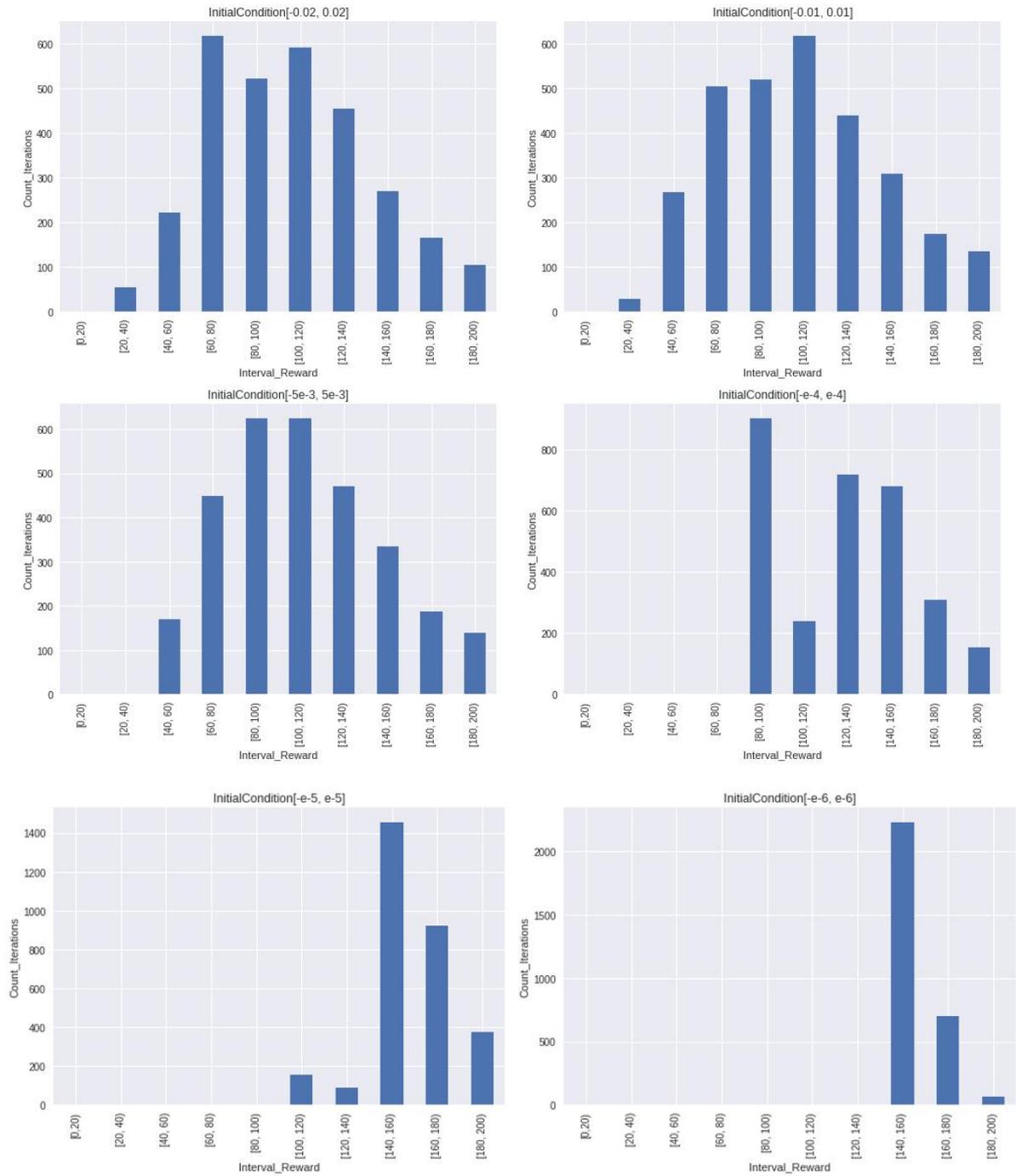

Fig. 5.18 Reward distribution in tests under different initial states interval

It can be seen from the figures: When the randomness is reduced, the obtained rewards by the same agent also tend to be concentrated. Crucially, **under 0 initial condition**s (deterministic initial states), the rewards concentrate on a certain value (**163**)**.** This gives us **essential inspiration**: the policy learned by the IIL agent is deterministic rather than a stochastic policy. So, the actions taken may be the same for the same initial condition, so are the trajectories. Thus, the number of rewards that can be obtained should also be the same.

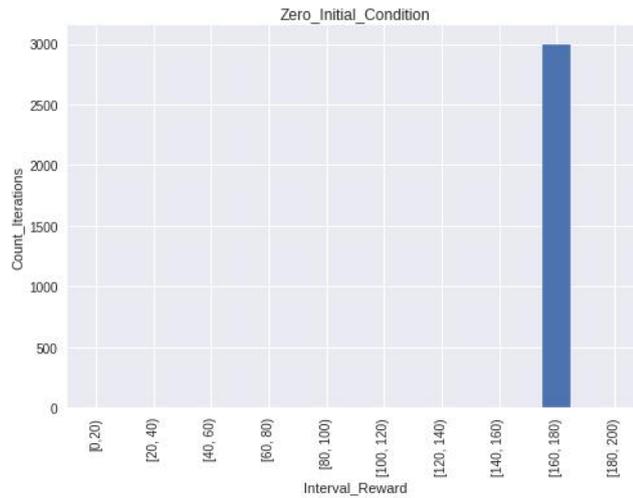

Fig. 5.19 Reward distribution under 0 initial states (fixed value: 163)

Overall, Hypothesis 2 holds. Moreover, hypothesis 3 can be proposed from the experimental results in Hypothesis 2.

**Hypothesis 3: Deterministic Policy May be Poor at Certain Initial States (Biased Policy)**

Since the policy learned by the agent is deterministic, it can be inferred that the randomness of the initial states causes obtained rewards to be scattered. Then, under certain initial states without randomness, the trajectories should be the same, and therefore the obtained rewards should also remain the same.

We modify the initial states to a series of values and visualize the test results. The following charts are divided into two groups: group one contains the test results under initial states with positive values, while group two includes the equivalent negative counterparts.

The line charts summarize the testing results over all the initial states.

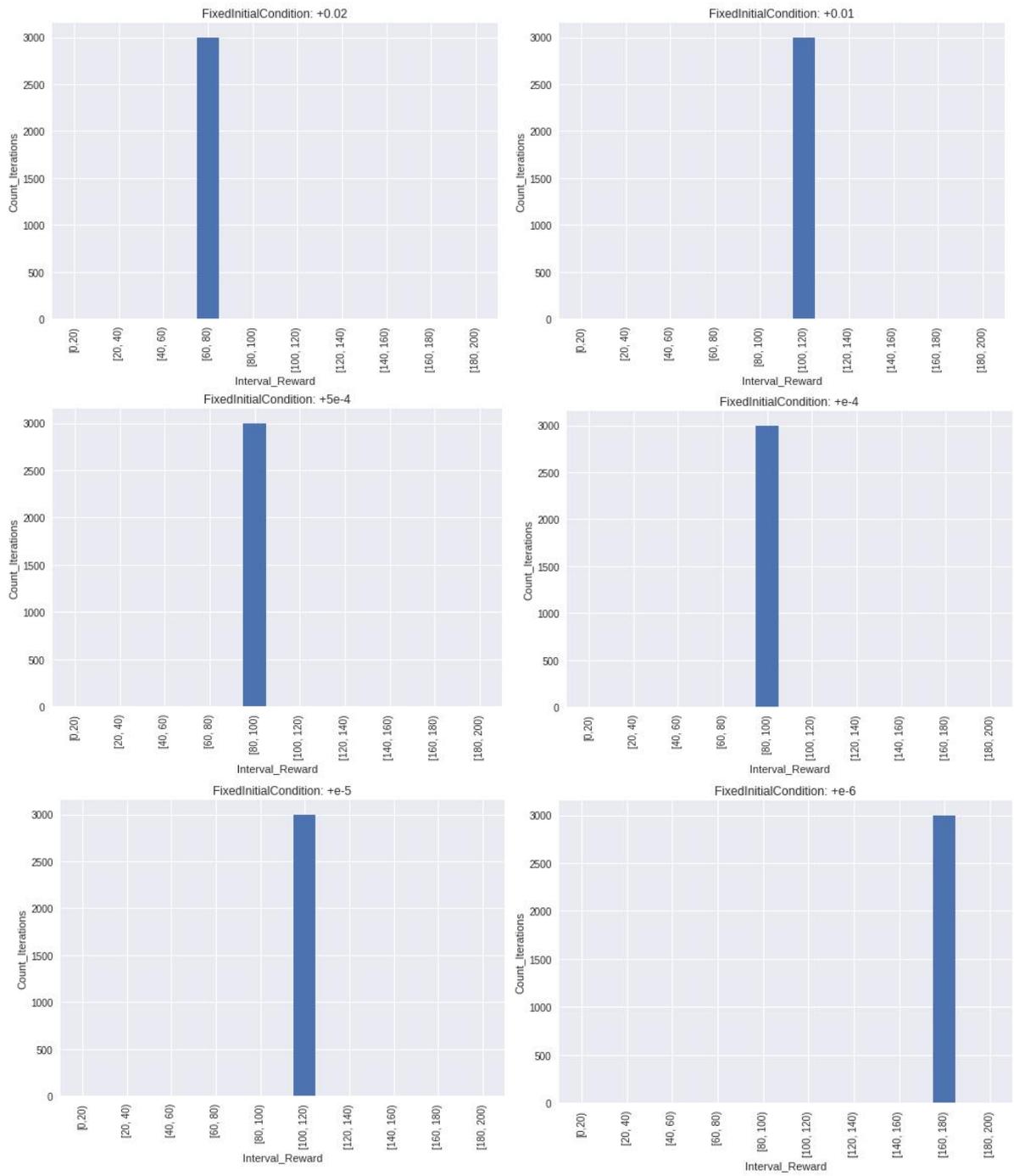

Fig. 5.20 Reward distribution under positively valued initial states (absolute value decreases from top left to bottom right)

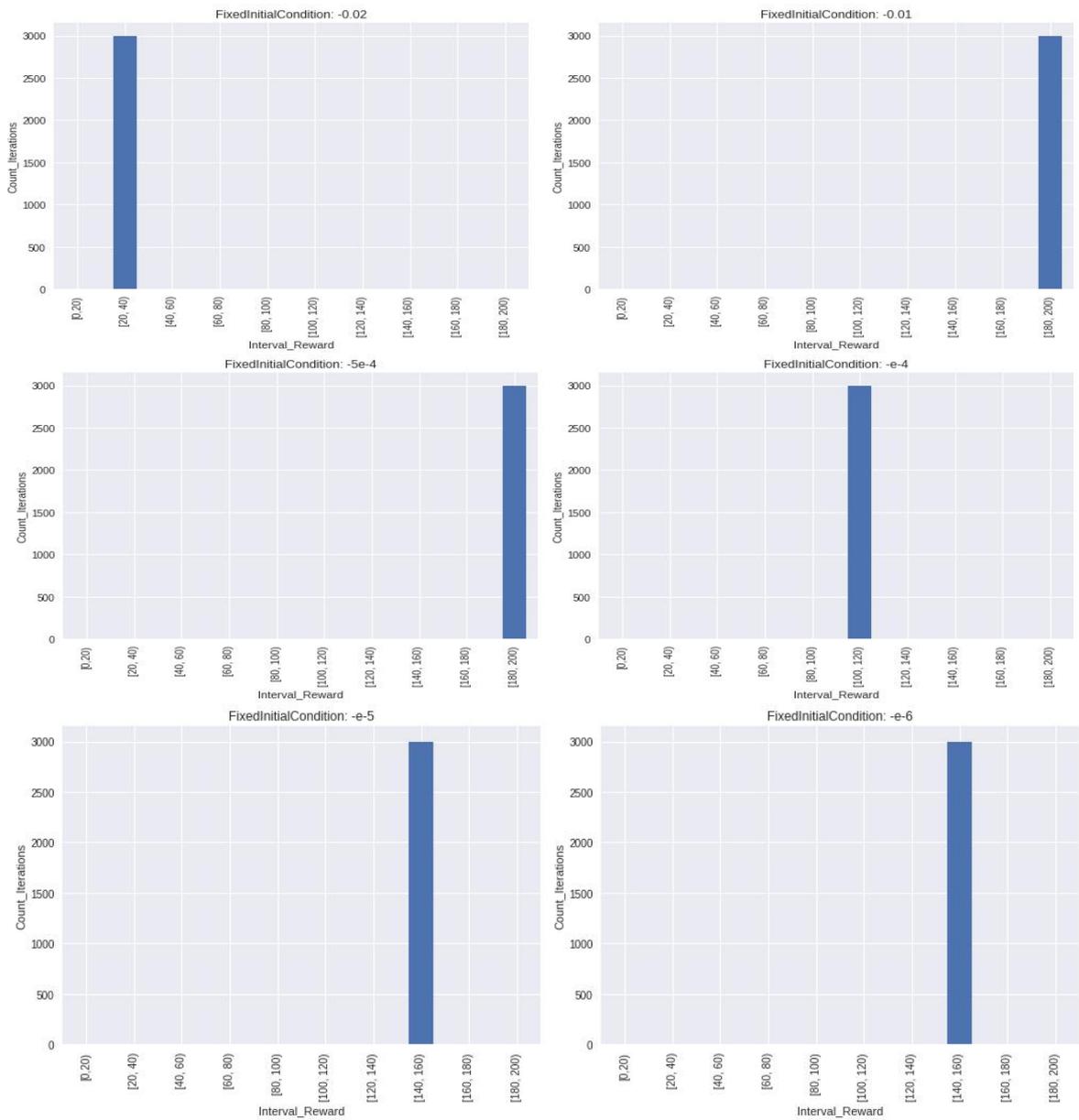

Fig. 5.21 Reward distribution under negatively valued initial states (absolute value decreases from top left to bottom

right)

Similarly, we test the unique DCOACH agent, which is unqualified to completely solve the Cart-Pole problem, under a set of deterministic initial states. Results are as follows:

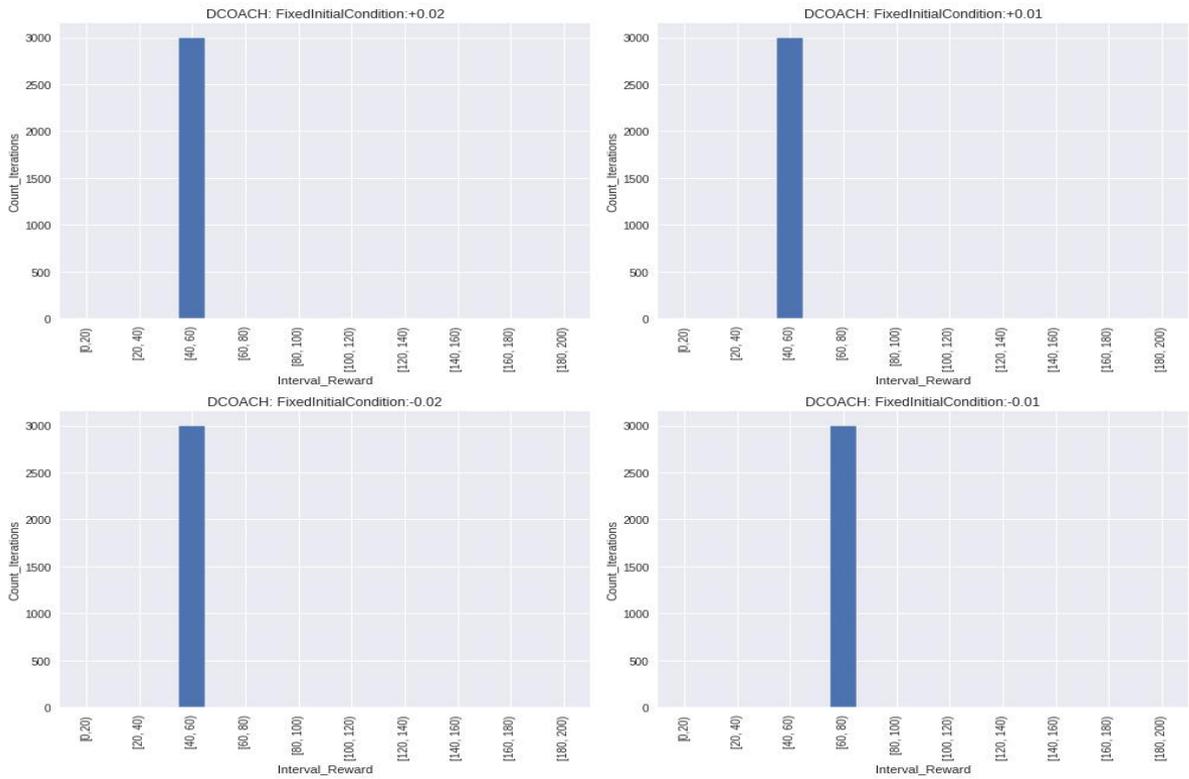

Fig. 5.22 Same scenario with DCOACH agent (deterministic rewards under deterministic initial states)

Summarizing the above testing results, we have:

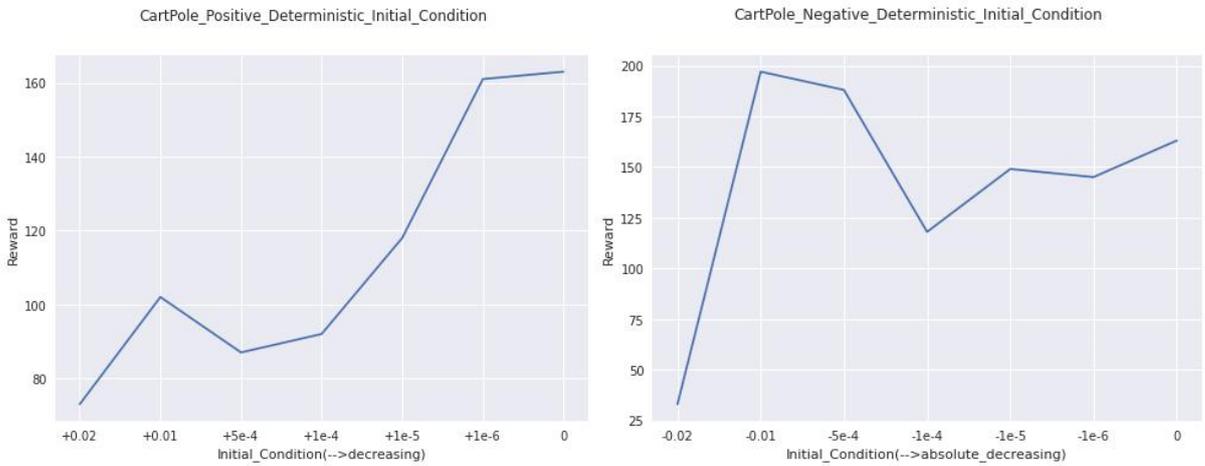

Fig. 5.23 Obtained rewards under different initial states (a series of deterministic values)

Conclusion:

(1) What the IIL agents learn in the Cart-Pole scenario is a deterministic policy. One disadvantage of the learning-based control method is that the "end-to-end" characteristic of the algorithm makes it lack interpretability. For a Fully Observed Markov Decision Process, the optimal and deterministic strategy must exist[33]. At this time, the optimal deterministic policy gives the algorithm more interpretability;

(2) **The learned policy may be biased**. It can be seen from the line chart that the number of rewards an agent can obtain is not inversely proportional to the absolute value of the initial conditions of the environment. The obtained rewards under "bad" initial states are not necessarily low, which is determined by the training process. The reason for the sudden instability in the above-mentioned training processes is insufficient training for a certain initial condition. **In other words, the agent forms a biased policy in a brief training period, which may be good or bad at solving problems under certain initial states**. The reason is still the limited training data (short training process), which can explain why the initial conditions did not worsen in the 22nd, 45th, and 50th rounds, but the performance deteriorated.

(3) Both of the IIL agents failed to completely solve the Cart-Pole problem in the tests because instability took place in the 50th training episodes, which was not corrected and thus, reversely 'broke' the policy.

(4) Some extra training episodes under certain initial states could be a solution for those 'imperfect' agents. This also indicates that because of the obtained deterministic policy, interactive imitation learning algorithms are more 'controllable', making it possible to reinforce the existing policy.

### 5.2.5 Reinforce Training

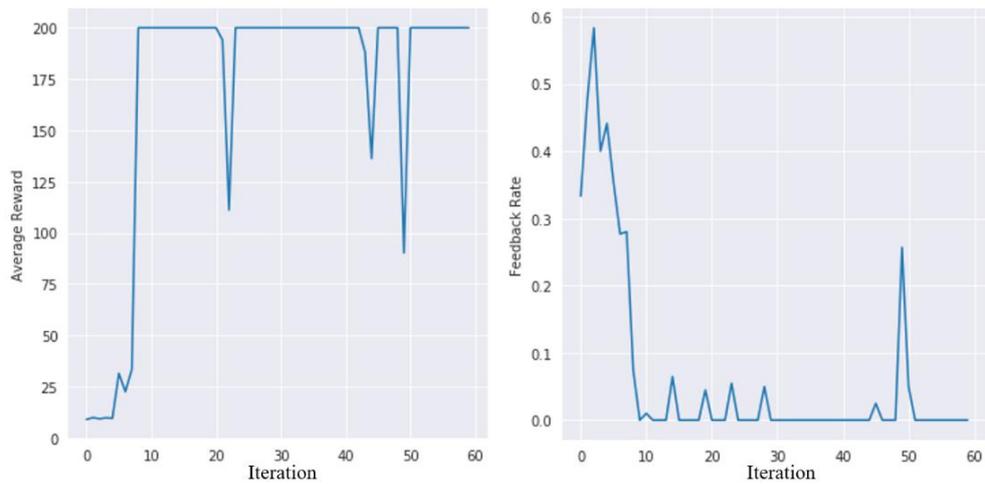

Fig. 5.24 ten more training episodes for the "imperfect" agent

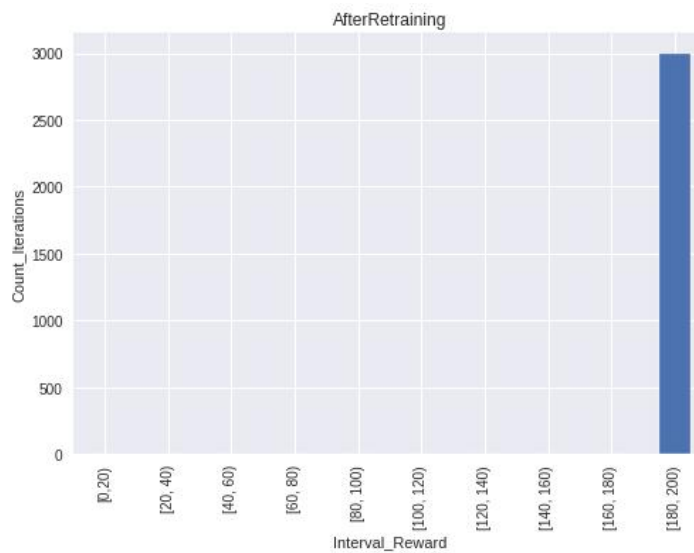

Fig. 5.25 The agent became "perfect" after adding extra training episodes

After ten extra training episodes, the policy successfully handles the Cart-Pole problem, as the figure above shows, which proves that 'reinforce training' is feasible for the deterministic policy obtained by IIL agent.

### 5.2.6 Conclusion

(1) This chapter compares two IIL algorithms with model-free reinforcement learning methods through a large number of training and testing experiments. The **superiority** of the IIL methods in the Cart-Pole scenario has been proven. At the same time, the performance of using IIL based on feedback from state space is equivalent to DCOACH algorithm in the Cart-Pole scenario since teaching in both algorithms is **equally intuitive** here.

(2) The **significantly shortened training procedure** proves that **transferring human knowledge** to the agent through interactions is useful in Cart-Pole scenario, which represents a low-dimensional fully observable MDP process with discrete action space. At the same time, the performance of the IIL agents is also way better compared with the regular RL algorithms.

(3) Considering the **instability** phenomenon, this chapter makes several hypotheses, validates them with extensive experiments, and gives corresponding conclusions. The policy learned by the agent is a **deterministic** and a bit **biased policy**. Specific 'not fully developed' initial states caused sudden instability of the agents.

There were two agents, whose policy did not completely solve the Cart-Pole problem in the testing experiments. This is because instability took place in the 50th training episode, which was not corrected and thus, reversely 'broke' the policy. Nonetheless, it is also proved that this kind of policy can be strengthened through extra training episodes.

(4) Finally, based on the existing experimental data, **the stability of the DCOACH agent** is slightly better than that of the IIL based on the state-space feedback. However, this only indicates the training process. For robots based on IIL algorithms, the training process and the testing (working) process are entirely separated. From the test results, both can solve the Cart-Pole problem very well. The DCOACH agent is more stable during the training process perhaps because it has only one neural network model, while its counterpart has two, which makes convergence harder. The specific reasons need more research to verify.

(5) The advantage of giving feedback in state space is that the natural decision-making process of humans is from the ideal state transition to the actions to be executed. In the Cart-Pole scenario, it is equally intuitive giving feedback in both ways. This also indicates how should one choose a suitable algorithm for task completion. The next chapter will compare both algorithms in Reacher scenario, where feedback mechanism is way more different than here.

(6) It is worth mentioning that the analysis towards the sudden instability is not only for Cart-Pole scenario. Instead, it indicates what our agent has learned using IIL algorithms and advises choosing model in different scenes.

## 5.3 Reacher

Chapter 5.3 mainly trains, tests, and compares two agents based on interactive imitation learning algorithms with RL baselines. The results show that with the participation and feedback of human demonstrators, the IIL agents are significantly superior to the RL agents in both the speed of convergence and the model's performance.

Secondly, the second half of Chapter 5.3 takes the Cart-Pole scenario as an example, conducts extensive hypothesis-driven experiments on the interpretability of the two IIL agents, and discusses their training process.

This chapter compares the DCOACH algorithm and the TIPS algorithm in the Reacher task. Unlike the ordinary Reacher experiment, in this experiment, the agent has no access to the target object's location. Instead, the correction signals from the human demonstrator will be used as an offset to guide the agents to complete the task. Also, the object will have a fixed location throughout different episodes.

The training process in this chapter takes 50 episodes for each. After each training round, nine test rounds are carried out, and the average rewards is taken as the metrics.

### 5.3.1 DCOACH: 'Arduous but Fruitless'

According to the feedback provided by the instructors, the training process of using the DCOACH algorithm is challenging. It requires high mental concentration. Demonstrators are required to give feedback to control each joint of the robotic arm, which requires much practice and familiarity.

As for the physical and mental feeling of training two kinds of IIL agents: giving corrections using a keyboard is extremely hard for training DCOACH agent in this scenario. Because the arrow keys correspond to the two robotic joints, it is hard to be coordinated by a non-expert human demonstrator.

The figure below is the first training round. It can be seen that the number of rewards obtained by the agent during the entire training process is unstable, and the frequency of human feedback is also fluctuating. There is no evident trend of development of agent's policy.

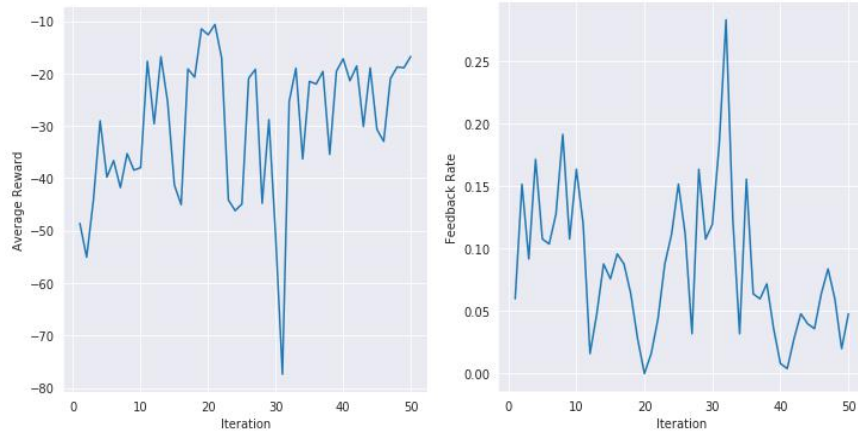

Fig. 5.26 "Warm-up" training of DCOACH agent in Reacher

After the first warm-up experiment, a formal group of experiments was started. Totally five experiments were carried out. Each lasts 50 episodes. The corresponding learning curves and feedback rate are plotted as follows:

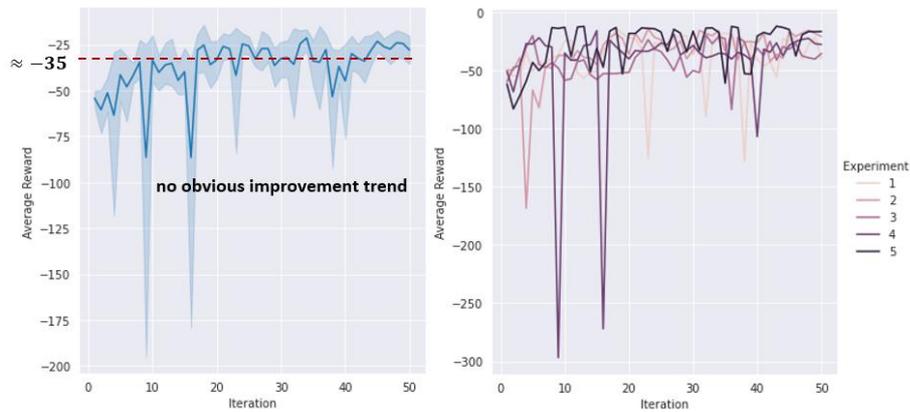

Fig. 5.27 Learning curves of DCOACH agent in Reacher

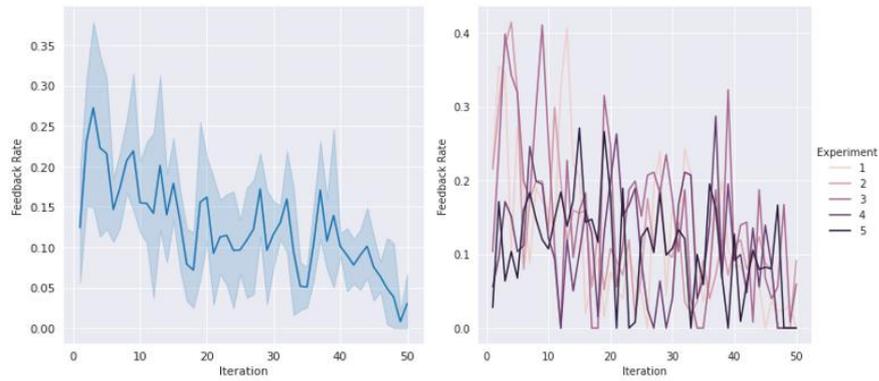

Fig. 5.28 Curves of feedback rate of DCOACH agent in Reacher

It can be seen that despite many efforts, the training effect is not ideal, as the obtained rewards have no apparent upward trend. During the training process, it is difficult for human demonstrators to provide feedback to coordinate the movement of the two joints of the robotic arm, especially under the premise that the agent will output random movements to the robotic arm.

### 5.3.2 TIPS: Efficiency and Easiness

To summarize why it is difficult for human demonstrators to provide feedback during the training of the DCOACH agent, the arrow keys are used to control the torque of the robotic joints, which is not intuitive and requires much skilled work.

In contrast, using the arrow keys to provide feedback on the relative location of the end effector of the robotic arm and the target ball is more in line with human intuitive perception. According to feedback from the demonstrators:

(1) During the training process, providing the feedback signal is very intuitive. One can get started quickly without much skilled work

(2) The training experience is relatively easy, and there is no sense of urgency and need to maintain concentration throughout the whole training process.

The figure below is the training results of the TIPS agent in Reacher-v2 with non-expert human demonstrators. As can be seen, even when the human demonstrator has no experience before, the entire training curve shows a certain pattern: the level of rewards has increased significantly, the overall feedback rate continues to decline.

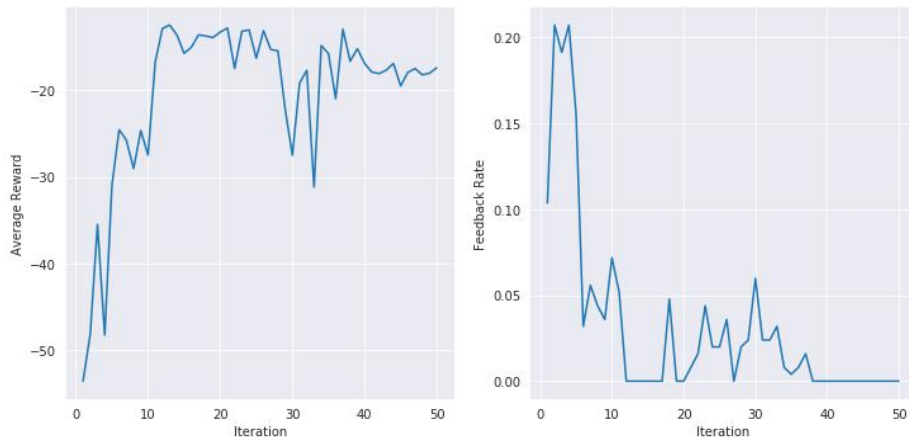

Fig. 5.29 "Warm-up" training of TIPS agent in Reacher

After the warm-up group, the agent is trained five times. The figure below shows the change in the obtained rewards and feedback rates during the five training sessions.

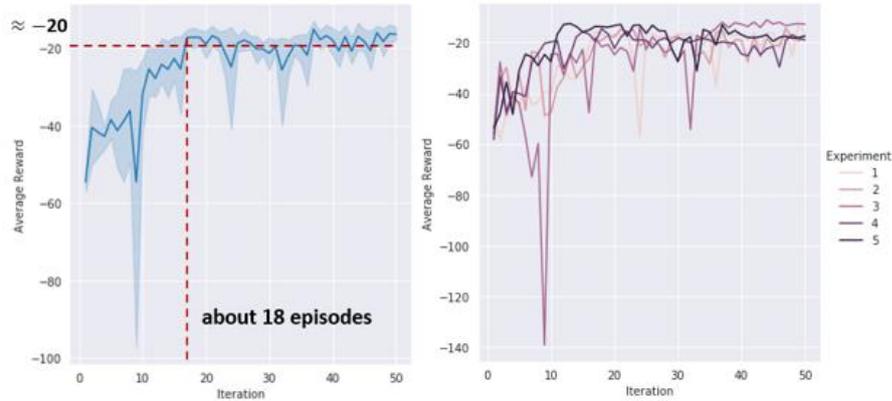

Fig. 5.30 Learning curves of TIPS agent in Reacher

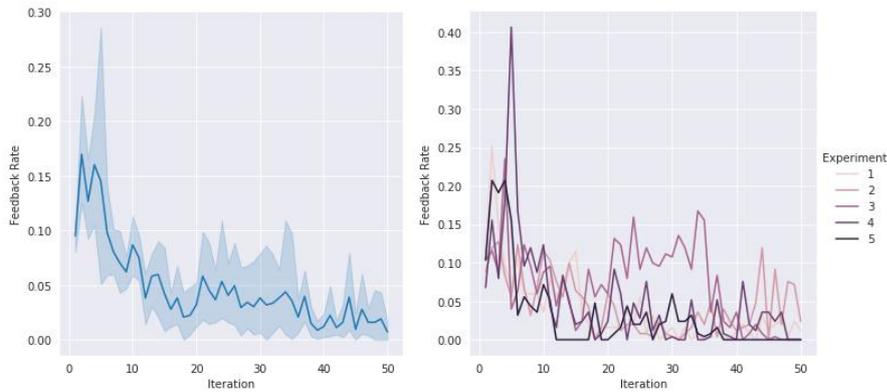

Fig. 5.31 Curves of feedback rate of TIPS agent in Reacher

It can be seen that the training results are relatively stable, the variance of the agent's performance in the five experiments is slight. The obtained rewards are stable at about -20 after forming an initial policy. However, another phenomenon was observed during the training process: the robot arm's end-effector often oscillates around the target point and cannot be completely accurately stabilized at the object. Chapter 5.4.3 gives an analysis and solution for this phenomenon.

### 5.3.3 Adjustment for Solving Oscillation Problem

According to the experimental results from chapter 5.4.2, TIPS can solve the problem very well already. Non-expert human demonstrators can also train a qualified agent, and the whole procedure does not require much effort.

However, a new problem was noticed during the training of the TIPS agent: In order to quickly guide the agent to learn the initial policy, the error correction constant $e$ is rather big, which leads to: after guiding the robot to form an initial policy, **a large error correction constant $e$ cannot reinforce the end-effector precisely stay at the target point, which causes the oscillation around the target point.** We have tried to reduce the error correction constant $e$ to 0.2, 0.1······ times of the original, but it is even difficult to form the initial policy. Because at the early stage, only a relatively large $e$ is intense enough to force the agent to the direction of the object.

Thus, we took the following solution:

(1) Using two-level feedback signals, when the agent forms a stable policy and the end-effector moves around the target point, a small error correction constant is used to perform more precise control.

(2) Adding a stop signal in the action space. The demonstrator should continue to give a stop signal when the end-effector reaches the target point. **The purpose is to make the Demo-Buffer have more data of stopping at the target point**, thereby enhancing the agent's stability near the target point.

```python
# define two levels of feedback signals
def key_press(self, k, mod):
        if k == key.SPACE:
                self.restart = True
        if k == key.LEFT:              # use normal signals initially
                self.h_fb = H_LEFT
        if k == key.RIGHT:
                self.h_fb = H_RIGHT
        if k == key.UP:
                self.h_fb = H_UP
        if k == key.DOWN:
                self.h_fb = H_DOWN
        if k == key.Z:
                self.h_fb = H_HOLD
        if k == key.X:
                self.h_fb = DO_NOTHING
        if k == key.J:
                self.h_fb = G_H_LEFT      # use finer signals around the target
        if k == key.L:
                self.h_fb = G_H_RIGHT
        if k == key.I:
                self.h_fb = G_H_UP
        if k == key.K:
                self.h_fb = G_H_DOWN
```

The training process after the optimization is as follows:

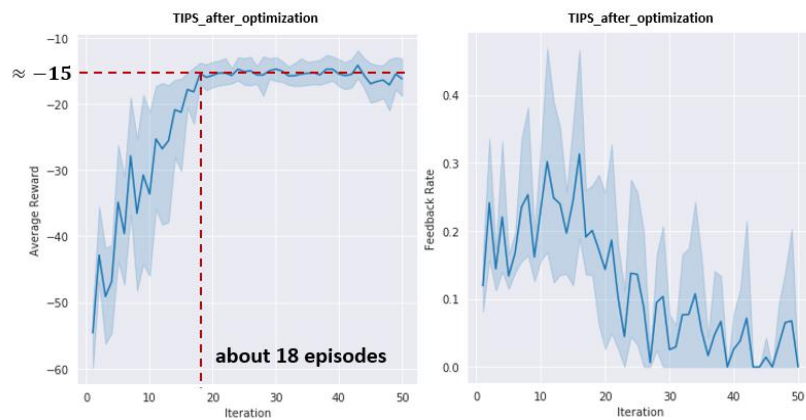

Fig. 5.32 Learning curves of TIPS after optimization

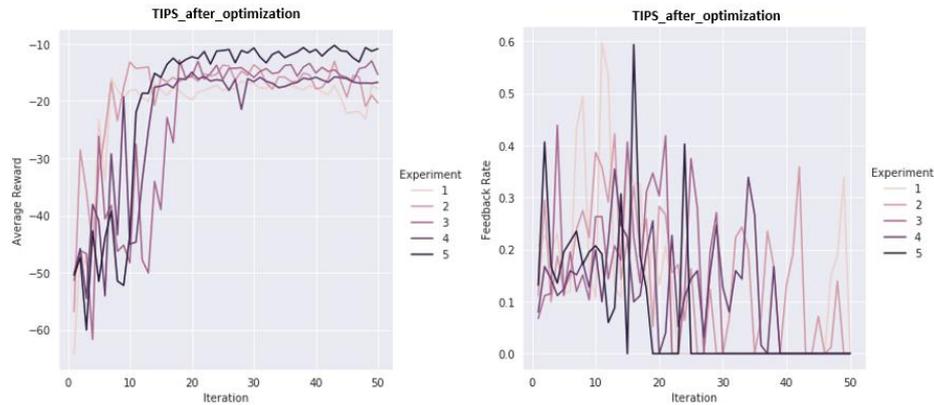

Fig. 5.33 Curves of feedback rate of TIPS after optimization

It can be seen from the figure, the training process after optimization is different from the previous one. Two points are worth noticing:

(1) After adjustment, the stabilized rewards are significantly higher than before (about -15 compared to about -20. The absolute value of the reward number represents the distance to the center of the target ball; the model performance is improved by about 25 %). The oscillation phenomenon of the agent near the target point completely disappears, and the end-effector can quickly stable itself (see more in videos).

(2) In the latter training process, the feedback rate is increased because the forced stop signals are also counted.

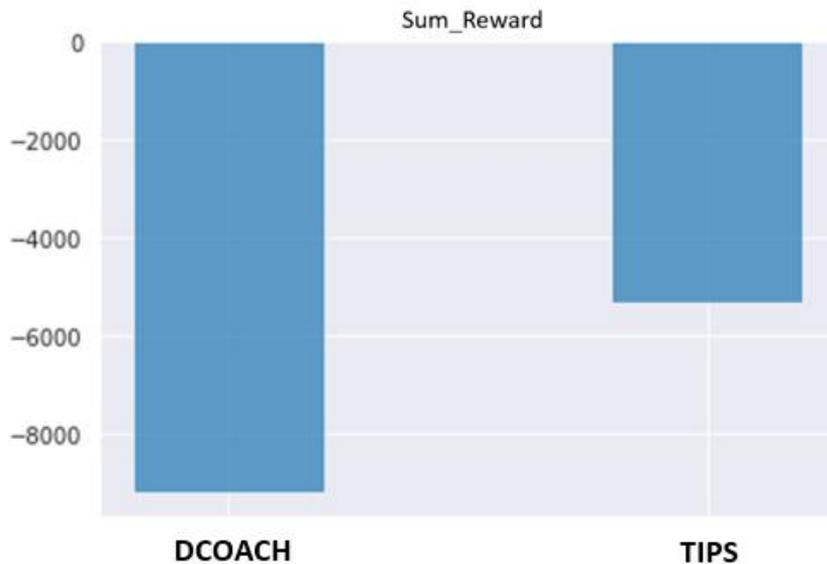

Fig. 5.34 Accumulated rewards of the two agents

Finally, the optimized TIPS model is compared with the DCOACH agent. The ordinate in the figure is rewards obtained by the agent throughout the five 50-rounded training experiments. The reward is a negative value, so the smaller means closer to the target. It can be seen that the performance of the TIPS agent is almost twice than that of the DCOACH algorithm.

### 5.3.4 Conclusion

Chapter 5.4 mainly compares the performance of two IIL algorithms in the Reacher scenario, where the agent should plan their trajectory for reaching the target object. The results show that under the premise of non-expert demonstrators participating in the training, IIL based on action space does not provide feedback that conforms to human intuitive perception, resulting in not being able to provide high-quality feedback signals and unsatisfactory performance.

In contrast, the Reacher scene is more suitable for IIL algorithms based on state space. The relative location between the target point and the end-effector of the robotic arm is more intuitive and suitable for keyboard-based correction.

In addition, towards the problem of the TIPS agent oscillating near the target point during training, the model is adjusted by using two-level feedback signals and forced stop signals. The Demo-Buffer has more state-action pairs of being stopped at the target point, **which completely solves the problem and increases the model performance by 25%.**

Overall, in this task scenario, IIL based on state-space fully displays its advantage. However, note that coupling the motion of different joints can also simplify the demonstration for a specific task. But the way of coupling may vary with the change of task scenarios, which may harm the generality and flexibility of the algorithms. Aiming to enhance the easiness of programming of the robots, IIL based on state space should be considered a better choice.

**5.4 Lunar-Lander-Continuous**

Chapter 5.4 Applies interactive imitation learning algorithms to perform control tasks in a continuous action space and static task scene. In contrast, Chapter 5.3 applies IIL algorithms to a dynamical scenario with discrete action space. This chapter focuses on a highly dynamic task with continuous action space—Lunar-Lander.

The baseline used in this chapter is Double DQN (Deep Q-Network), where a target network is used to eliminate the 'moving target' problem and is more robust. Because of the time limitation, we did not implement Double DQN in the Lunar-Lander scenario. Instead, we use the experimental results at GitHub[2].

In the Lunar-Lander scenario, the task is to smoothly land the spacecraft from different initial states on the platform in the middle of the screen. If the specified time is exceeded or the spacecraft touches the ground incorrectly or flies out of the screen, the task will fail. The initial states of the spacecraft are arbitrary. That is, it may have a significant speed and deflection angular velocity. The agent must learn to adjust for different initial states and finally achieve a smooth landing.

If the agent stops on the landing platform with a correct posture and state, it will receive a reward of 100-140 points. If the landing is stable, it will receive an additional 100 points' reward. Among them, the spacecraft has unlimited fuel, but firing the main engine (pushing the spacecraft upward) will get -0.3 points for each timestep. When the spaceship crashes (-100 points' reward) or flies out of the screen (failure), or lands smoothly (success), this round is considered to be over. When the number of cumulative rewards for this round exceeds 200, the problem is considered "solved".

In this segment, 160 rounds of training are conducted for each agent. Similar to the two scenarios above, there are nine parallel testing rounds after each training episode. Their average rewards are used as the metrics. After training, 500-rounded testing is carried out for each agent, whose policy will not be updated anymore.

5.4.1 Double DQN in Lunar-Lander-Continuous

Table 5.2 State space of Lunar-Lander-v2

| Dimension | State |
|:---:|:---:|
| 0 | Horizontal coordinates |
| 1 | Vertical coordinates |
| 2 | Horizontal velocity |
| 3 | Vertical velocity |
| 4 | Deflection angle |

[2]  https://github.com/svpino/lunar-lander

| 5 | Angular velocity |
|---|---|
| 6 | Touching ground of foot 1 |
| 7 | Touching ground of foot 2 |

The state-space of this task scenario is represented as an array of length 8. The last two digits are binary, where 0 represents not touching the ground of the leg of the spaceship and 1 vice versa.

Instead of doing gradient ascent step on the RL objective, DQN uses neural network as function approximator to learn a Q-function, representing the prospective obtained rewards from the current state and action. Using two neural networks as the Q-network and target network, the problem of moving target is eliminated by Double DQN, which is considered more robust. More details of the RL algorithms are introduced in Appendix.

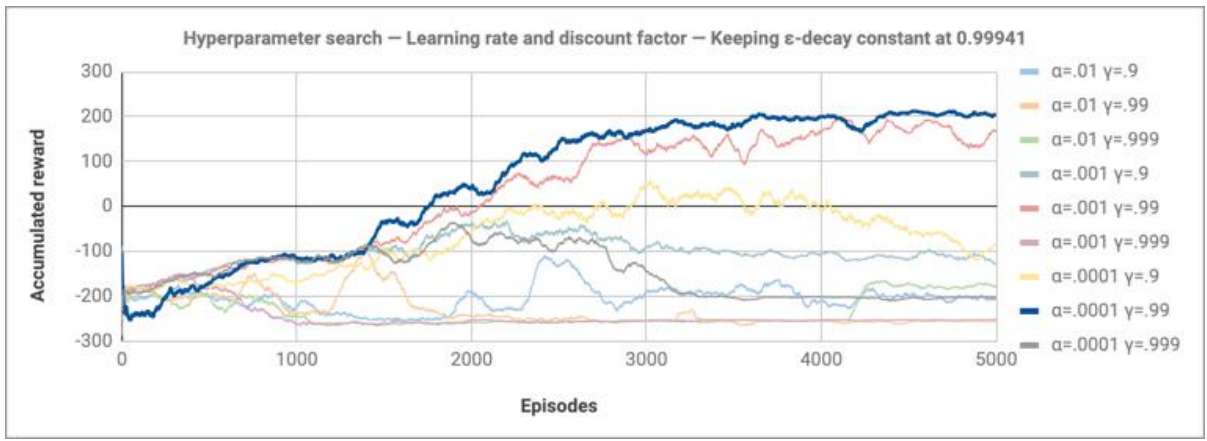

Fig. 5.35 Double DQN agent in Lunar-Lander with fixed $\varepsilon$ and various $\alpha$ and $\gamma$

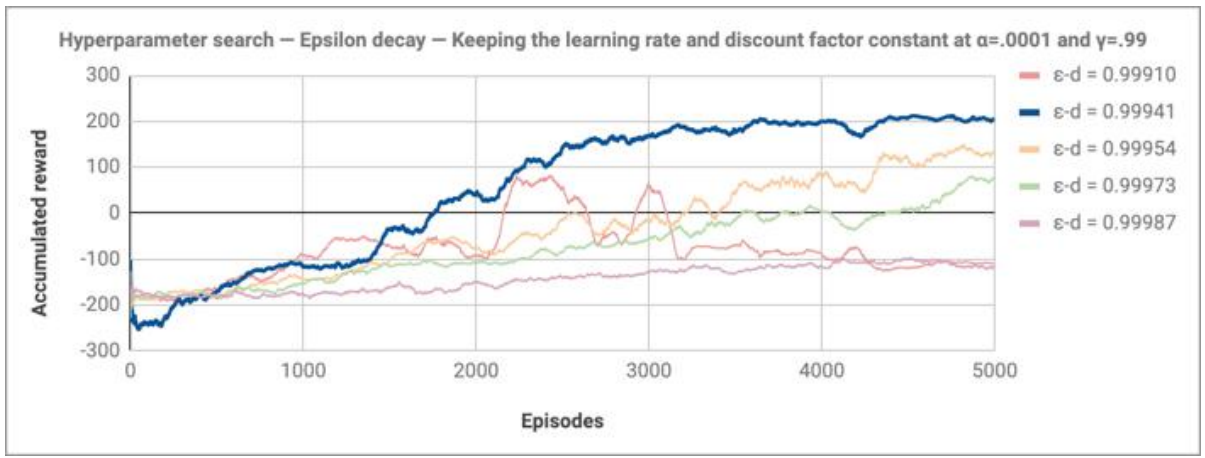

Fig. 5.36 Double DQN agent in Lunar-Lander with fixed $\alpha$ and $\gamma$ and various $\varepsilon$

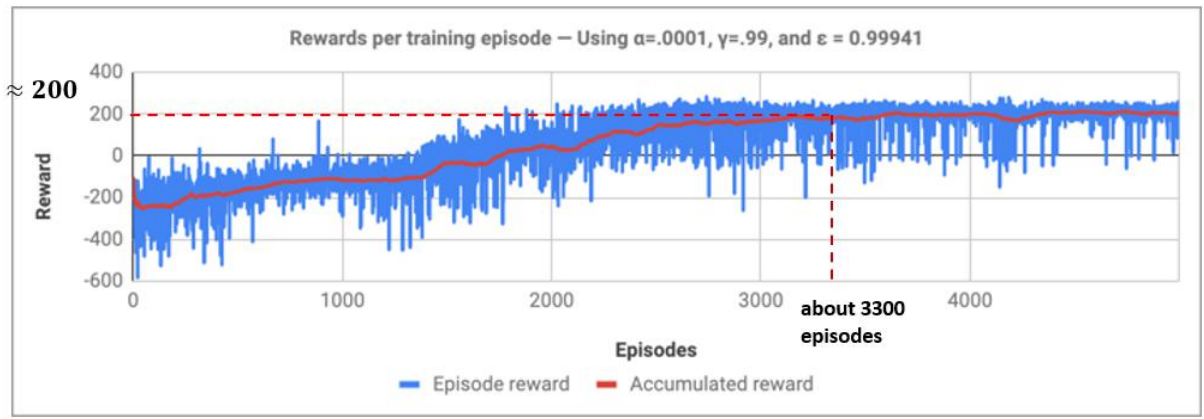

Fig. 5.37 Learning curve of the optimal Double DQN agent

The above is the training curve of using Double DQN to solve the Lunar-Lander problem (using grid search to experiment with different learning rates, decay rates, and greedy coefficients). First of all, it is worthy of recognition that in the case of greedy coefficient $\varepsilon_d = 0.99941$, the decay rate of rewards $\gamma = 0.99$ and learning rate $\alpha = 0.0001$, the experiment result is outstanding. The rewards obtained by the agent have shown a steady increasing trend.

In the end, around 3300 rounds, the number of rewards of the agent can stably reach about 200. However, even after reaching 200 stably, the number of rewards it receives is also very unstable, which is also a common problem in conventional reinforcement learning.

In summary, even an excellent algorithm in this scenario has problems such as difficulty in convergence and unstable reward numbers in dynamic scenarios. The next chapter will show the experimental results of using the IIL algorithm agent to perform control tasks in the Lunar-Lander scene. Finally, the obtained rewards can stabilize at about 200 (from about the $3000_{th}$ round afterward). However, even after acquiring a qualified policy, the RL agent may still suffer from instability, as the figure shows.

### 5.4.2 TIPS in Lunar-Lander

The previous chapter showed the experimental results of using Double DQN for solving Lunar-Lander problem. It can be said that the results of which have been excellent.

This chapter applies IIL algorithms in the same task scenario. Moreover, in section 5.3.3, we analyze the interactive imitation and reinforcement learning algorithms in dynamic scenarios and give comparative suggestions for applying different algorithms.

**Too Intense Correction Signals on Angular Velocity Leads to Mission Failure**

In the pre-experiments, the agent often had a sudden increase in the tilt angle, which caused its horizontal velocity to be too high and the task to fail. After analyzing the state transition data of the spacecraft during the training rounds, we believed that the key was to control the horizontal speed.

The horizontal speed control is performed by the deflection angle of the spacecraft and the propulsion of the main engine. So a solution similar to that in the Reacher scenario was adopted: a two-level feedback signal was used. A smaller error correction constant is used to correct the deflection angle, while a larger one controls the vertical velocity.

In the formal experiments, the being optimized model with two levels of error correction constant is trained for 160 rounds. It can solve the problem well, as the following shows:

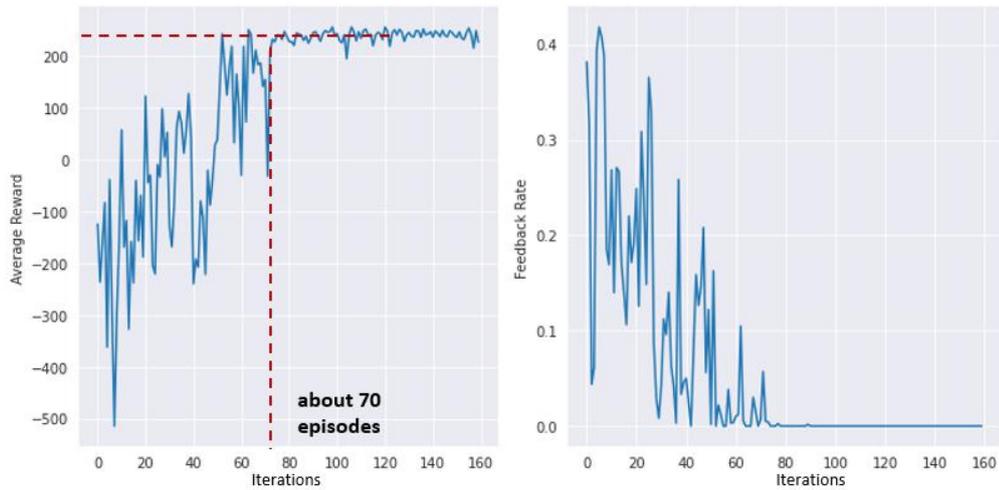

Fig. 5.38 Learning curve of TIPS agent in Lunar-Lander

In the first 50 episodes, the overall obtained rewards increase but are highly unstable. The reason may be: With the limited training data at beginning, local overfitting often takes place. Nevertheless, the situation is improved with the increase in training data. About 70 episodes later, the agent can continuously obtain rewards above 200 and the problem is solved.

Compared with Double DQN in chapter 5.5.1, the performance of IIL agent is obviously better. At the same time, **it is worth noticing that** no instability occurs like in Cart-Pole scenario. The reason may be that, in this scenario, more training episodes mean more training data and states are visited, which can make the learned policy more robust.

After the training session, 500-rounded testing is carried out, the obtained rewards distribute as follows:

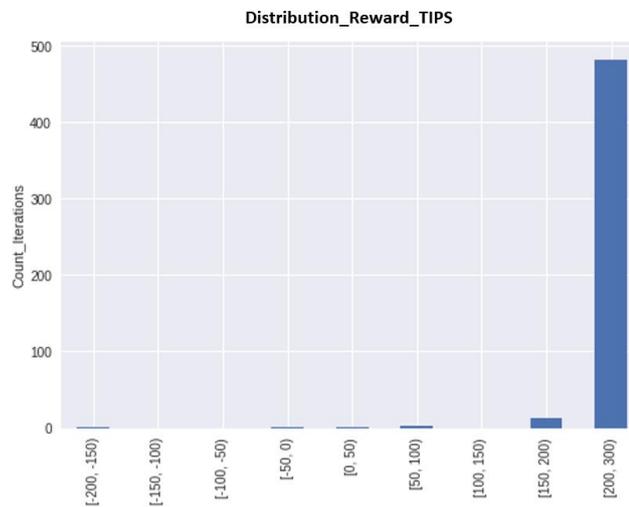

Fig. 5.39 Reward distribution of TIPS agent in 500 testing episodes

It can be seen that, except for the sporadic instability, the agent can solve the Lunar-Lander problem very well.

### 5.4.3 DCOACH in Lunar-Lander

The previous section trains and tests TIPS agents in the Lunar-Lander scenario. TIPS agents solve the problem in a way shorter procedure than that of the RL agent.

This section implements the counterpart DCOACH in Lunar-Lander. In this scenario, providing feedback both in the action and state spaces is equally intuitive. Thus, the goal is not to compare the two algorithms but rather to validate different IIL algorithms in this kind of task and compare them with the RL agents.

The training procedure is as follows:

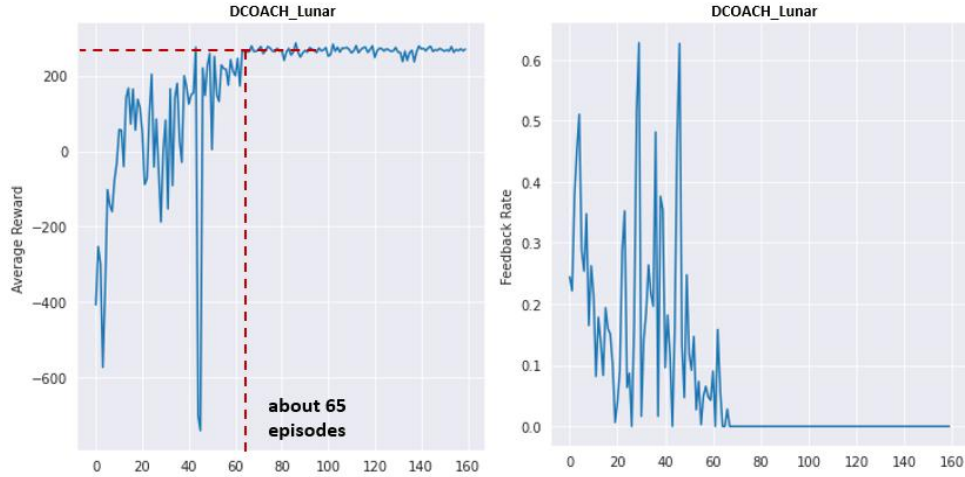

Fig. 5.40 Learning curve of DCOACH agent in Lunar-Lander

It can be seen from the figure, DCOACH based on the feedback in action-space can also perfectly solve the Lunar-Lander problem. After approximately 65 episodes, the obtained rewards can be stabilized above 200. Compared with the training process of the TIPS agent, in the very first beginning, the obtained rewards increase more quickly. The reasons may be the following:

(1) First of all, when training the DCOACH agent, the human demonstrator has already accumulated some experience in this task scenario, which makes the demonstration smoother with DCOACH agent.

(2) Because the DCOACH agent has only one neural network model, thus it can be easier to be corrected by the human demonstrator, while the TIPS agent has two.

500-rounded testing is carried out after training, the distribution of obtained rewards is as follows:

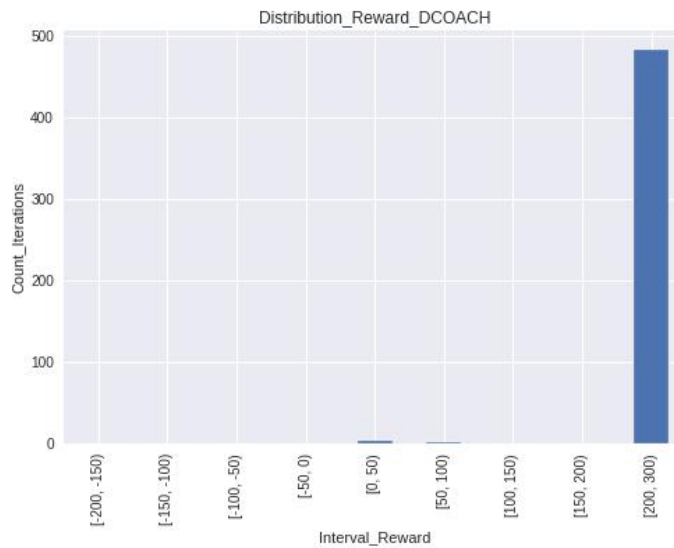

Fig. 5.41 Reward distribution of TIPS agent in 500 testing episodes

It can be seen that DCOACH algorithm can solve Lunar-Lander problem with good stability as well, which again validates the conclusion of this thesis: when providing feedback has no significant difference in action space or state space, the performance of the two algorithms is close to each other (or the stability of DCOACH is a bit better).

### 5.4.4 Stochastic Policy

As mentioned above, the standard IIL algorithm yields a deterministic policy, which may work decently well in some cases, e.g., in Cart-Pole problem.

In any fully observed MDP problems, there must exist an optimal policy, which is deterministic.[28] Nevertheless, in many cases in the real world, the state may not be fully observable. In this case, a stochastic policy must outperform the counterpart.

Also, even if in the case of MDP problem, a human may not be able to provide the perfect demonstrations. The suboptimality of human behavior suggests that simply imitating the action labels provided by humans does not possibly make sense sometimes. In the problem of Lunar-Lander, even if the state is accessible, the highly dynamic scenario still leads to some wrong demonstrations, which makes the obtained deterministic policy might not be optimal.

Regarding this, we set another group of experiments, where IIL agents using Gaussian policy $\pi_\theta(a_t|s_t) \sim \mathcal{N}(\mu, \sigma)$, where $\mu$ is the mapped optimal action by the IIL algorithm, and $\sigma$ is a hyper-parameter.

The following are the learning curves of the IIL agent using TIPS and Gaussian policy with the variance $\sigma = 0.2$.

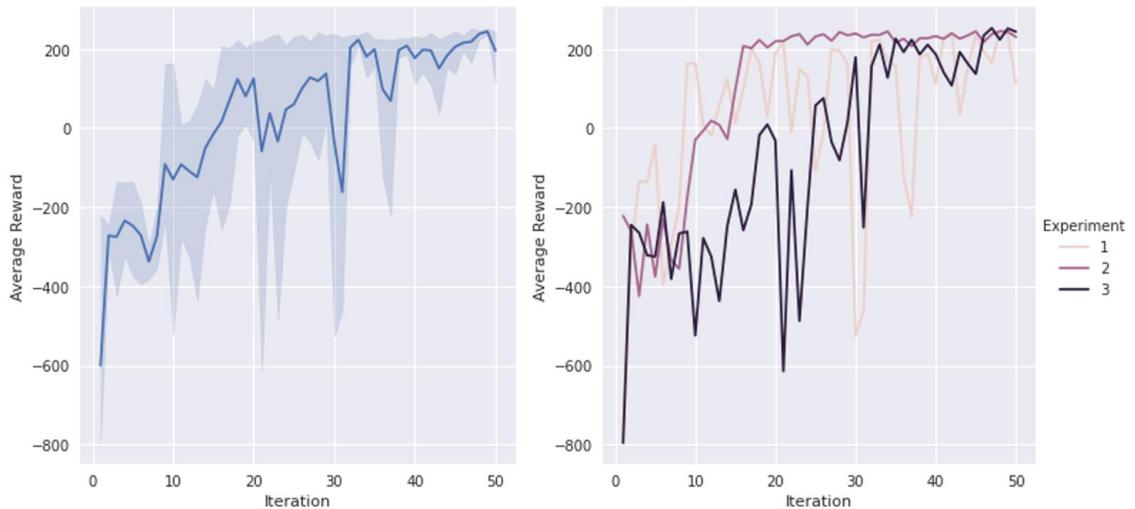

Fig. 5.42 Training process of TIPS agents with Gaussian policy

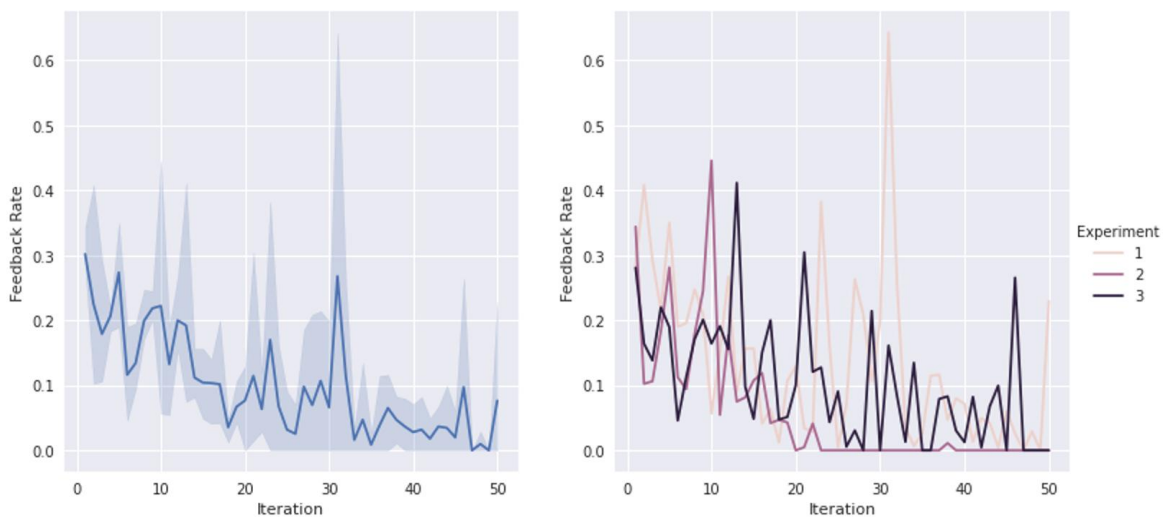

Fig. 5.43 Feedback rate of the training process of TIPS agents with Gaussian policy

As can be seen from the figures, by using Gaussian policy, the training process of IIL agent in Lunar-Lander is much shortened (to 20-30 episodes). The reason may be: the added randomness in policy could offset the suboptimality of human feedback and gain more robustness in the beginning stage of training.

After the training stage, 500-rounded tests are conducted to each agent:

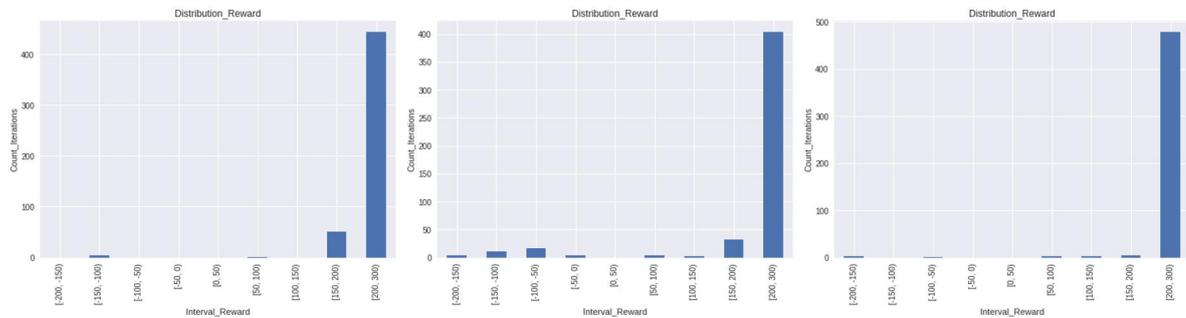

Fig. 5.44 Reward distribution in 3 testing experiments

As a result, the three stochastic policies solve the Lunar-Lander problems very well in tests after 50 episodes of training, which suggests the advantage of stochastic policy even in solving MDP problems in highly dynamic scenarios.

### 5.4.5 Conclusion

First of all, a significant advantage of IIL algorithm is that it can significantly reduce the number of rounds required for training compared with the use of RL algorithms in a trial and error manner. At the same time, a fundamental difficulty in RL can be solved: teaching the agent to balance between the immediate rewards and the potential rewards in the future, namely **delaying enjoyment**.

Humans can work hard for their goals, even if it is difficult to see results in a short time or there are other temptations around them. The reason is that humans have the ability or quality to delay enjoyment. In the Lunar-Lander problem, if the spacecraft lands at a very fast speed and in an imperfect posture, it can get a low but positive reward. However, the mission's goal is to allow the spacecraft to hover in the air and slowly adjust its attitude. During the whole process, the tilt angle should not be too large to land smoothly. Since reinforcement learning is carried out through trial and error, the agent can be trapped in a "local optimum" and choose to land quickly instead of waiting for long-term rewards.

In contrast, the training process with interactions can avoid this problem well because humans can help the agent make decisions and choose the timing of landing. This transferring of human knowledge makes IIL agents be good at delaying enjoyment.

Nevertheless, at the same time, the Lunar-Lander problem is also representative. It can be seen from the curve that at the beginning of the training process using the two IIL algorithms, the obtained rewards fluctuate greatly or even slightly decline. There are two main reasons: (1) Human demonstrators cannot do this task requiring high response speed and coordination ability. They cannot provide perfect training data. However, we can see that, later, even non-expert demonstrators can train a good model. (2) The IIL agents can quickly form an initial policy through limited training data, but this also makes it easy to have "local over-fitting". The model is overfitted to the nearest action and trajectory, which is why it is necessary to train the model using shuffled data in a batch manner.

In the later stage of the training process, when the agent has a stable policy solving the problem, corrections become even easier. Therefore, for agent training using interactive imitation learning algorithms, **the difficulty lies in the early stage of training**. When the agent's policy and behavior are random, how to improve their behavior through providing feedback, and the difficulty of this step also depends on the task. For instance, in highly dynamic scenarios like Lunar-Lander, it is more difficult to give corrections. It also inspires us to choose algorithms: non-expert demonstrators can also train a good model by providing correction feedback, which is proved in section 5.6, where human control is performed as a comparison. In highly dynamic scenarios, the difficulty is to form an initial policy through corrections, which can sometimes be very hard for human nature. At this time, IIL algorithm may not outperform the RL algorithms.

Stochastic policy, e.g., Gaussian policy, may present a better solution for addressing the suboptimality of human demonstrations in highly dynamic scenarios. As section 5.4.4 suggests: Gaussian policy works better for shaping the initial policy and addressing the problem.

## 5.5 Non-expert Human Directly Perform Tasks

Besides, as a comparison, non-expert humans are required to perform the control tasks directly. The following figure shows their performance in Cart-Pole and Lunar-Lander scenarios. In the same episodes, the obtained rewards of the agents dramatically outperform that of humans. Which can also prove that, with giving feedback in a correction signals manner, non-expert human can also train a model, who has way better performance than themselves.

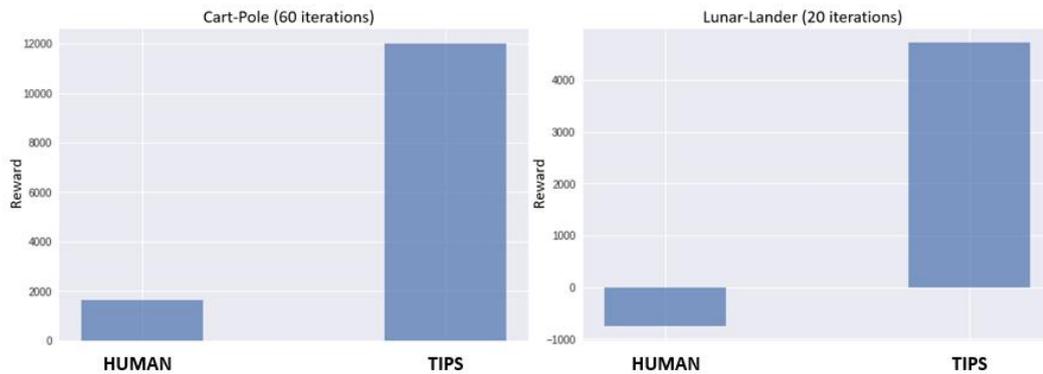

Fig. 5.45 Comparison between human and agent

## 5.6 KUKA-Reach

In this section, the IIL algorithm DCOACH is employed to train the agent to complete the reaching task, where the target object's position is random among four choices. The goal of this setting is to validate the IIL algorithm in high-dimensional space. RGB image from the camera is used as input of the KUKA robotic arm. The agent should learn to classify the position of the goal and plan the trajectory of reaching through interactions with humans. A convolutional neural network is used as the encoder.

The following is the learning curve of the training procedure:

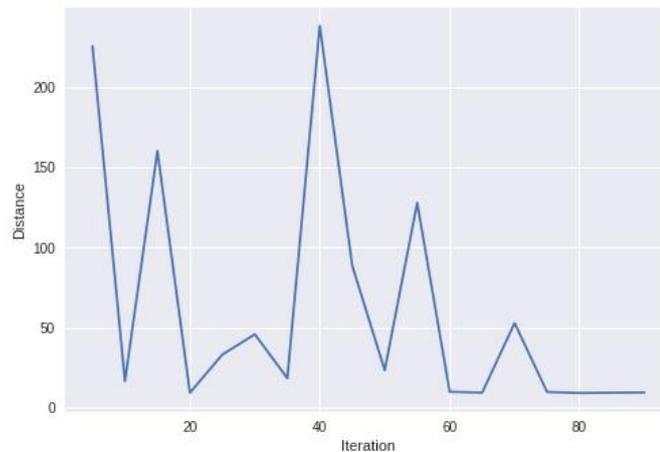

Fig. 5.46 Distance between end-effector and object in training process

It can be seen from the curve that the terminated distance between the end-effector of the robotic arm and the object is decreasing significantly as the training iteration goes up. But sometimes, the distance fluctuates because the training data of some specific locations dominate, which influences the policy for reaching the object on other positions. Also, in the very first beginning, the policy is not robust enough against interference. As the training episodes increase (about 60 iterations), the performance of the agent converges gradually. Thus, it is essential to keep balance in quantity of the different groups of training data, here means the different positions of the object.

After 90 episodes of training, the robotic arm can solve the problem very well with a deterministic policy. In the 50 rounds of testing experiments, it can almost perfectly reach the goal on three out of four positions.

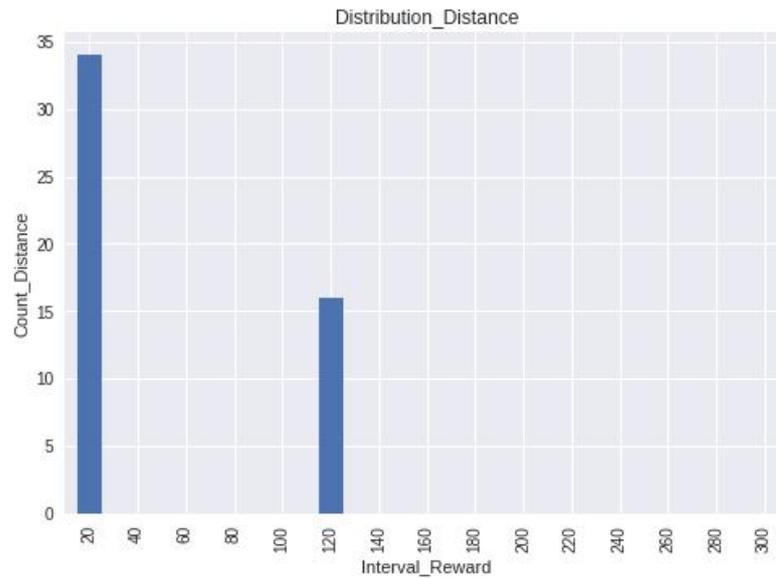

Fig. 5.47 Tesing results in 50 episodes

It is worth noticing that only add extra training episodes for the specific position cannot improve the entire policy since it will also harm the performance of the policy for the other positions.

# 6. Summary

So far, all the theoretical and experimental parts have been displayed above. The most crucial conclusion includes the following four aspects:

(1)  In scenarios where giving feedback in action space and state space makes no significant difference, such as Cart-Pole and Lunar-Lander, both IIL methods can perform way better than regular model-free RL methods in terms of speed of convergence and stability. Also, from the existing experimental results, we can say that giving feedback in action space is a bit more stable than in state space, where a forward dynamics model is required.

(2)  Nevertheless, when providing feedback in state space is more intuitive, the difference becomes extremely huge with non-expert human demonstrators. In the Reacher scenario, IIL in state-space obtains rewards twice than in action space.

(3)  The two original IIL algorithms yield a deterministic policy in the tasks, which has advantages in terms of interpretability. Also, the learned policy is biased regarding different initial states. One can take 'reinforcement training' in this situation to optimize the existing policy.

(4)  In highly dynamic scenarios with continuous action, a stochastic policy like Gaussian policy presents better performance, possibly because of the offset against the suboptimality of human behaviors.

At the same time, two main problems are solved in the process of experiments:

(1)  Through a series of hypothesis-driven experiments, reasons and solutions for the instability are given. And the interpretability of the algorithm is also explored.

(2)  In the Reacher scenario, the observed oscillation phenomenon around the target point is eliminated by adapting the model and using two levels of correction and forced stop signals. The robustness of the model is enhanced, and the performance is increased by 25%.

Besides, in the KUKA-Reach task with high-dimensional input data, the agent combined with an encoder could learn to classify the position of the goal and plan the trajectory in very few training episodes, which also validates the IIL with convolutional neural network design.

Summarizing the entire thesis: two kinds of IIL algorithms are implemented and compared with RL methods in both statistic and dynamic scenarios, where action space may be discrete or continuous. It can be said that knowledge from human demonstration can significantly reduce the required training episodes and increase agents' performance. Also, the 'delaying enjoyment' problem can be eliminated, which is discussed in chapter 5.5.4.

In the meantime, for training an IIL agent to perform control tasks in a highly dynamic scenario, the difficulty lies in the early stage of training. The human demonstrator should provide correct feedback, even when the agent's behavior is highly random, which could be challenging in some tasks. At this time, a stochastic policy like Gaussian presents a better choice for shaping the initial policy.

Overall, the prerequisite of using IIL algorithms is the task, where humans should be able to give correction signals (not necessarily being able to perform the entire task). In this term, providing feedback in state space can sometimes dramatically simplify the demonstration. Nevertheless, a more complex model should also be taken into consideration. Especially when inputs are very high-dimensional, such as raw images, learning a precise model of state representation and prediction may become extremely hard.

# Reference

[1]   J. Kober, J. A. Bagnell, J. Peters, "Reinforcement Learning in Robotics: A survey, " in The International Journal of Robotics Research (IJRR0), 2013, pp. 1231–1237.

[2]   J. Ijspeert, J. Nakanishi, and S. Schaal, "Movement imitation with nonlinear dynamical systems in humanoid robots," in Proc. IEEE Int. Conf. Robotics and Automation (ICRA), 2002, pp. 1398–1403.

[3]   O. Vinyals, T. Ewalds, S. Bartunov, P. Georgiev, A. S. Vezhnevets, M. Yeo, A. Makhzani, H. K¨uttler, J. Agapiou, J. Schrittwieser, et al., "Starcraft ii: A new challenge for reinforcement learning," arXiv preprint arXiv:1708.04782, 2017.

[4]   J. MacGlashan, M. K. Ho, R. Loftin, B. Peng, G. Wang, D. L. Roberts, M. E. Taylor, and M. L. Littman, "Interactive learning from policy-dependent human feedback," in 34th International Conference on Machine Learning - Volume 70. JMLR. org, 2017, pp. 2285–2294.11.COACH journal paper

[5]   C. Wen, J. Lin, T. Darrell, D. Jayaraman, Y. Gao, "Fighting Copycat Agents in Behavioral Cloning from Observation Histories," arXiv preprint arXiv: 2010.14876v1, 2020.

[6]   P. d. Haan, D. Jayaraman, S. Levine, "Causal Confusion in Imitation Learning," arXiv preprint at arXiv:1905.11979, 2019.

[7]   S. Ross and D. Bagnell, "Efficient reductions for imitation learning," in International Conference on Artificial Intelligence and Statistics, 2010, pp. 661–668.

[8]   M. Kelly, C. Sidrane, K. Driggs-Campbell, and M. Kochenderfer, "HG-DAgger: Interactive imitation learning with human experts," in 2019 International Conference on Robotics and Automation (ICRA). IEEE, 2019, pp. 8077–8083.

[9]   R. Perez-Dattari, C. Celemin, G. Franzese, J. Ruiz-del Solar, and J. Kober, "Interactive learning of temporal features for control: Shaping policies and state representations from human feedback," IEEE Robotics & Automation Magazine, 2020.

[10] S. Jauhri, C. Celemin, J. Kober, "Interactive Imitation Learning in State-Space," arXiv preprint at arXiv: 2008.00524v2.

[11] Y. Liu, A. Gupta, P. Abbeel, and S. Levine, "Imitation from observation: Learning to imitate behaviors from raw video via context translation," In 2018 IEEE International Conference on Robotics and Automation (ICRA), pages 1118–1125. IEEE, 2018. doi:10.1109/ICRA.2018.8462901.

[12] Silver, D., Huang, A., Maddison, C., Guez, A., Sifre, L., van den Driessche, G., Schrittwieser, J., Antonoglou, I., Panneershelvam, V., Lanctot, M., Dieleman, S., Grewe, D., Nham, J., Kalchbrenner, N., Sutskever, I., Lillicrap, T., Leach, M., Kavukcuoglu, K., Graepel, T. and Hassabis, D. 2016," Mastering the game of Go with deep neural networks and tree search, " Nature, 529, 7587 (Jan. 2016), 484-489.

[13] V. Mnih, K. Kavukcuoglu, D. Silver, et. al. "Playing Atari with Deep Reinforcement Learning,"

arXiv preprint at arXiv: arXiv:1312.5602, 2013.

[14] Y. Song, M. Steinweg, E. Kaufmann, D. Scaramuzza, "Autonomous Drone Racing with Deep Reinforcement Learning," arXiv preprint at arXiv: 2103.08624, 2021.

[15] Williams, Ronald J. "Simple statistical gradient-following algorithms for connectionist reinforcement learning." Reinforcement Learning. Springer, Boston, MA, 1992. 5-32.

[16] T. Degris, M. White, R. S. Sutton, "Off-Policy Actor-Critic," arXiv preprint at arXiv: 1205.4839v5.

[17] V. Mnih, K. Kavukcuoglu, D. Silver, A. Graves, L. Antonoglou, D. Wierstra, M. Riedmiller, "Autonomous Drone Racing with Deep Reinforcement Learning," arXiv preprint at arXiv: 1312.5602v1.

[18] V. Mnih, K. Kavukcuoglu, D. Silver, A. A. Rusu, J. Veness, M. G. Bellemare, A. Graves, M. Riedmiller, A. K. Fidjeland, G. Ostrovski et al., "Human-level control through deep reinforcement learning," Nature, vol. 518, no. 7540, p. 529, 2015.

[19] S. Sharifzadeh, I. Chiotellis, R. Triebel, and D. Cremers, "Learning to drive using inverse reinforcement learning and deep q-networks," arXiv preprint arXiv:1612.03653, 2016.

[20] W. B. Knox and P. Stone, "Interactively shaping agents via human reinforcement: The TAMER framework," in Fifth International Conference on Knowledge Capture, 2009.

[21] S. Bhatnagar, R. S. Sutton, M. Ghavamzadeh, and M. Lee, "Natural actor–critic algorithms," Automatica, vol. 45, no. 11, pp. 2471–2482, 2009.

[22] F. Torabi, G. Warnell, P. Stone, "Behavioral Cloning from Observation," arXiv preprint at arXiv: 1805.01954, 2018.

[23] M. Deisenroth and C. Rasmussen, "A model-based and data-efficient approach to policy search," in ICML, 2011.

[24] M. Bojarski, D. D. Testa, D. Dworakowski, et al, "End to End learning for Self-Driving Cars," arXiv preprint at arXiv: 1604.07316, 2016.

[25] P. Haan, D. Jayaraman, S. Levine, "Causal Confusion in Imitation Learning," arXiv preprint arXiv: 1905.11979v2, 2019.

[26] S. Ross, G. J. Gordon, J. A. Bagnell, "A Reduction of Imitation Learning and Structured Prediction to No-Regret Online Learning".

[27] Gagniuc, Paul A. (2017), "Markov Chains: From Theory to Implementation and Experimentation, " USA, NJ: John Wiley & Sons. pp. 1–235.

[28] C. Celemin and J. Ruiz-del Solar, "An interactive framework for learning continuous actions policies based on corrective feedback," Journal of Intelligent & Robotic Systems, pp. 1–20, 2018.

[29] R. Perez-Dattari, C. Celemin, J. Ruiz-del Solar, J. Kober, "Interactive Learning with Corrective Feedback for Policies based on Deep Neural Networks," arXiv preprint at arXiv: 1810.00466, 2018.

[30] A. Nair, D. Chen, P. Agrawal, P. Isola, P. Abbeel, J. Malik, and S. Levine, "Combining self-supervised learning and imitation for vision-based rope manipulation," in 2017 IEEE

International Conference on Robotics and Automation (ICRA), pages 2146–2153. IEEE, 2017. doi: 10.1109/ICRA.2017.7989247.

[31] G. Brockman, V. Cheung, L. Pettersson, et al, "OpenAI Gym," arXiv preprint at arXiv: 1606.01540, 2016.

[32] E. Todorov, T. Erez, Y. Tassa, "MuJoCo: A physics engine for model-based control," University of Washington.

[33] Martin L Puterman, "Markov Decision Processes.: Discrete Stochastic Dynamic Programming," John Wiley & Sons, 2014.

# Acknowledgments

Ending here, still feeling unreal. Five years in the Department of Automotive Studies come to an end finally. As a senior undergraduate student, I am more clear than ever where I am going for the next stage of my life.

Firstly, I want to thank myself, who tries his best to be assiduous and handle everything in a 100% manner. I want to say: you still get a long way to go, but life has no limitations. Be brave, the only thing that is certain in your life is uncertainty. Try your best to do the rest 15% whenever you have already done the 85%, and you will not be regretful and say that you could have done something! I hope that when the stronger, smarter you look back at these words, you will feel lucky one day. The only thing you need is just time.

Equally important, I feel super lucky that I could work with my supervisor Prof. Yu, who is incredibly friendly and nice. As an external student out of the College of Electronics and Information Engineering, I got even more attention, which I appreciate a lot. From the daily greetings to the online Q&A, to the offline meetings every week, even if he is busy, this spirit has always inspired me to become a more diligent person who takes details seriously. Wish one day I could also help more people for their study and life.

Last but not least, I thank my parents for always supporting every decision I made in my life. When I could not help but want to give up, I realized that I was not alone. Just like Benjamin Button, no matter how far he has traveled to get home, he will always meet the older man who told him that he had been struck down by lightning seven times. No matter which corner of the world he went to, his home would always be the most reliable one. The harbor has never changed.

Time flies, five years have passed. The five years in Tongji are the most confused but brightest of youth. No matter which corner of the world I am in, I will definitely remember the things and responsibilities that I got here. For being a better man, I will spare no effort.

# Appendix

**The Initial States of the Environment in the 9 Testing Experiments After Each Training Episode**

In chapter 5.3.4, we discussed the influence of the initial states on the model performance. In hypothesis 1, the change of initial states during the training procedure of the unique 'imperfect' IIL agent is analyzed. The following is the original data of the four dimensions of the initial states.

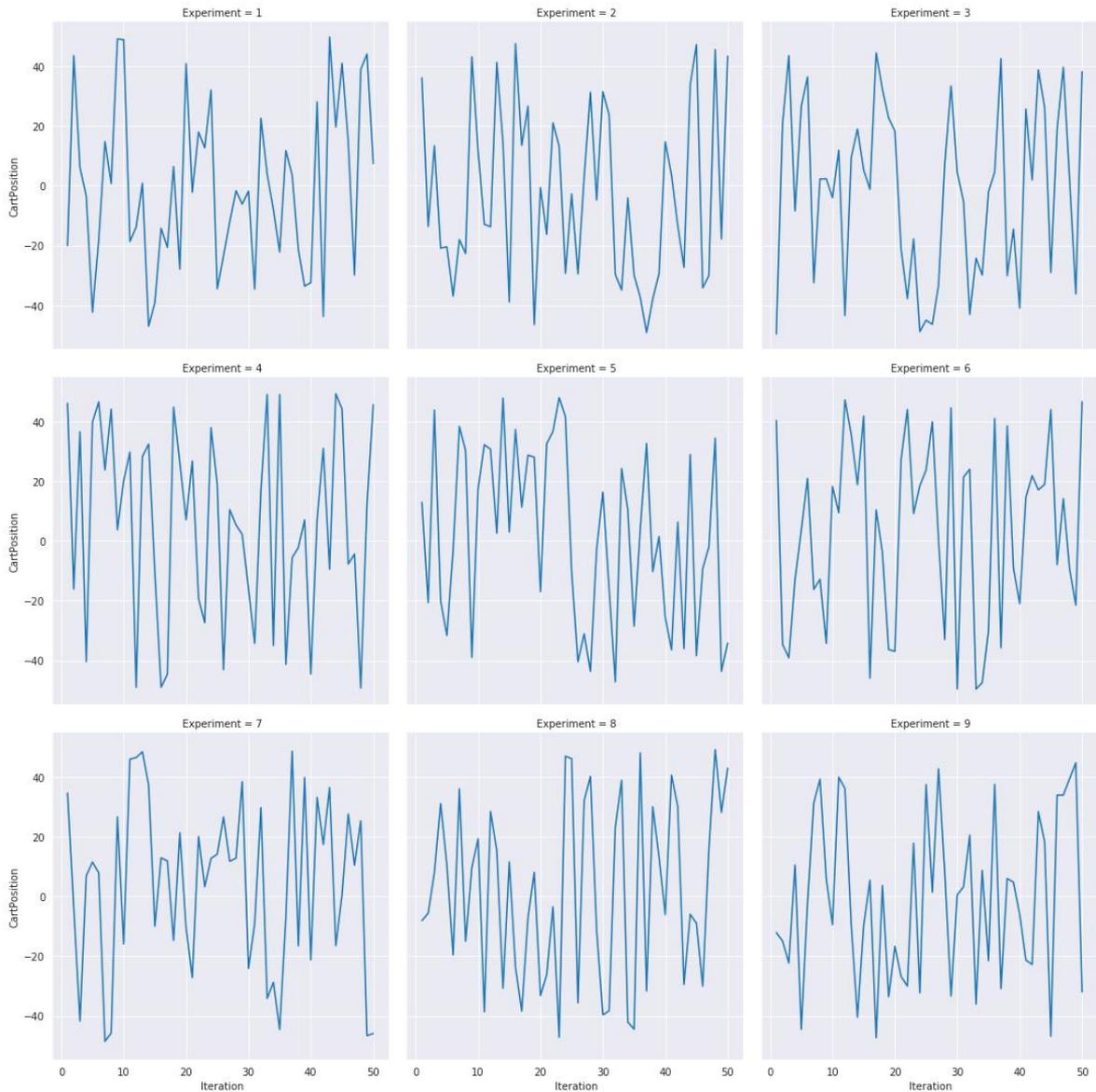

Fig. 8.1 The change of initial cart position

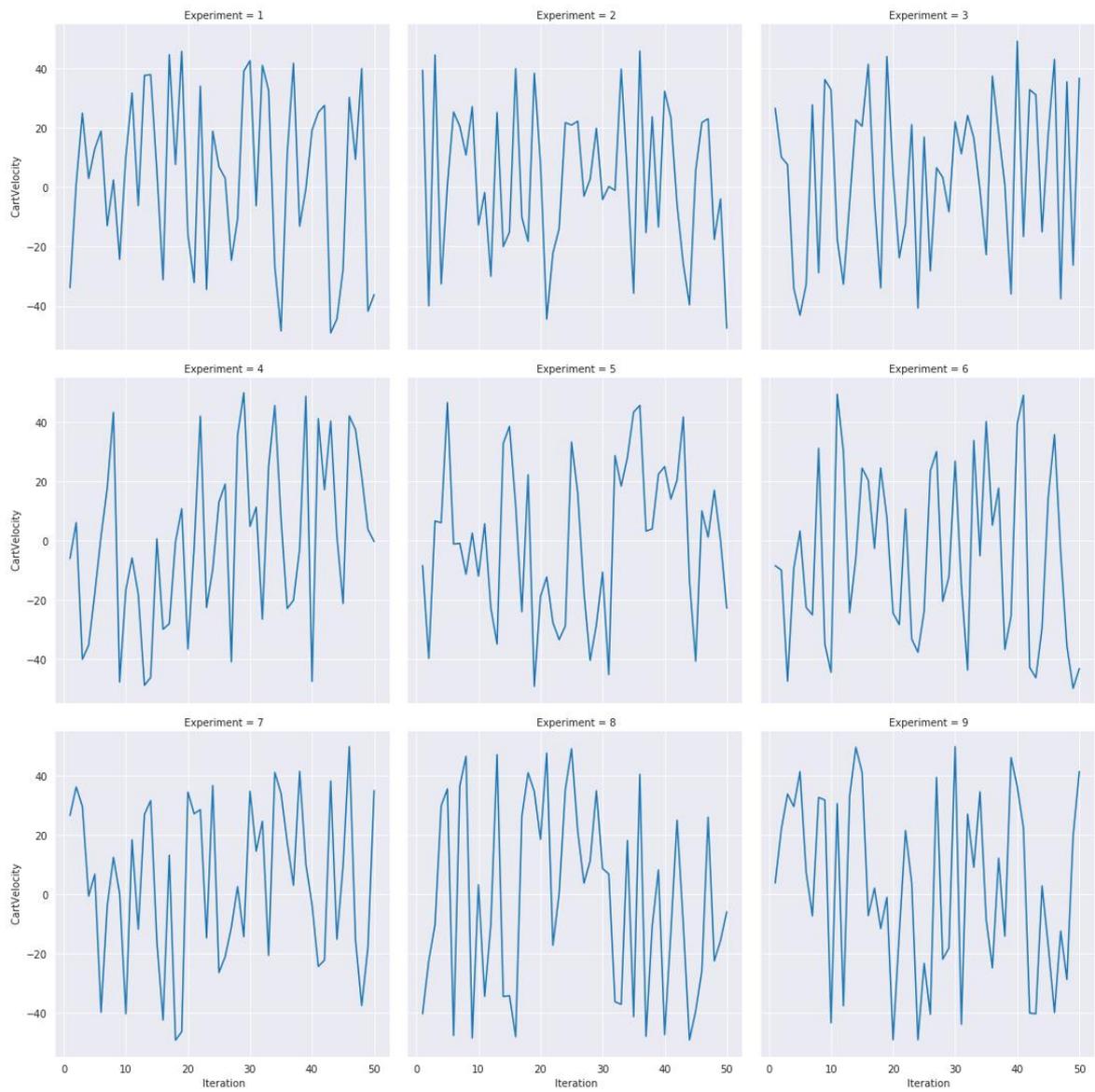

Fig. 8.2 The change of initial cart velocity

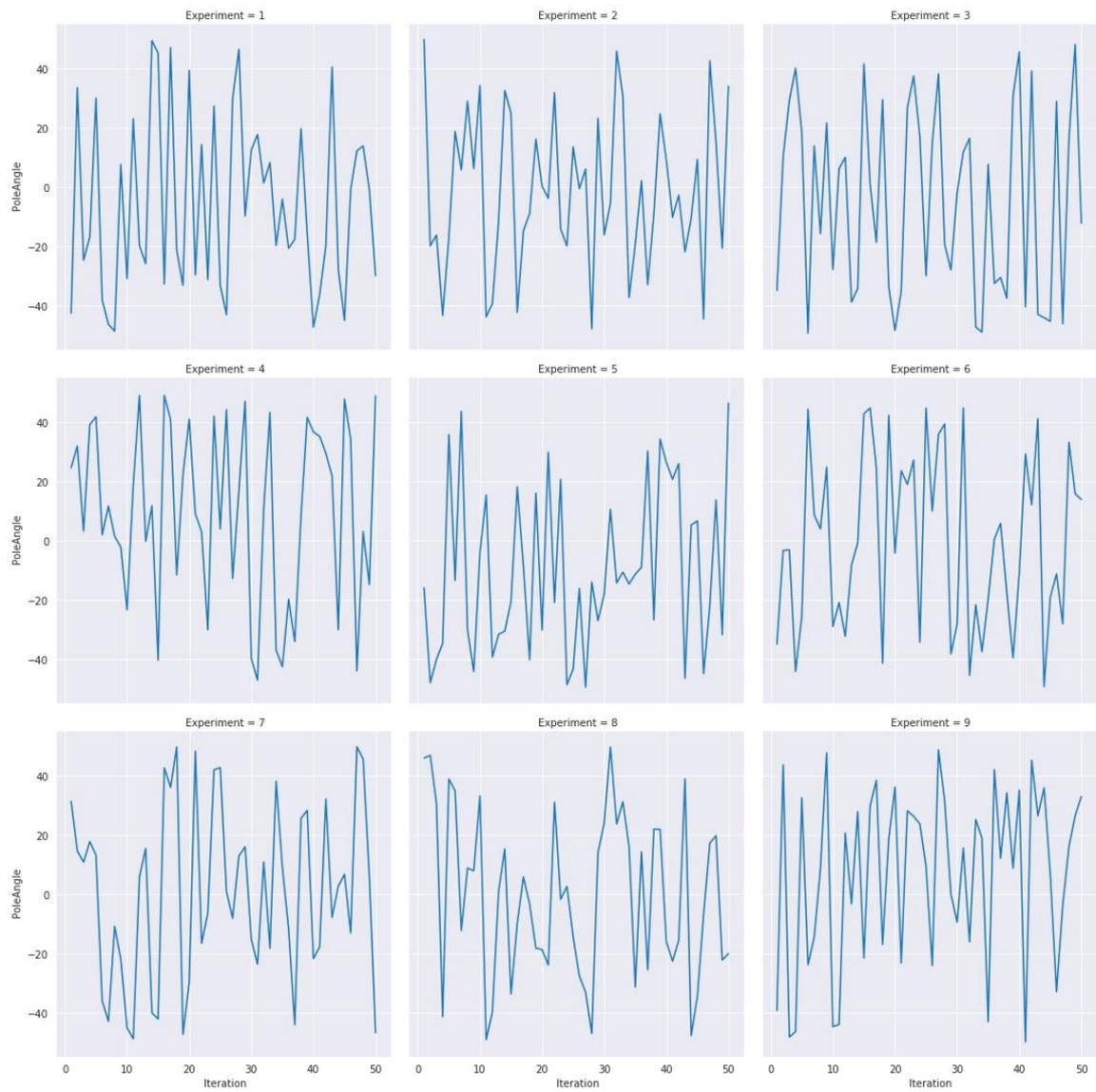

Fig. 8.3 The change of initial pole angle

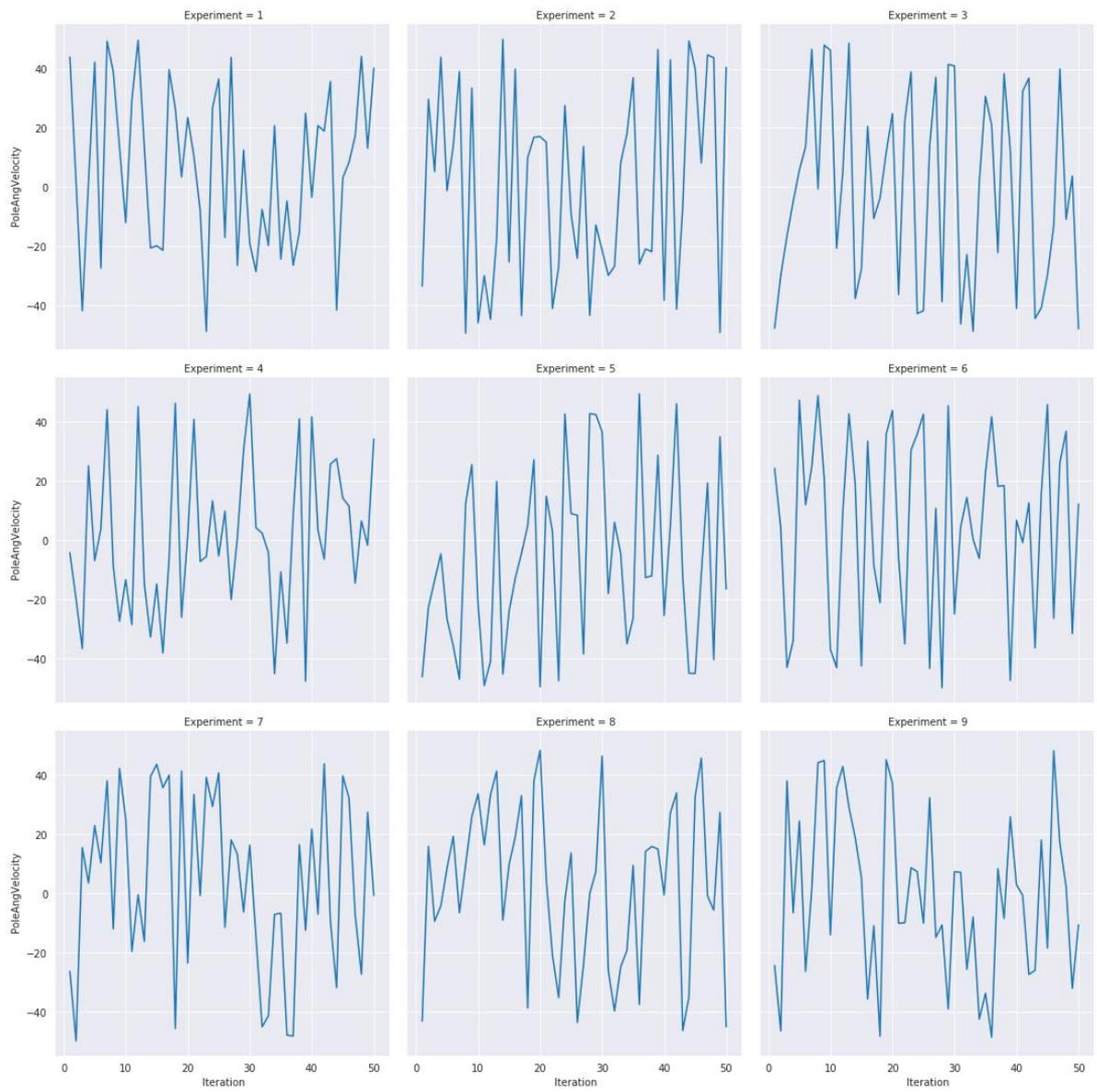

Fig. 8.4 The change of initial tip velocity

**Introduction to Basic Model-Free Reinforcement Learning Algorithms**

This chapter mainly introduces three basic model-free algorithms, which are used in this article as metrics, including Policy Gradient algorithm, Actor-Critic algorithm, which are based on gradient step on the RL objective, and the Deep Q-Network, which is based on value function learning.

Policy Gradient

First, the RL objective is described as: in a sequential decision-making process, find the parameters $\theta$ of the model, which maximizes the accumulated rewards obtained by the agent over the trajectory $\tau$.

$$\theta^* = argmax_\theta \, E_{\tau \sim p_{\theta(\tau)}} \left[ \sum_t r(s_t, a_t) \right] \qquad （8.1）$$

Write $E_{\tau \sim p_{\theta(\tau)}} \left[ \sum_t r(s_t, a_t) \right]$ as $J(\theta)$, we have：

$$J(\theta) = \mathbb{E}_{\tau \sim \pi_{\theta(\tau)}} \left[ r(\tau) \right] = \int \pi_\theta(\tau) r(\tau) d\tau \xrightarrow{\textit{partial differentiation}} \nabla_\theta J(\theta) = \int \nabla_\theta \pi_\theta(\tau) r(\tau) d\tau$$
$$= \int \pi_\theta \nabla_\theta \log \pi_\theta(\tau) r(\tau) d\tau$$

$$（8.2）$$

According to the equation (8.3) ($log \; trick$)：

$$p_\theta(\tau) \nabla_\theta \log p_\theta(\tau) = p_\theta(\tau) \frac{\nabla_\theta p_\theta(\tau)}{p_\theta(\tau)} = \nabla_\theta p_\theta(\tau)$$

$$（8.3）$$

Equation (8.2) can be written as：

$$\nabla_\theta J(\theta) = \mathbb{E}_{\tau \sim \pi_{\theta(\tau)}} [\nabla_\theta \log \pi_\theta(\tau) r(\tau)]$$
$$= \mathbb{E}_{\tau \sim \pi_{\theta(\tau)}} \left[ \left( \sum_{t=1}^T \nabla_\theta \log \pi_\theta(a_t | s_t) \right) \left( \sum_{t=1}^T r(s_t, a_t) \right) \right]$$

$$（8.4）$$

So that $\nabla_\theta J(\theta)$ is written as a value that can be estimated by sampling. Optimizing $\theta$ by gradient descent：

$$\theta_{k+1} = \theta_k + \alpha \nabla_\theta J(\pi_\theta) \longleftarrow \approx \frac{1}{N} \sum_{i=1}^N \sum_{t=1}^T \nabla_\theta \log \pi_\theta(a_t^i | s_t^i) r(s_t^i, a_t^i)$$

$$（8.5）$$

In the equation (8.5), 'reward to go' $\hat{r} = \sum_{t'=t}^T r(s, a)$ is often used to simplify estimation.

Actor-Critic

In the Policy Gradient algorithm, reward form single samples $r(s_t^i, a_t^i)$ is used to substitute a whole distribution (Monto Carlo Estimation). In practice, it suffers from a problem of high variance. As solutions, algorithms like 'Trust Region Policy Gradient' are introduced.
Also, using Actor-Critic algorithm can reduce the variance, which uses the advantage function as estimation with state-related baseline.

$$A(s_t, a_t) = Q(s_t, a_t) - V(s_t) \qquad （8.6）$$

Equation (8.6) is the advantage function, which is the difference between Q Function and the Value Function. This is the 'Critic' part of the algorithm.

And the Q Function can be approximated by equation (8.7)

$$Q(s_t, a_t) \approx r(s_t, a_t) + V(s_{t+1})$$

（8.7）

The gradient step is taken on the 'Actor' part, which uses Advantage Function instead of 'reward to go' and reduces variance by means of state-related baseline.

$$\nabla_\theta J(\theta) \approx \frac{1}{N} \sum_{i=1}^{N} \sum_{t=1}^{T} \nabla_\theta \log \pi_\theta(a_t^i | s_t^i) A_\omega(s_t^i, a_t^i)$$

（8.8）

Deep Q-Learning

Another model-free RL algorithm is Q Function-based, which has no explicit policy gradient step. For example, in Deep Q Learning, Bellman Backup is optimized using dynamic programming approach.

$$Q^{new}(s_t, a_t) \leftarrow \underbrace{Q(s_t, a_t)}_{\text{old value}} + \underbrace{\alpha}_{\text{learning rate}} \cdot \left( \underbrace{\overbrace{\underbrace{r_t}_{\text{reward}} + \underbrace{\gamma}_{\text{discount factor}} \cdot \underbrace{\max_a Q(s_{t+1}, a)}_{\text{estimate of optimal future value}} - \underbrace{Q(s_t, a_t)}_{\text{old value}}}^{\text{temporal difference}}}_{\text{new value (temporal difference target)}} \right)$$

（8.9）

As mentioned in chapter 5.5, in DQN, the optimal future value is estimated from the current Q network, which is being updated. This causes the problem of optimizing a moving target. Using the target network chosen from a few steps ago, the so-named Double DQN gains much more robustness in practice.